\documentclass[final]{elsarticle}
\makeatletter
\def\ps@pprintTitle{%
 \let\@oddhead\@empty
 \let\@evenhead\@empty
 \def\@oddfoot{}%
 \let\@evenfoot\@oddfoot}
\makeatother

\usepackage[utf8]{inputenc}
\usepackage{fullpage}
\usepackage{amsmath}
\usepackage{amsfonts}
\usepackage{amssymb}
\usepackage{graphicx}
\usepackage{algorithmic}
\usepackage{algorithm}
\usepackage{color}
\usepackage{subfigure}
\usepackage{multirow}
\usepackage{booktabs}
\usepackage{tikz}
\usepackage[raggedright]{sidecap}
\sidecaptionvpos{figure}{c} 
\usepackage{url} 

\newcommand{\argmin}{\operatornamewithlimits{arg\ min}}

\def\x{{\mathbf x}}
\def\y{{\mathbf y}}
\def\h{{\mathbf h}}
\def\b{{\mathbf b}}
\def\W{{\mathbf W}}
\def\R{{\mathbf R}}

\definecolor{darkgreen}{rgb}{0,0.6,0.2}

\begin{document}
\begin{frontmatter}
\title{An overview and comparative analysis of Recurrent Neural Networks for Short Term Load Forecasting}

\author[uit]{Filippo Maria Bianchi\thanks{filippo.m.bianchi@uit.no}}
\author[sap]{Enrico Maiorino}
\author[uit]{Michael C. Kampffmeyer}
\author[sap]{Antonello Rizzi}
\author[uit]{Robert Jenssen}
\address[uit]{Machine Learning Group, Dept. of Physics and Technology, UiT The Arctic University of Norway, Troms\o{}, Norway}
\address[sap]{Dept. of Information Engineering, Electronics and Telecommunications, Sapienza University, Rome, Italy}

\begin{abstract}
The key component in forecasting demand and consumption of resources in a supply network is an accurate prediction of real-valued time series. 
Indeed, both service interruptions and resource waste can be reduced with the implementation of an effective forecasting system. Significant research has thus been devoted to the design and development of methodologies for short term load forecasting over the past decades.
A class of mathematical models, called Recurrent Neural Networks, are nowadays gaining renewed interest among researchers and they are replacing many practical implementation of the forecasting systems, previously based on static methods.
Despite the undeniable expressive power of these architectures, their recurrent nature complicates their understanding and poses challenges in the training procedures.
Recently, new important families of recurrent architectures have emerged and their applicability in the context of load forecasting has not been investigated completely yet.
In this paper we perform a comparative study on the problem of Short-Term Load Forecast, by using different classes of state-of-the-art Recurrent Neural Networks. 
We test the reviewed models first on controlled synthetic tasks and then on different real datasets, covering important practical cases of study.
We provide a general overview of the most important architectures and we define guidelines for configuring the recurrent networks to predict real-valued time series.
\end{abstract}

\end{frontmatter}
\section{Introduction}

Forecasting the demand of resources within a distribution network of energy, telecommunication or transportation is of fundamental importance for managing the limited availability of the assets.
An accurate Short Term Load Forecast (STLF) system \cite{DeGooijer2006443} can reduce high cost of over- and under-contracts on balancing markets due to load prediction errors.
Moreover, it keeps power markets efficient and provides a better understanding of the dynamics of the monitored system \cite{simchi1999designing}. 
On the other hand, a wrong prediction could cause either a load overestimation, which leads to the excess of supply and consequently more costs and contract curtailments for market participants, or a load underestimation, resulting in failures in gathering enough provisions, thereby more costly supplementary services \cite{bunn2000forecasting, ruiz2008short}. 
These reasons motivated the research of forecasting models capable of reducing this financial distress, by increasing the load forecasting accuracy even by a small percent \cite{deihimi2012application, peng2014novel, shen2008interday, bianchi2015prediction, 7286732}. 

The load profile generally follows cyclic and seasonal patterns related to human activities and can be represented by a real-valued time series. 
The dynamics of the system generating the load time series can vary significantly during the observation period, depending on the nature of the system and on latent, external influences. 
For this reason, the forecasting accuracy can change considerably among different samples even when using the same prediction model \cite{deihimi2013short}.
Over the past years, the STLF problem has been tackled in several research areas \cite{jan2005did} by means of many different model-based approaches, each one characterized by different advantages and drawbacks in terms of prediction accuracy, complexity in training, sensitivity to the parameters and limitations in the tractable forecasting horizon \cite{2017arXiv170208025D}. 

Autoregressive and exponential smoothing models represented for many years the baseline among systems for time series prediction \cite{hyndman2008forecasting}.
Such models require to properly select the lagged inputs to identify the correct model orders, a procedure which demands a certain amount of skill and expertise \cite{box2011time}. 
Moreover, autoregressive models make explicit assumptions about the nature of system under exam.
Therefore, their use is limited to those settings in which such assumptions hold and where \textit{a-priori} knowledge on the system is available \cite{box1964analysis}. 
\citet{taylor2008comparison} showed that for long forecasting horizons a very basic averaging model, like AutoRegressive Integrated Moving Average or Triple Exponential Smoothing, can outperform more sophisticated alternatives.
However, in many complicated systems the properties of linearity and even stationarity of the analyzed time series are not guaranteed. 
Nonetheless, given their simplicity, autoregressive models have been largely employed as practical implementations of forecast systems.

The problem of time series prediction has been approached within a function approximation framework, by relying on the embedding procedure proposed by \citet{takens1981detecting}. 
Takens' theorem transforms the prediction problem from time extrapolation to phase space interpolation. In particular, by properly sampling a time dependent quantity $s(t)$, it is possible to predict the value of the $k$-th sample from the previous samples, given an appropriate choice of the sampling frequency $\tau$ and the number of samples $m$: $s[k] = f(s[k-\tau], \ldots, s[k-m \cdot \tau])$.
Through the application of phase-space embedding, regression methods, such as Support Vector Regression (an extension of Support Vector Machines in the continuum) have been applied in time series prediction \cite{sapankevych2009time}, either by representing the sequential input as a static domain, described by frequency and phase, or by embedding sequential input values in time windows of fixed length.
The approach can only succeed if there are no critical temporal dependencies exceeding the windows length, making the SVM unable to learn an internal state representation for sequence learning tasks involving time lags of arbitrary length. 
Other universal function approximators such as Feed-Forward Artificial Neural Networks \cite{Hornik1989359} and ANFIS (Adaptive Network-Based Fuzzy Inference System) \cite{jang1993anfis} have been employed in time series prediction tasks by selecting a suitable interval of past values from the time series as the inputs and by training the network to forecast one or a fixed number of future values \cite{Zhang199835,Hippert2001,Law2000331, Tsaur2002397, kon2005neural, Palmer2006781, Claveria2014220}. 
The operation is repeated to forecast next values by translating the time window of the considered inputs \cite{Kourentzes2013198}. 
While this approach proved to be effective in many circumstances \cite{DiazRobles20088331,plummer2000time,TEIXEIRA2012445,JTR:JTR2016}, it does not treat temporal ordering as an explicit feature of the time series and, in general, is not suitable in cases where the time series have significantly different lengths. 
On this account, a Recurrent Neural Network (RNN) is a more flexible model, since it encodes the temporal context in its feedback connections, which are capable of capturing the time varying dynamics of the underlying system \cite{schafer2007approximators,bianchi2017temporal}. 

RNNs are a special class of Neural Networks characterized by internal self-connections, which can, in principle, any nonlinear dynamical system, up to a given degree of accuracy \cite{Schafer2007}. 
RNNs and their variants have been used in many contexts where the temporal dependency in the data is an important implicit feature in the model design.
Noteworthy applications of RNNs include sequence transduction \cite{graves2012sequence}, language modeling \cite{Graves2013,Pascanu_Mikolov_Bengio, tomavs2012statistical,sutskever2011generating}, speech recognition \cite{graves2011practical}, learning word embeddings \cite{mikolov2013distributed}, audio modeling \cite{oord2016wavenet}, handwriting recognition \cite{graves2009offline,graves2008unconstrained}, and image generation \cite{gregor2015draw}. 
In many of these works a popular variant of RNN was used, called Long-Short Term Memory \cite{hochreiter1997long}.
This latter has recently earned significant attention due to its capability of storing information for very long periods of time. 

As an RNN processes sequential information, it performs the same operations on every element of the input sequence.
Its output, at each time step, depends on previous inputs and past computations. 
This allows the network to develop a memory of previous events, which is implicitly encoded in its hidden state variables. 
This is certainly different from traditional feedforward neural networks, where it is assumed that all inputs (and outputs) are independent of each other. 
Theoretically, RNNs can remember arbitrarily long sequences.
However, their memory is in practice limited by their finite size and, more critically, by the suboptimal training of their parameters. 
To overcome memory limitations, recent research efforts have led to the design of novel RNN architectures, which are equipped with an external, permanent memory capable of storing information for indefinitely long amount of time \cite{DBLP:journals/corr/WestonCB14, DBLP:journals/corr/GravesWD14}.

Contrarily to other linear models adopted for prediction, RNNs can learn functions of arbitrary complexity and they can deal with time series data possessing properties such as saturation or exponential effects and nonlinear interactions between latent variables.
However, if the temporal dependencies of data are prevalently contained in a finite and small time interval, the use of RNNs can be unnecessary. 
In these cases performances, both in terms of computational resources required and accuracy, are generally lower than the ones of time-window approaches, like ARIMA, SVM, Multi-Layer Perceptron and ANFIS. 
On the other hand, in many load forecasting problems the time series to be predicted are characterized by long temporal dependencies, whose extent may vary in time or be unknown in advance.
In all these situations, the use of RNNs may turn out to be the best solution. 

Despite the STLF problem has been one of the most important applications for both early RNNs models \cite{Gers2001} and most recent ones \cite{2017arXiv170404110F}, an up-to-date and comprehensive analysis of the modern RNN architectures applied to the STLF problem is still lacking. 
In several recent works on STFL, NARX networks (see Sec. \ref{sec:narx}) or Echo State Networks (see Sec. \ref{sec:esn}) are adopted for time series prediction and their performance is usually compared with standard static models, rather than with other RNN architectures.
With this paper, we aim to fill these gaps by performing a comparative study on the problem of STLF using different classes of state-of-the-art RNNs. 
We provide an introduction to the RNN framework, presenting the most important architectures and their properties.
We also furnish the guidelines for configuring and training the different RNN models to predict real-valued time series.
In practice, we formulate the STLF problem as the prediction of a real-valued univariate time series, given its past values as input.
In some cases, beside the time series of past target values, additional ``context'' time series are fed to the network in order to provide exogenous information related to the environment in which the system to be modeled operates. 

The paper is structured as follows. 

In Sec. \ref{sec:rnn} we provide a general overview of a standard RNN architecture and we discuss its general properties. 
We also discuss the main issues encountered in the training phase, the most common methodologies for learning the model parameters and common ways of defining the loss function to be optimized during the training.

In Sec. \ref{sec:rnn_architectures}, we present the most basic architecture, called Elman RNN, and then we analyze two important variants, namely the Long-Short Term Memory and Gated Recurrent Units networks. 
Despite the recent popularity of these architectures \cite{greff2015lstm}, their application to prediction of real-valued time series has been limited so far \cite{malhotra2015long}. 
For each RNN, we provide a brief review, explaining its main features, the approaches followed in the training stage and a short list of the main works concerning time series prediction in which the specific network has been applied.

Successively, in Sec. \ref{sec:other_architectures} we illustrate two particular RNN architectures, which differ from the previous ones, mainly due to their training procedure.
In particular, we analyze the Nonlinear AutoRegressive with eXogenous inputs (NARX) neural network and the Echo State Network (ESN). 
These architectures have been successfully applied in the literature of time series prediction and they provide important advantages with respect to traditional models, due to their easy applicability and fast training procedures.

In Sec. \ref{sec:synth_data} we describe three synthetic datasets, used to test and to compare the computational capabilities of the five RNN architectures in a controlled environment.

In Sec. \ref{sec:real_data}, we present three real-world datasets of time series relative to the load profile in energy distribution and telecommunication networks. 
For each dataset, we perform a series of analysis with the purpose of choosing a suitable preprocessing for the data.

Sec. \ref{sec:experiments} is dedicated to the experiments and to the discussion of the performance of the RNN models.
The first part of the experimental section focuses on the benchmark tests, while in the second part we employ the RNNs to solve STLF tasks on real-world time series.

Finally, in Sec. \ref{sec:conclusions} we discuss our conclusions.

\section{Properties and Training in Recurrent Neural Networks}
\label{sec:rnn}

RNNs are learning machines that recursively compute new states by applying transfer functions to previous states and inputs. 
Typical transfer functions are composed by an affine transformation followed by a nonlinear function, which are chosen depending on the nature of the particular problem at hand. 
It has been shown by \citet{maass2007computational} that RNNs possess the so-called universal approximation property, that is, they are capable of approximating arbitrary nonlinear dynamical systems (under loose regularity conditions) with arbitrary precision, by realizing complex mappings from input sequences to output sequences \cite{siegelmann1991turing}. 
However, the particular architecture of an RNN determines how information flows between different neurons and its correct design is crucial in the realization of a robust learning system. 
In the context of prediction, an RNN is trained on input temporal data $\x(t)$ in order to reproduce a desired temporal output $\y(t)$. $\y(t)$ can be any time series related to the input and even a temporal shift of $\x(t)$ itself. 
The most common training procedures are gradient-based, but other techniques have been proposed, based on derivative-free approaches or convex optimization \cite{schmidhuber2007training, jaeger2001echo}. 
The objective function to be minimized is a loss function, which depends on the error between the estimated output $\hat \y(t)$ and the actual output of the network $\y(t)$.
An interesting aspect of RNNs is that, upon suitable training, they can also be executed in generative mode, as they are capable of reproducing temporal patterns similar to those they have been trained on \cite{gregor2015draw}. 

\begin{SCfigure}[1.2][!ht]
\centering
    \includegraphics[scale=0.8, keepaspectratio]{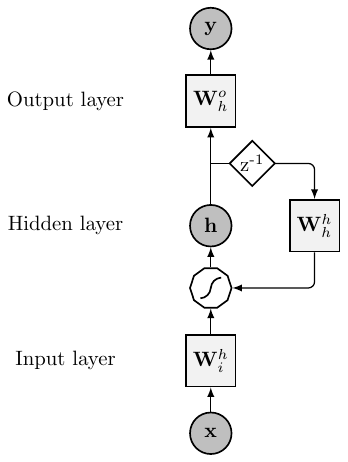}
    \caption{Schematic depiction of a simple RNN architecture. The circles represent input $\mathbf{x}$, hidden, $\mathbf{h}$, and output nodes, $\mathbf{y}$, respectively. The solid squares $\mathbf{W}_{i}^{h}$, $\mathbf{W}_{h}^{h}$ and $\mathbf{W}_{h}^{o}$ are the matrices which represent input, hidden and output weights respectively. Their values are commonly tuned in the training phase through gradient descent. The polygon represents the non-linear transformation performed by neurons and $\text{z}^{\text{-1}}$ is the unit delay operator.}
    \label{fig:rnn}
\end{SCfigure}

The architecture of a simple RNN is depicted in Fig.~\ref{fig:rnn}.
%
%
In its most general form an RNN can be seen as a weighted, directed and cyclic graph that contains three different kinds of nodes, namely the input, hidden and output nodes \cite{NIPS2016_6303}.
Input nodes do not have incoming connections, output nodes do not have outgoing connections, hidden nodes have both. 
An edge can connect two different nodes which are at the same or at different time instants. 
In this paper, we adopt the time-shift operator $\text{z}^{n}$ to represent a time delay of $n$ time steps between a source and a destination node. 
Usually $n=-1$, but also lower values are admitted and they represent the so called skip connections \cite{DBLP:journals/corr/KoutnikGGS14}. 
Self-connecting edges always implement a lag operator with $|n| \geq 1$. 
In some particular cases, the argument of the time-shift operator is positive and it represents a forward-shift in time \cite{sutskever2010temporal}. 
This means that a node receives as input the content of a source node in a future time interval. 
Networks with those kind of connections are called bidirectional RNNs and are based on the idea that the output at a given time may not only depend on the previous elements in the sequence, but also on future ones \cite{schuster1997bidirectional}. 
These architectures, however, are not reviewed in this work as we only focus on RNNs with $n=-1$.

While, in theory, an RNN architecture can model any given dynamical system, practical problems arise during the training procedure, when model parameters must be learned from data in order to solve a target task.
Part of the difficulty is due to a lack of well established methodologies for training different types of models.
This is also because a general theory that might guide designer decisions has lagged behind the feverish pace of novel architecture designs \cite{2016arXiv161101232S, DBLP:journals/corr/Lipton15}.
A large variety of novel strategies and heuristics have arisen from the literature in the past the years \cite{Montavon2012,Scardapane201781} and, in many cases, they may require a considerable amount of expertise from the user to be correctly applied.
While the standard learning procedure is based on gradient optimization, in some RNN architectures the weights are trained following different approaches \cite{WIDM:WIDM1200,jaeger2002tutorial}, such as real-time recurrent learning \cite{williams1989learning}, extended Kalman filters \cite{haykin2001kalman} or evolutionary algorithms \cite{john1992holland}, and in some cases they are not learned at all \cite{lukovsevivcius2009reservoir}. 

\subsection{Backpropagation Through Time}
\label{sec:bppt}

\begin{figure}[htp!]
    \centering
    \includegraphics[scale=0.8, keepaspectratio]{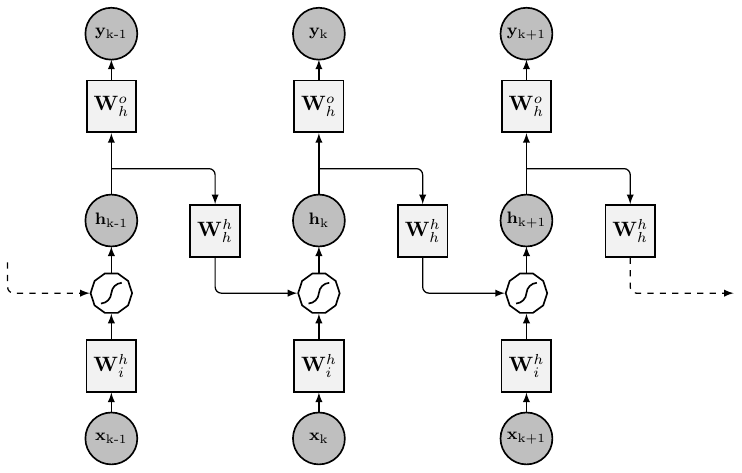}
    \caption{The diagram depicts the RNN from Fig.~\ref{fig:rnn}, being unfolded (or unrolled) into a FFNN. 
    As we can see from the image, each input $\mathbf{x}_t$ and output $\mathbf{y}_t$ are relative to different time intervals. 
    Unlike a traditional deep FFNN, which uses different parameters in each layer, an unfolded RNN shares the same weights across every time step. 
    In fact, the input weights matrix $\W_i^h$, the hidden weights matrix $\W_h^h$ and the output weights matrix $\W_h^o$ are constrained to keep the same values in each time interval.}
    \label{fig:BPPT}
\end{figure}

Gradient-based learning requires a closed-form relation between the model parameters and the loss function. 
This relation allows to propagate the gradient information calculated on the loss function back to the model parameters, in order to modify them accordingly. 
While this operation is straightforward in models represented by a directed acyclic graph, such as a FeedForward Neural Network (FFNN), some caution must be taken when this reasoning is applied to RNNs, whose corresponding graph is cyclic. 
Indeed, in order to find a direct relation between the loss function and the network weights, the RNN has to be represented as an equivalent infinite, acyclic and directed graph. 
The procedure is called \textit{unfolding} and consists in replicating the network's hidden layer structure for each time interval, obtaining a particular kind of FFNN. 
The key difference of an unfolded RNN with respect to a standard FFNN is that the weight matrices are constrained to assume the same values in all replicas of the layers, since they represent the recursive application of the same operation.

Fig.~\ref{fig:BPPT} depicts the unfolding of the RNN, previously reported in Fig.~\ref{fig:rnn}. 
Through this transformation the network can be trained with standard learning algorithms, originally conceived for feedforward architectures.
This learning procedure is called Back Propagation Through Time (BPTT)~\cite{rumelhart1985learning} and is one of the most successful techniques adopted for training RNNs. 
However, while the network structure could in principle be replicated an infinite number of times, in practice the unfolding is always truncated after a finite number of time instants.
This maintains the complexity (depth) of the network treatable and limits the issue of the vanishing gradient (as discussed later).
In this learning procedure called Truncated BPPT~\cite{williams1990efficient}, the folded architecture is repeated up to a given number of steps $\tau_b$, with $\tau_b$ upperbounded by the time series length $T$. 
The size of the truncation depends on the available computational resources, as the network grows deeper by repeating the unfolding, and on the expected maximum extent of time dependencies in data.
For example, in a periodic time series with period $t$ it may be unnecessary, or even detrimental, to set $\tau_b > t$.

Another variable we consider is the frequency $\tau_f$ at which the BPTT calculates the backpropagated gradients.
In particular, let us define with $\mathrm{BPTT}(\tau_b,\tau_f)$ the truncated backpropagation that processes the sequence one time step at a time, and every $\tau_f$ time steps, it runs BPTT for $\tau_b$ time steps \cite{sutskever2013training}.
Very often the term $\tau_f$ is omitted in the literature, as it is assumed equal to 1, and only the value for $\tau_b$ is specified.
We refer to the case $\tau_f = 1$ and $\tau_b = n$ as \emph{true} BPTT, or $\mathrm{BPTT}(n,1)$.

In order to improve the computational efficiency of the BPTT, the ratio $\tau_b / \tau_f$ can be decremented, effectively reducing the frequency of gradients evaluation.
An example, is the so-called \emph{epochwise} BPTT or $\mathrm{BPTT}(n,n)$, where $\tau_b = \tau_f$ \cite{williams1995gradient}.
In this case, the ratio $\tau_b / \tau_f = 1$.
However, the learning procedure is in general much less accurate than $\mathrm{BPTT}(n,1)$, since the gradient is truncated too early for many values on the boundary of the backpropagation window.

A better approximation of the true BPTT is reached by taking a large difference $\tau_b - \tau_f$, since no error in the gradient is injected for the earliest $\tau_b - \tau_f$ time steps in the buffer.
A good trade-off between accuracy and performance is $\mathrm{BPTT}(2n,n)$, which keeps the ratio $\tau_b / \tau_f = 2$ sufficiently close to 1 and the difference $\tau_b - \tau_f = n$ is large as in the true BPTT \cite{williams1990efficient}.
Through preliminary experiments, we observed that $\mathrm{BPTT}(2n,n)$ achieves comparable performance to $\mathrm{BPTT}(n,1)$, in a significantly reduced training time.
Therefore, we followed this procedure in all our experiments.

\subsection{Gradient descent and loss function}
\label{sec:grad_descent}

Training a neural network commonly consists in modifying its parameters through a gradient descent optimization, which minimizes a given loss function that quantifies the accuracy of the network in performing the desired task. 
The gradient descent procedure consists in repeating two basic steps until convergence is reached. 
First, the loss function $L_k$ is evaluated on the RNN configured with weights $\W_k$, when a set of input data $\mathcal{X}_k$ are processed (forward pass).
Note that with $\W_k$ we refer to \textit{all} network parameters, while the index $k$ identifies their values at epoch $k$, as they are updated during the optimization procedure. 
In the second step, the gradient $\partial L_k/ \partial \W_k$ is back-propagated through the network in order to update its parameters (backward pass). 

In a time series prediction problem, the loss function evaluates the dissimilarity between the predicted values and the actual future values of the time series, which is the ground truth. 
The loss function can be defined as
\begin{equation}
  \label{eq:loss_fun}
  L_k = E\left( \mathcal{X}_k , \mathcal{Y}_k^*; \W_k \right) + R_{\lambda}\left(\W_k\right),
\end{equation}
where $E$ is a function that evaluates the prediction error of the network when it is fed with inputs in $\mathcal{X}_k$, in respect to a desired response $\mathcal{Y}_k^*$.
$R_{\lambda}$ is a regularization function that depends on a hyperparameter $\lambda$, which weights the contribution of the regularization in the total loss. 

The error function $E$ that we adopt in this work is Mean Square Error (MSE). It is defined as
\begin{equation}
  \label{eq:MSE}
  \mathrm{MSE}(\mathcal{Y}_k,\mathcal{Y}_k^*) = \frac{1}{|\mathcal{X}_k|} \sum \limits_{\x \in \mathcal{X}_k} \left( \y_\x - \y_\x^* \right)^2,
\end{equation}
where $\y_\x \in \mathcal{Y}_k$ is the output of the RNN (configured with parameters $\W_k$) when the input $\x \in \mathcal{X}_k$ is processed and $\y_\x^* \in \mathcal{Y}_k^*$ is the ground-truth value that the network must learn to reproduce.

The regularization term $R_{\lambda}$ introduces a bias that improves the generalization capabilities of the RNN, by reducing overfitting on the training data.  
In this work, we consider four types of regularization:

\begin{enumerate}
  \item $\mathrm{L}_1$: the regularization term in Eq. \ref{eq:loss_fun} has the form $R_{\lambda}(\W_k) = \lambda_1\|\W_k \|_1$. $\mathrm{L}_1$ regularization enforces sparsity in the network parameters, is robust to noisy outliers and it can possibly deliver multiple optimal solutions. However, this regularization can produce unstable results, in the sense that a small variation in the training data can yield very different outcomes. 
  \item $\mathrm{L}_2$: in this case, $R_{\lambda}(\W_k) = \lambda_2\| \mathbf{W_k} \|_2$. 
  This function penalizes large magnitudes in the parameters, favouring dense weight matrices with low values. 
  This procedure is more sensitive to outliers, but is more stable than $\mathrm{L}_1$. 
  Usually, if one is not concerned with explicit features selection, the use of $\mathrm{L}_2$ is preferred. 
  \item \textit{Elastic net penalty:} combines the two regularizations above, by joining both $\mathrm{L}_1$ and $\mathrm{L}_2$ terms as $R_{\lambda}(\W_k) = \lambda_1 \left\|\W_k\right\|_1 + \lambda_2 \|\W_k\|_2$. 
  This regularization method overcomes the shortcomings of the $\mathrm{L}_1$ regularization, which selects a limited number of variables before it saturates and, in case of highly correlated variables, tends to pick only one and ignore the others. 
  Elastic net penalty generalizes the $\mathrm{L}_1$ and $\mathrm{L}_2$ regularization, which can be obtained by setting $\lambda_2=0$ and $\lambda_1=0$, respectively.
  \item \textit{Dropout:} rather than defining an explicit regularization function $R_{\lambda}(\cdot)$, dropout is implemented by keeping a neuron active during each forward pass in the training phase with some probability. Specifically, one applies a randomly generated mask to the output of the neurons in the hidden layer. The probability of each mask element to be 0 or 1 is defined by a hyperparameter $p_\mathrm{drop}$. Once the training is over, the activations are scaled by $p_\mathrm{drop}$ in order to maintain the same expected output.
  Contrarily to feedforward architectures, a naive dropout in recurrent layers generally produces bad performance and, therefore, it has usually been applied only to input and output layers of the RNN \cite{pham2014dropout}. 
  However, in a recent work, \citet{2015arXiv151205287G} shown that this shortcoming can be circumvented by dropping the same network units in each epoch of the gradient descent. 
  Even if this formulation yields a slightly reduced regularization, nowadays this approach is becoming popular \cite{2016arXiv160703474Z, DBLP:journals/corr/ChePCSL16} and is the one followed in this paper.  
\end{enumerate}

Beside the ones discussed above, several other kinds of regularization procedures have been proposed in the literature. 
Examples are the stochastic noise injection \cite{neelakantan2015adding} and the max-norm constraint \cite{lee2010practical}, which, however, are not considered in our experiments.

\subsection{Parameters update strategies}
\label{sec:strategies}

Rather than evaluating the loss function over the entire training set to perform a single update of the network parameters, a very common approach consists in computing the gradient over mini-batches $\mathcal{X}_k$ of the training data. 
The size of the batch is usually set by following rules of thumb \cite{Bengio2012}. 

This gradient-update method is called Stochastic Gradient Descent (SGD) and, in presence of a non-convex function, its convergence to a local minimum is guaranteed (under some mild assumptions) if the learning rate is sufficiently small \cite{bottou-mlss-2004}.
The update equation reads
\begin{equation}
  \label{eq:sgd}
  \W_{k+1} = \W_{k} + \eta \nabla L_k(\W_{k}),
\end{equation}
where $\eta$ is the \emph{learning rate}, an important hyperparameter that must be carefully tuned to achieve an effective training \cite{bottou2012stochastic}. 
In fact, a large learning rate provides a high amount of kinetic energy in the gradient descent, which causes the parameter vector to bounce, preventing the access to narrow area of the search space, where the loss function is lower. 
On the other hand, a strong decay can excessively slow the training procedure, resulting in a waste of computational time.

Several solutions have been proposed over the years, to improve the convergence to the optimal solution \cite{Bottou2012}.
During the training phase it is usually helpful to anneal $\eta$ over time or when the performance stops increasing. 
A method called \textit{step decay} reduces the learning rate by a factor $\alpha$, if after a given number of epochs the loss has not decreased. 
The \textit{exponential decay} and the \textit{fractional decay} instead, have mathematical forms $\eta=\eta_0 e^{-\alpha k}$ and $\eta=\frac{\eta_0}{(1+\alpha k)}$, respectively. 
Here $\alpha$ and $\eta_0$ are hyperparameters, while $k$ is the current optimization epoch.
In our experiments, we opted for the step decay annealing, when we train the networks with SGD.

Even if SGD usually represents a safe optimization procedure, its rate of convergence is slow and the gradient descent is likely to get stuck in a saddle point of the loss function landscape \cite{NIPS2014_5486}.
Those issues have been addressed by several alternative strategies proposed in the literature for updating the network parameters. 
In the following we describe the most commonly used ones.

\paragraph{Momentum} 
In this first-order method, the weights $\W_k$ are updated according to a linear combination of the current gradient $\nabla L_k(\W_{k})$ and the previous update $\mathbf{V}_{k-1}$, which is scaled by a hyperparameter $\mu$:
\begin{equation}
	\label{eq:momentum}
	\begin{aligned}
	\mathbf{V}_k & = \mu \mathbf{V}_{k-1} - \eta\nabla L_k(\W_{k}), \\
	\W_{k+1} & = \W_{k} + \mathbf{V}_k.
	\end{aligned}
\end{equation}
With this approach, the updates will build up velocity toward a direction that shows a consistent gradient \cite{sutskever2013training}. A common choice is to set $\mu=0.9$.

A variant of the original formulation is the \textit{Nesterov momentum}, which often achieves a better convergence rate, especially for smoother loss functions \cite{nesterov1983method}. 
Contrarily to the original momentum, the gradient is evaluated at an approximated future location, rather than at the current position. The update equations are
\begin{equation}
  \label{eq:nesterovMomentum}
  \begin{aligned}
  \mathbf{V}_k & = \mu \mathbf{V}_{k-1} - \eta\nabla L_k(\W_{k}+\mu\mathbf{V}_{k-1}), \\
  \W_{k+1} & = \W_{k} + \mathbf{V}_k.
  \end{aligned}
\end{equation}

\paragraph{Adaptive learning rate}
The first adaptive learning rate method, proposed by \citet{duchi2011adaptive}, is Adagrad. 
Unlike the previously discussed approaches, Adagrad maintains a different learning rate for each parameter. 
Given the update information from all previous iterations $\nabla L_k\left( \W_{j} \right)$, with $j \in \{0, 1, \cdots, k\}$, a different update is specified for each parameter $i$ of the weight matrix:
\begin{equation}
	\label{eq:adagrad}
	\begin{aligned}
	\W_{k+1}^{(i)} = \W_{k}^{(i)} -\eta \frac{\nabla L_k\left( \W_{k}^{(i)} \right)}{\sqrt{ \sum_{j=0}^{k} \nabla L_k\left( \W_{j}^{(i)} \right)^2} + \epsilon},
	\end{aligned}
\end{equation}
where $\epsilon$ is a small term used to avoid division by $0$.
A major drawback with Adagrad is the unconstrained growth of the accumulated gradients over time. 
This can cause diminishing learning rates that may stop the gradient descent prematurely.

A procedure called RMSprop \cite{tieleman2012lecture} attempts to solve this issue by using an exponential decaying average of square gradients, which discourages an excessive shrinkage of the learning rates:
\begin{equation}
  \label{eq:rmsprop}
  \begin{aligned}
  v_k^{(i)} & =
  \begin{cases}
  (1-\delta) \cdot v_{k-1}^{(i)} + \delta \nabla L_k\left(\W_k^{(i)}\right)^2 & \text{if} \; \nabla L_k \left( \W_{k}^{(i)} \right) > 0\\
  (1-\delta) \cdot v_{k-1}^{(i)} & \text{otherwise}
  \end{cases} \\
  \W_{k+1}^{(i)}  & = \W_k^{(i)} - \eta v_k^{(i)}.
  \end{aligned}
\end{equation}
According to the update formula, if there are oscillation in gradient updates, the learning rate is reduced by $1-\delta$, otherwise it is increased by $\delta$. 
Usually the decay rate is set to $\delta = 0.01$.

Another approach called Adam and proposed by \citet{kingma2014adam}, combines the principles of Adagrad and momentum update strategies.
Usually, Adam is the adaptive learning method that yields better results and, therefore, is the gradient descent strategy most used in practice.
Like RMSprop, Adam stores an exponentially decaying average of gradients squared, but it also keeps an exponentially decaying average of the moments of the gradients. 
The update difference equations of Adam are
\begin{equation}
  \label{eq:adam}
  \begin{aligned}
  m_k &= \beta_1 m_{k-1} + (1-\beta_1) \nabla L_k \left( \W_{k}^{(i)} \right),\\
  v_k &= \beta_2 v_{k-1} + (1-\beta_2) \nabla L_k \left( \W_{k}^{(i)} \right)^2,\\
  \hat{m}_k &= \frac{m_k}{1-\beta_1^k}, \;\; \hat{v}_k = \frac{v_k}{1-\beta_2^k}, \\
  \W_{k+1} &= \W_{k} + \frac{\eta}{\sqrt{\hat{v}_k+\epsilon}} \hat{m}_k \; .
  \end{aligned}
\end{equation}
$m$ corresponds to the first moment and $v$ is the second moment. 
However, since both $m$ and $v$ are initialized as zero-vectors, they are biased towards $0$ during the first epochs. 
To avoid this effect, the two terms are corrected as $\hat{m}_t$ and $\hat{v}_t$.
Default values of the hyperparameters are $\beta_1 = 0.9$, $\beta_2 = 0.999$ and $\varepsilon = 10^{-8}$.

\paragraph{Second-order methods}
The methods discussed so far only consider first-order derivatives of the loss function.
Due to this approximation, the landscape of the loss function locally looks and behaves like a plane. 
Ignoring the curvature of the surface may lead the optimization astray and it could cause the training to progress very slowly.
However, second-order methods involve the computation of the Hessian, which is expensive and usually untreatable even in networks of medium size.
A Hessian-Free (HF) method that considers derivatives of the second order, without explicitly computing the Hessian, has been proposed by \citet{martens2010deep}. 
This latter, unlike other existing HF methods, makes use of the positive semi-definite Gauss-Newton curvature matrix and it introduces a damping factor based on the Levenberg-Marquardt heuristic, which permits to train networks more effectively. 
However, \citet{sutskever2013importance} showed that HF obtains similar performance to SGD with Nesterov momentum.
Despite being a first-order approach, Nestrov momentum is capable of accelerating directions of low-curvature just like a HF method and, therefore, is preferred due to its lower computational complexity.

\subsection{Vanishing and exploding gradient}
\label{sec:vanishing}

Increasing the depth in an RNN, in general, improves the memory capacity of the network and its modeling capabilities \cite{DBLP:journals/corr/PascanuGCB13}. 
For example, stacked RNNs do outperform shallow ones with the same hidden size on problems where it is necessary to store more information throughout the hidden states between the input and output layer \cite{sutskever2014sequence}.
One of the principal drawback of early RNN architectures was their limited memory capacity, caused by the \textit{vanishing} or \textit{exploding gradient} problem \cite{el1995hierarchical}, which becomes evident when the information contained in past inputs must be retrieved after a long time interval \cite{hochreiter2001gradient}.
To illustrate the issue of vanishing gradient, one can consider the influence of the loss function $L_t$ (that depends on the network inputs and on its parameters) on the network parameters $\W_t$, when its gradient is backpropagated through the unfolded 
The network Jacobian reads as
\begin{equation}
  \label{eq:jacobian}
  \frac{\partial L[t]}{\partial W} = \sum_\tau \frac{\partial L[t]}{\partial h[t]}\frac{\partial h[t]}{\partial h[\tau]}\frac{\partial h[\tau]}{\partial W}.
\end{equation}

In the previous equation, the partial derivatives of the states with respect to their previous values can be factorized as
\begin{equation}
	\label{eq:vanishing}
	\frac{\partial h[t]}{\partial h[\tau]} = \frac{\partial h[t]}{\partial h[t-1]} \dots \frac{\partial h[\tau +1]}{\partial h[\tau]} = f_t^{'} \dots f_{\tau+1}^{'}.
\end{equation}

To ensure local stability, the network must operate in a ordered regime \cite{7765110}, a property ensured by the condition $|f_t^{'}| < 1$.
However, in this case the product expanded Eq. \ref{eq:vanishing} rapidly (exponentially) converges to 0, when $t-\tau$ increases.
Consequently, the sum in Eq. \ref{eq:jacobian} becomes dominated by terms corresponding to short-term dependencies and the vanishing gradient effect occurs.
As principal side effect, the weights are less and less updated as the gradient flows backward through the layers of the network.
On the other hand, the phenomenon of exploding gradient appears when $|f_t^{'}| > 1$ and the network becomes locally unstable. 
Even if global stability can still be obtained under certain conditions, in general the network enters into a chaotic regime, where its computational capability is hindered \cite{7817870}.

Models with large recurrent depths exacerbate these gradient-related issues, since they posses more nonlinearities and the gradients are more likely to explode or vanish. 
A common way to handle the exploding gradient problem, is to clip the norm of the gradient if it grows above a certain threshold. 
This procedure relies on the assumption that exploding gradients only occur in contained regions of the parameters space.
Therefore, clipping avoids extreme parameter changes without overturning the general descent direction \cite{pascanu2012understanding}. 

On the other hand, different solutions have been proposed to tackle the vanishing gradient issue.
A simple, yet effective approach consists in initializing the weights to maintain the same variance withing the activations and back-propagated gradients, as one moves along the network depth.
This is obtained with a random initialization that guarantees the variance of the components of the weight matrix in layer $l$ to be $\mathrm{Var}(\W_l)=2/(N_{l-1}+N_{l+1})$, $N_{l-1}$ and $N_{l+1}$ being the number of units in the previous and the next layer respectively \cite{glorot2010understanding}.
\citet{he2015delving} proposed to initialize the network weights by sampling them from an uniform distribution in $[0,1]$ and then rescaling their values by $1/\sqrt{N_h}$, $N_h$ being the total number of hidden neurons in the network.
Another option, popular in deep FFNN, consists in using ReLU \cite{DBLP:conf/icml/NairH10} as activation function, whose derivative is $0$ or $1$, and it does not cause the gradient to vanish or explode. 
Regularization, besides preventing unwanted overfitting in the training phase, proved to be useful in dealing with exploding gradients. 
In particular, $\mathrm{L}_1$ and $\mathrm{L}_2$ regularizations constrain the growth of the components of the weight matrices and consequently limit the values assumed by the propagated gradient \cite{Pascanu_Mikolov_Bengio}.
Another popular solution is adopting gated architectures, like Long Short-Term Memory (LSTM) or Gated Recurrent Unit (GRU), which have been specifically designed to deal with vanishing gradients and allow the network to learn much longer-range dependencies.
\citet{NIPS2015_5850} proposed an architecture called \textit{Highway Network}, which allows information to flow across several layers without attenuation. 
Each layer can smoothly vary its behavior between that of a plain layer, implementing an affine transform followed by a non-linear activation, and that of a layer which simply passes its input through.
Optimization in highway networks is virtually independent of depth, as information can be routed (unchanged) through the layers.
The Highway architecture, initially applied to deep FFNN \cite{2015arXiv151203385H}, has recently been extended to RNN where it dealt with several modeling and optimization issues \cite{2016arXiv160703474Z}.

Finally, gradient-related problems can be avoided by repeatedly selecting new weight parameters using random guess or evolutionary approaches \cite{john1992holland, gomez2003robust}; in this way the network is less likely to get stuck in local minima. 
However, convergence time of these procedures is time-consuming and can be impractical in many real-world applications.
A solution proposed by \citet{schmidhuber2007training}, consists in evolving only the weights of non-linear hidden units, while linear mappings from hidden to output units are tuned using fast algorithms for convex problem optimization.

\section{Recurrent Neural Networks Architectures}
\label{sec:rnn_architectures}

In this section, we present three different RNN architectures trainable through the BPPT procedure, which we employ to predict real-valued time series. 
First, in Sec. \ref{sec:ernn} we present the most basic version of RNN, called Elman RNN. 
In Sec. \ref{sec:LSTM} and \ref{sec:GRU} we discuss two gated architectures, which are LSTM and GRU.
For each RNN model, we provide a quick overview of the main applications in time series forecasting and we discuss its principal features.

\subsection{Elman Recurrent Neural Network}
\label{sec:ernn}

The Elman Recurrent Neural Network (ERNN), also known as \textit{Simple RNN} or \textit{Vanillan RNN}, is depicted in Fig.~\ref{fig:rnn} and is usually considered to be the most basic version of RNN. 
Most of the more complex RNN architectures, such as LSTM and GRU, can be interpreted as a variation or as an extension of ERNNs.

ERNN have been applied in many different contexts. 
In natural language processing applications, ERNN demonstrated to be capable of learning grammar using a training set of unannotated sentences to predict successive words in the sentence \cite{elman1995language, ogata2007two}. 
\citet{Mori1993} studied ERNN performance in short-term load forecasting and proposed a learning method, called ``diffusion learning'' (a sort of momentum-based gradient descent), to avoid local minima during the optimization procedure. 
\citet{Cai2007} trained a ERNN with a hybrid algorithm that combines particle swarm optimization and evolutionary computation to overcome the local minima issues of gradient-based methods. 
Furthermore, ERNNs have been employed by \citet{Cho2003323} in tourist arrival forecasting and by \citet{Mandal2006} to predict electric load time series. 
Due to the critical dependence of electric power usage on the day of the week or month of the year, a preprocessing step is performed to cluster similar days according to their load profile characteristics. 
\citet{Chitsaz2015} proposes a variant of ERNN called Self-Recurrent Wavelet Neural Network, where the ordinary nonlinear activation functions of the hidden layer are replaced with wavelet functions.
This leads to a sparser representation of the load profile, which demonstrated to be helpful for tackling the forecast task through smaller and more easily trainable networks.

The layers in a RNN can be divided in \emph{input layers}, \emph{hidden layers} and the \emph{output layers} (see Fig.~\ref{fig:rnn}).
While input and output layers are characterized by feedforward connections, the hidden layers contain recurrent ones. 
At each time step $t$, the input layer process the component $\mathbf{x}[t] \in \mathbb{R}^{N_i}$ of a serial input $\mathbf{x}$. 
The time series $\mathbf{x}$ has length $T$ and it can contain real values, discrete values, one-hot vectors, and so on. 
In the input layer, each component $\mathbf{x}[t]$ is summed with a bias vector $\b_i \in \mathbb{R}^{N_h}$ ($N_h$ is the number of nodes in the hidden layer) and then is multiplied with the input weight matrix $\W_i^h \in \mathbb{R}^{N_i \times N_h}$.
Analogously, the internal state of the network $\mathbf{h}[t-1] \in \mathbb{R}^{N_h}$ from the previous time interval is first summed with a bias vector $\mathbf{b}_h \in \mathbb{R}^{N_h}$ and then multiplied by the weight matrix $\W_h^h \in \mathbb{R}^{N_h \times N_h}$ of the recurrent connections.
The transformed current input and past network state are then combined and processed by the neurons in the hidden layers, which apply a non-linear transformation.
The difference equations for the update of the internal state and the output of the network at a time step $t$ are:
\begin{equation}
  \label{eq:ernn _state}
  \begin{aligned}
  \h[t] &= f \left( \W_i^h \left( \x[t] + \b_i \right) + \W_h^h \left( \h[t-1] +\b_h \right) \right), \\
  \y[t] &= g \left( \W_h^o \left( \h[t] + \b_o \right) \right),
  \end{aligned}
\end{equation}
where $f(\cdot)$ is the activation function of the neurons, usually implemented by a sigmoid or by a hyperbolic tangent. 
The hidden state $\h[t]$ conveys the content of the memory of the network at time step $t$, is typically initialized with a vector of zeros and it depends on past inputs and network states. 
The output $\y[t] \in \mathbb{R}^{N_o}$ is computed through a transformation $g(\cdot)$, usually linear, on the matrix of the output weights $\W_h^o \in \mathbb{R}^{N_r \times N_o}$ applied to the sum of the current state $\h[t]$ and the bias vector $\b_o \in \mathbb{R}^{N_o}$.
All the weight matrices and biases can be trained through gradient descent, according to the BPPT procedure.
Unless differently specified, in the following to compact the notation we omit the bias terms by assuming $\mathbf{x} = [\mathbf{x}; 1]$, $\mathbf{h} = [\mathbf{h}; 1]$, $\mathbf{y} = [\mathbf{y}; 1]$ and by augmenting $\W_i^h$, $\W_h^h$, $\W_h^o$ with an additional column.

\subsection{Long Short-Term Memory}
\label{sec:LSTM}

The Long Short-Term Memory (LSTM) architecture was originally proposed by \citet{hochreiter1997long} and is widely used nowadays due to its superior performance in accurately modeling both short and long term dependencies in data.
LSTM tries to solve the vanishing gradient problem by not imposing any bias towards recent observations, but it keeps constant error flowing back through time.
LSTM works essentially in the same way as the ERNN architecture, with the difference that it implements a more elaborated internal processing unit called \textit{cell}.

LSTM has been employed in numerous sequence learning applications, especially in the field of natural language processing. 
Outstanding results with LSTM have been reached by \citet{graves2009offline} in unsegmented connected handwriting recognition, by \citet{graves2013speech} in automatic speech recognition, by \citet{eck2002finding} in music composition and by \citet{gers2001lstm} in grammar learning. 
Further successful results have been achieved in the context of image tagging, where LSTM have been paired with convolutional neural network, to provide annotations on images automatically \cite{Vinyals_2015_CVPR}.

However, few works exist where LSTM has been applied to prediction of real-valued time series.
\citet{Ma2015} evaluated the performances of several kinds of RNNs in short-term traffic speed prediction and compared them with other common methods like SVMs, ARIMA, and Kalman filters, finding that LSTM networks are nearly always the best approach. 
\citet{Pawlowski2015} utilized ensembles of LSTM and feedforward architectures to classify the danger from concentration level of methane in a coal mine, by predicting future concentration values. 
By following a hybrid approach, \citet{Felder2010} trains a LSTM network to output the parameter of a Gaussian mixture model that best fits a wind power temporal profile. 

While an ERNN neuron implements a single nonlinearity $f(\cdot)$ (see Eq. \ref{eq:ernn _state}), a LSTM cell is composed of 5 different nonlinear components, interacting with each other in a particular way.
The internal state of a cell is modified by the LSTM only through linear interactions.
This permits information to backpropagate smoothly across time, with a consequent enhancement of the memory capacity of the cell.
LSTM protects and controls the information in the cell through three gates, which are implemented by a sigmoid and a pointwise multiplication.
To control the behavior of each gate, a set of parameters are trained with gradient descent, in order to solve a target task.

\begin{SCfigure}[2.5][t]
  \centering
  \includegraphics[scale=0.7, keepaspectratio]{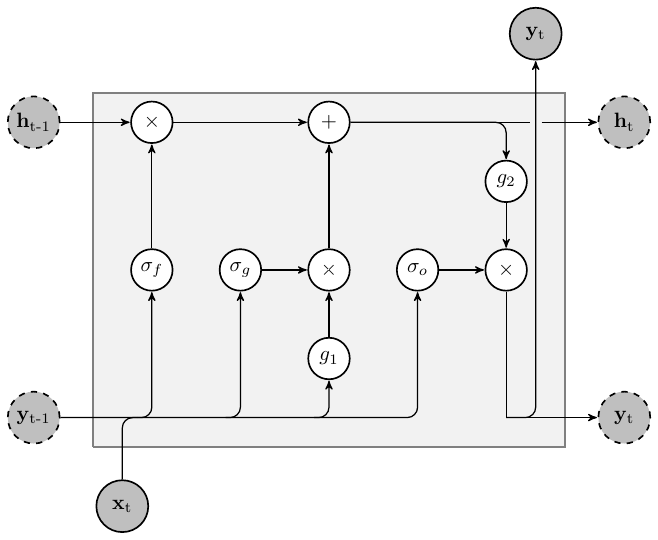}
  \caption{Illustration of a cell in the LSTM architecture. 
  Dark gray circles with a solid line are the variables whose content is exchanged with the input and output of the cell.
  Dark gray circles with a dashed line represent the internal state variables, whose content is exchanged between the cells of the hidden layer. 
  Operators $g_1$ and $g_2$ are the non-linear transformation, usually implemented as a hyperbolic tangent. 
  White circles with $+$ and $\times$ represent linear operations, while $\sigma_f$, $\sigma_u$ and $\sigma_o$ are the sigmoids used in the forget, update and output gates respectively.}
  \label{fig:lstm}
\end{SCfigure}

Since its initial definition \cite{hochreiter1997long}, several variants of the original LSTM unit have been proposed in the literature.
In the following, we refer to the commonly used architecture proposed by \citet{Graves2005602}. 
A schema of the LSTM cell is depicted in Fig.~\ref{fig:lstm}.
%

The difference equations that define the forward pass to update the cell state and to compute the output are listed below. 
\begin{equation}
  \label{eq:lstm_equations}
  \begin{aligned}
  \text{forget gate}:     \;\; & \sigma_f[t]           = \sigma\left( \W_f \x[t] + \R_f \mathbf{y}[t-1] + \b_f \right), \\
  \text{candidate state}: \;\; & \mathbf{\tilde{h}}[t] = g_1\left(      \W_h \x[t] + \R_h \mathbf{y}[t-1] + \b_h \right), \\
  \text{update gate}:     \;\; & \sigma_u[t]           = \sigma\left( \W_u \x[t] + \R_u \mathbf{y}[t-1] + \b_u \right), \\
  \text{cell state}:      \;\; & \mathbf{h}[t]         = \sigma_u[t]  \odot \mathbf{\tilde{h}}[t] + \sigma_f[t] \odot \mathbf{h}[t-1], \\
  \text{output gate}:     \;\; & \sigma_o[t]           = \sigma\left( \W_o \x[t] + \R_o \mathbf{y}[t-1] + \b_o \right), \\
  \text{output}:          \;\; & \mathbf{y}[t]         = \sigma_o[t] \odot g_2(\mathbf{h}[t]).
  \end{aligned}
\end{equation}
$\x[t]$ is the input vector at time $t$.
$\W_f$, $\W_h$, $\W_u$, and $\W_o$ are rectangular weight matrices, that are applied to the input of the LSTM cell. 
$\R_f$, $\R_h$, $\R_u$, and $\R_o$ are square matrices that define the weights of the recurrent connections, while $\b_f$, $\b_h$, $\b_u$, and $\b_o$ are bias vectors.
The function $\sigma(\cdot)$ is a sigmoid \footnote{the logistic sigmoid is defined as $\sigma(x) = \frac{1}{1+e^{-x}}$ }, while $g_1(\cdot)$ and $g_2(\cdot)$ are pointwise non-linear activation functions, usually implemented as hyperbolic tangents that squash the values in $[-1,1]$.
Finally, $\odot$ is the entrywise multiplication between two vectors (Hadamard product).

Each gate in the cell has a specific and unique functionality.
The \textit{forget gate} $\sigma_f$ decides what information should be discarded from the previous cell state $\mathbf{h}[t-1]$.
The \textit{input gate} $\sigma_u$ operates on the previous state $\mathbf{h}[t-1]$, after having been modified by the forget gate, and it decides how much the new state $\mathbf{h}[t]$ should be updated with a new candidate $\mathbf{\tilde{h}}[t]$.
To produce the output $\mathbf{y}[t]$, first the cell filters its current state with a nonlinearity $g_2(\cdot)$.
Then, the \textit{output gate} $\sigma_o$ selects the part of the state to be returned as output.
Each gate depends on the current external input $\x[t]$ and the previous cells output $\mathbf{y}[t-1]$.

As we can see from the Fig.~\ref{fig:lstm} and from the forward-step equations, when $\sigma_f=\mathbf{1}$ and $\sigma_u=\mathbf{0}$, the current state of a cell is transferred to the next time interval exactly as it is.
By referring back to Eq. \ref{eq:vanishing}, it is possible to observe that in LSTM the issue of vanishing gradient does not occur, due to the absence of nonlinear transfer functions applied to the cell state.
Since in this case the transfer function $f(\cdot)$ in Eq. \ref{eq:vanishing} applied to the internal states is an identity function, the contribution from past states remains unchanged over time.
However, in practice, the update and forget gates are never completely open or closed due to the functional form of the sigmoid, which saturates only for infinitely large values.
As a result, even if long term memory in LSTM is greatly enhanced with respect to ERNN architectures, the content of the cell cannot be kept completely unchanged over time.

\subsection{Gated Recurrent Unit}
\label{sec:GRU}

The Gated Recurrent Unit (GRU) is another notorious gated architecture, originally proposed by \citet{cho2014learning}, which adaptively captures dependencies at different time scales. 
In GRU, forget and input gates are combined into a single update gate, which adaptively controls how much each hidden unit can remember or forget.
The internal state in GRU is always fully exposed in output, due to the lack of a control mechanism, like the output gate in LSTM.

GRU were firstly tested by \citet{cho2014learning} on a statistical machine translation task and reported mixed results.
In an empirical comparison of GRU and LSTM, configured with the same amount of parameters, \citet{DBLP:journals/corr/ChungGCB14} concluded that on some datasets GRU can outperform LSTM, both in terms of generalization capabilities and in terms of time required to reach convergence and to update parameters.
In an extended experimental evaluation, \citet{zaremba2015empirical} employed GRU to (i) compute the digits of the sum or difference of two input numbers, (ii) predict the next character in a synthetic XML dataset and in the large words dataset Penn TreeBank, (iii) predict polyphonic music. The results showed that the GRU outperformed the LSTM on nearly all tasks except language modeling when using a naive initialization. 
\citet{bianchi2017temporal} compared GRU with other recurrent networks on the prediction of superimposed oscillators.
However, to the best of author's knowledge, at the moment there are no researches where the standard GRU architecture has been applied in STLF problems.

\begin{SCfigure}[2.5][t]
  \centering
  \includegraphics[scale=0.7, keepaspectratio]{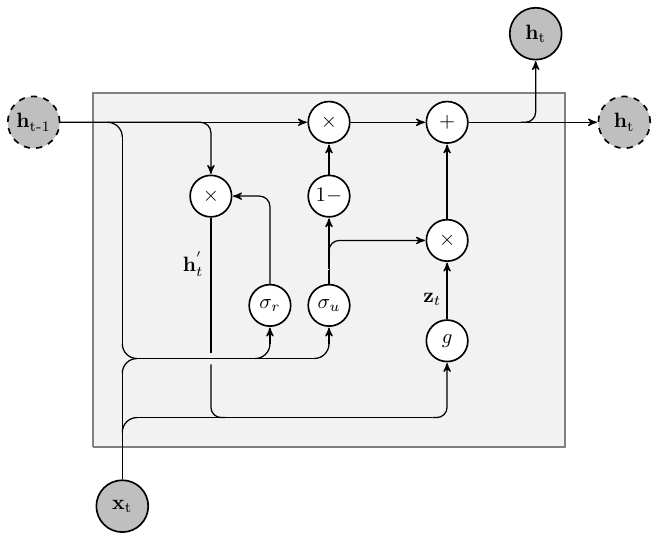}
  \caption{Illustration of a recurrent unit in the GRU architecture. 
	   Dark gray circles with a solid line are the variables whose content is exchanged with the input and output of the network. 
	   Dark gray circles with a dashed line represent the internal state variables, whose content is exchanged within the cells of the hidden layer. 
	   The operator $g$ is a non-linear transformation, usually implemented as a hyperbolic tangent. 
	   White circles with '$+$', '$-1$' and '$\times$' represent linear operations, while $\sigma_r$ and $\sigma_u$ are the sigmoids used in the reset and update gates respectively.}
  \label{fig:gru}
\end{SCfigure}

A schematic depiction of the GRU cell is reported in Fig.~\ref{fig:gru}.
%
%
GRU makes use of two gates.
The first is the \textit{update gate}, which controls how much the current content of the cell should be updated with the new candidate state.
The second is the \textit{reset gate} that, if closed (value near to 0), can effectively reset the memory of the cell and make the unit act as if the next processed input was the first in the sequence. 
The state equations of the GRU are the following:
\begin{equation}
\begin{aligned}
\text{reset gate}:	\;\; & \mathbf{r}[t] = \sigma \left( \W_{r} \h[t-1] + \R_{r} \x[t] +\b_r \right), \\
\text{current state}:	\;\; & \h'[t] = \h[t-1] \odot \mathbf{r}[t], \\
\text{candidate state}:	\;\; & \mathbf{z}[t] = g\left(\W_{z}\h'[t] + \R_{z} \x[t] + \b_z \right), \\
\text{update gate}:	\;\; & \mathbf{u}[t] = \sigma \left(\W_u \h[t-1] + \R_u \x[t] + \b_u \right), \\
\text{new state}:	\;\; & \h[t] = (1-\mathbf{u}[t]) \odot \h[t-1] + \mathbf{u}[t] \odot \mathbf{z}[t].
\end{aligned}
\end{equation}
Here, $g(\cdot)$ is a non-linear function usually implemented by a hyperbolic tangent.

In a GRU cell, the number of parameters is larger than in the an ERNN unit, but smaller than in a LSTM cell. 
The parameters to be learned are the rectangular matrices $\W_r$, $\W_z$, $\W_u$, the square matrices $\R_r$, $\R_z$, $\R_u$, and the bias vectors $\b_r$, $\b_z$, $\b_u$.

\section{Other Recurrent Neural Networks Models}
\label{sec:other_architectures}

In this section we describe two different types of RNNs, which are the Nonlinear AutoRegressive eXogenous inputs neural network (NARX) and the Echo State Network (ESN).
Both of them have been largely employed in STLF.
These two RNNs differ from the models described in Sec. \ref{sec:rnn_architectures}, both in terms of their architecture and in the training procedure, which is not implemented as a BPPT.
Therefore, some of the properties and training approaches discussed in Sec. \ref{sec:rnn} do not hold for these models.

\subsection{NARX Network}
\label{sec:narx}

NARX networks are recurrent dynamic architectures with several hidden layers and they are inspired by discrete-time nonlinear models called Nonlinear AutoRegressive with eXogenous inputs \cite{leontaritis1985input}.
Differently from other RNNs, the recurrence in the NARX network is given only by the feedback on the output, rather than from the whole internal state. 

NARX networks have been employed in many different applicative contexts, to forecast future values of the input signal \cite{diaconescu2008use, lin1997delay}. 
\citet{menezes2008} showed that NARX networks perform better on predictions involving long-term dependencies. 
\citet{5212326} used NARX in conjunction with an input embedded according to Takens method, to predict highly non-linear time series.
NARX are also employed as a nonlinear filter, whose target output is trained by using the noise-free version of the input signal \cite{napoli2010nonlinear}. 
NARX networks have also been adopted by \citet{plett2003adaptive} in a gray-box approach for nonlinear system identification.

A NARX network can be implemented with a MultiLayer Perceptron (MLP), where the next value of the output signal $\y[t] \in \mathbb{R}^{N_y}$ is regressed on $d_y$ previous values of the output signal and on $d_x$ previous values of an independent, exogenous input signal $\x[t]\in \mathbb{R}^{N_x}$ \cite{billings2013nonlinear}. 
The output equation reads
\begin{equation}
  \label{eq:narx_output}
  \y[t] = \phi\left(\x[t-d_x],\ldots,\x[t-1],\x[t],\y[t-d_y],\ldots,\y[t-1], \Theta\right),
\end{equation}
where $\phi(\cdot)$ is the nonlinear mapping function performed by the MLP, $\Theta$ are the trainable network parameters, $d_x$ and $d_y$ are the input and the output time delays. 
Even if the numbers of delays $d_x$ and $d_y$ is a finite (often small) number, it has been proven that NARX networks are at least as powerful as Turing machines, and thus they are universal computation devices \cite{siegelmann1997computational}.

The input $\mathbf{i}[t]$ of the NARX network has $d_xN_x+d_yN_y$ components, which correspond to a set of two Tapped-Delay Lines (TDLs), and it reads
\begin{equation}
  \label{eq:narx_input}
  \mathbf{i}[t] = 
  \left[\begin{array}{c}
  \left( \x[t-d_x], \ldots, \x[t-1] \right)^T\\
  \left( \y[t-d_y], \ldots, \y[t-1] \right)^T
  \end{array}\right]^T.
\end{equation}
  
The structure of a MLP network consists of a set of source nodes forming the input layer, $N_l \geq 1$ layers of hidden nodes, and an output layer of nodes. 
The output of the network is governed by the following difference equations
\begin{align}
  \label{eq:narx_output2}
  \mathbf{h}_1[t] &= f \left( \mathbf{i}[t], \theta_i \right), \\
  \mathbf{h}_{l}[t] &= f \left( \mathbf{h}_{l-1}[t-1], \theta_{h_l} \right), \\
  \mathbf{y}[t] &= g \left( \mathbf{h}_{N_l}[t-1], \theta_{o} \right),
\end{align}
where $\mathbf{h}_{l}[t] \in \mathbb{R}^{N_{h_l}}$ is the output of the $l$\textsuperscript{th} hidden layer at time $t$, $g(\cdot)$ is a linear function and $f(\cdot)$ is the transfer function of the neuron, usually implemented as a sigmoid or \textit{tanh} function. 

\begin{figure}[htp!]
    \centering
    \includegraphics[scale=1, keepaspectratio]{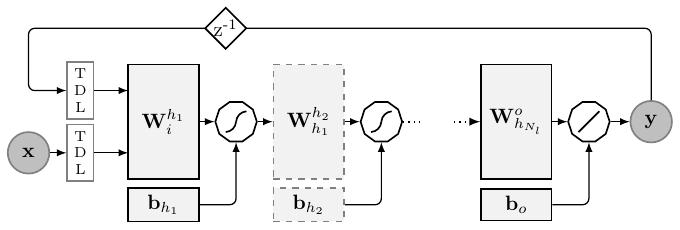}
    \caption{Architecture of the NARX network. 
    Circles represent input $\mathbf{x}$ and output $\mathbf{y}$, respectively. 
    The two TDL blocks are the tapped-delay lines. 
    The solid squares $\W_{i}^{h_1}$, $\W_{h_{N_l}}^{o}$, $\b_i$, and $\b_o$ are the weight matrices and the bias relative to the input and the output respectively. 
    The dashed squares are the weight matrices and the biases relative to the $N_l$ hidden layers -- in the figure, we report $\W_{h_1}^{h_2}$ and $\b_{h_2}$, relative to the first hidden layer.
    The polygon with the sigmoid symbol represents the nonlinear transfer function of the neurons and the one with the oblique line is a linear operation. 
    Finally, $\text{z}^{\text{-1}}$ is the backshift/lag operator.}
    \label{fig:narx}
\end{figure}

The weights of the neurons connections are defined by the parameters $\Theta = \left\{ \theta_i, \theta_o, \theta_{h_1}, \dots, \theta_{h_{N_l}} \right\}$. 
In particular, $\theta_i = \left\{ \W_i^{h_1} \in \mathbb{R}^{d_xN_x+d_yN_y \times N_{h_1}}, \b_{h_1} \in \mathbb{R}^{N_{h_1}} \right\}$ are the parameters that determine the weights in the input layer, $\theta_o = \left\{ \W_{h_{N_l}}^{o} \in \mathbb{R}^{N_{N_l} \times N_y }, \b_o \in \mathbb{R}^{N_y} \right\}$ are the parameters of the output layer and $\theta_{h_l} = \left\{ \W_{h_{l-1}}^{h_l} \in \mathbb{R}^{N_{h_{l-1}} \times N_{h_l}}, \b_{h_l} \in \mathbb{R}^{N_{h_l}} \right\}$ are the parameters of the $l$\textsuperscript{th} hidden layer.
A schematic depiction of a NARX network is reported in Fig.~\ref{fig:narx}.

\begin{figure}[htp!]
\centering
    \subfigure[Training mode]{
    \includegraphics[width=0.4\columnwidth]{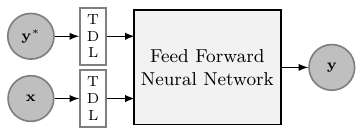}
    \label{fig:narx_train1}}
    ~
    \subfigure[Operational mode]{
    \includegraphics[width=0.4\columnwidth]{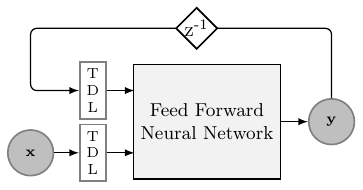}
    \label{fig:narx_train2}}
\caption{During the training, the desired input $\y^*$ is fed directly to the network. Once the network parameters have been optimized, the teacher signal is removed and the output $\y$ produced by the network is connected to the input with a feedback loop.}
\label{fig:narx_train}
\end{figure}

Due to the architecture of the network, it is possible to exploit a particular strategy to learn the parameters $\Theta$.
Specifically, during the training phase the time series relative to the desired output $\y^*$ is fed into the network along with the input time series $\x$. 
At this stage, the output feedback is disconnected and the network has a purely feed-forward architecture, whose parameters can be trained with one of the several, well-established standard backpropagation techniques. Notice that this operation is not possible in other recurrent networks such as ERNN, since the state of the hidden layer depends on the previous hidden state, whose ideal value is not retrievable from the training set. 
Once the training stage is over, the teacher signal of the desired output is disconnected and is replaced with the feedback of the predicted output $\y$ computed by the network. 
The procedure is depicted in Fig.~\ref{fig:narx_train}.

Similar to what discussed in Sec. \ref{sec:grad_descent} for the previous RNN architectures, the loss function employed in the gradient descent is defined as 
\begin{equation}
  \label{eq:narx_loss}
  L(\x,\y^*;\Theta) = \mathrm{MSE}(\y,\y^*) + \lambda_2 \| \Theta \|_2,
\end{equation}
where $\mathrm{MSE}$ is the error term defined in Eq. \ref{eq:MSE} and $\lambda_2$ is the hyperparameter that weights the importance of the $\mathrm{L}_2$ regularization term in the loss function.
Due to the initial transient phase of the network, when the estimated output $\y$ is initially fed back as network input, the first initial outputs are discarded.

Even if it reduces to a feed-forward network in the training phase, NARX network is not immune to the problem of vanishing and exploding gradients. 
This can be seen by looking at the Jacobian $\mathbf{J}_{\mathbf{h}}(t,n)$ of the state-space map at time $t$ expanded for $n$ time step.
In order to guarantee network stability, the Jacobian must have all of its eigenvalues inside the unit circle at each time step. However, this results in $\lim \limits_{n \rightarrow \infty} \mathbf{J}_{\mathbf{h}}(t,n) = 0$, which implies that NARX networks suffer from vanishing gradients, like the other RNNs \cite{lin1996learning}.

\subsection{Echo State Network}
\label{sec:esn}

While most hard computing approaches and ANNs demand long training procedures to tune the parameters through an optimization algorithm \cite{huang2005particle}, recently proposed architectures such as Extreme Learning Machines \cite{cambria2013extreme,7000606} and ESNs are characterized by a very fast learning procedure, which usually consists in solving a convex optimization problem. 
ESNs, along with Liquid State Machines \cite{maass2002real}, belong to the class of computational dynamical systems implemented according to the so-called \emph{reservoir computing} framework \cite{lukovsevivcius2009reservoir}. 

ESN have been applied in a variety of different contexts, such as static classification \cite{alexandre2009benchmarking}, speech recognition \cite{skowronski2007automatic}, intrusion detection \cite{1504645}, adaptive control \cite{6480841}, detrending of nonstationary time series \cite{desn}, harmonic distortion measurements \cite{4712533} and, in general, for modeling of various kinds of non-linear dynamical systems \cite{han2014fuzzy}.

ESNs have been extensively employed to forecast real valued time series. 
\citet{niu2012multi} trained an ESN to perform multivariate time series prediction by applying a Bayesian regularization technique to the reservoir and by pruning redundant connections from the reservoir to avoid overfitting.
Superior prediction capabilities have been achieved by projecting the high-dimensional output of the ESN recurrent layer into a suitable subspace of reduced dimension \cite{Lokse2017}. 
An important context of application with real valued time series is the prediction of telephonic or electricity load, usually performed 1-hour and a 24-hours ahead \cite{deihimi2012application, deihimi2013short, 7286732, varshneyhalf, bianchi2015prediction}. 
\citet{deihimi2013short} and \citet{peng2014novel} decomposed the time series in wavelet components, which are predicted separately using distinct ESN and ARIMA models, whose outputs are combined to produce the final result.
Important results have been achieved in the prediction of chaotic time series by \citet{li2012chaotic}. 
They proposed an alternative to the Bayesian regression for estimating the regularization parameter and a Laplacian likelihood function, more robust to noise and outliers than a Gaussian likelihood. 
\citet{jaeger2004harnessing} applied an ESN-based predictor on both benchmark and real dataset, highlighting the capability of these networks to learn amazingly accurate models to forecast a chaotic process from almost noise-free training data.

\begin{SCfigure}[1.2][!ht]
    \centering
    \includegraphics[scale=0.9, keepaspectratio]{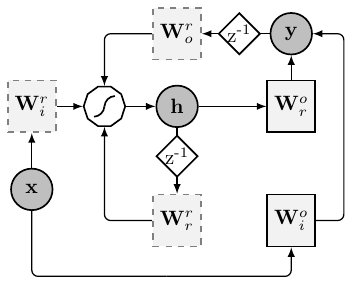}
    \caption{Schematic depiction of the ESN architecture. The circles represent input $\mathbf{x}$, state $\mathbf{h}$, and output $\mathbf{y}$, respectively. Solid squares $\mathbf{W}_{r}^{o}$ and $\mathbf{W}_{i}^{o}$, are the trainable matrices of the readout, while dashed squares, $\mathbf{W}_{r}^{r}$, $\mathbf{W}_{o}^{r}$, and $\mathbf{W}_{i}^{r}$, are randomly initialized matrices. The polygon represents the non-linear transformation performed by neurons and $\text{z}^{\text{-1}}$ is the unit delay operator.}
    \label{fig:esn}
\end{SCfigure}

An ESN consists of a large, sparsely connected, untrained recurrent layer of nonlinear units and a linear, memory-less read-out layer, which is trained according to the task that the ESN is demanded to solve. 
A visual representation of an ESN is shown in Fig.~\ref{fig:esn}
%

The difference equations describing the ESN state-update and output are, respectively, defined as follows:
\begin{align}
  \label{eq:esn_state_update}
  \mathbf{h}[t] =& f(\mathbf{W}_{r}^{r}\mathbf{h}[t-1] + \mathbf{W}_{i}^{r}\mathbf{x}[t] + \mathbf{W}_{o}^{r}\mathbf{y}[t-1] + \epsilon),\\
  \label{eq:esn_esn_output}
  \mathbf{y}[t] =& g(\mathbf{W}_{i}^{o}\mathbf{x}[t] + \mathbf{W}_{r}^{o}\mathbf{h}[t]),
\end{align}
where $\epsilon$ is a small noise term. 
The reservoir contains $N_h$ neurons whose transfer/activation function $f(\cdot)$ is typically implemented by a hyperbolic tangent.
The readout instead, is implemented usually by a linear function $g(\cdot)$.
At time instant $t$, the network is driven by the input signal $\mathbf{x}[t]\in \mathbb{R}^{N_i}$ and produces the output $\mathbf{y}[k] \in \mathbb{R}^{N_o}$, $N_i$ and $N_o$ being the dimensionality of input and output, respectively.
The vector $\mathbf{h}[t]$ has $N_h$ components and it describes the ESN (instantaneous) state.
The weight matrices $\mathbf{W}_r^r \in \mathbb{R}^{N_r \times N_r}$ (reservoir connections), $\mathbf{W}_i^r \in \mathbb{R}^{N_i \times N_r}$ (input-to-reservoir), and $\mathbf{W}_o^r \in \mathbb{R}^{N_o \times N_r}$ (output-to-reservoir feedback) contain real values in the $[-1, 1]$ interval drawn from a uniform distribution and are left untrained. 
Alternative options have been explored recently by \citet{rodan2011minimum} and \citet{appeltant2011information} to generate the connection weights.
The sparsity of the reservoir is controlled by a hyperparameter $R_c$, which determines the number of nonzero elements in $\mathbf{W}_r^r$.
According to the ESN theory, the reservoir $\mathbf{W}_r^r$ must satisfy the so-called ``echo state property'' (ESP) \cite{lukovsevivcius2009reservoir}. 
This means that the effect of a given input on the state of the reservoir must vanish in a finite number of time-instants. 
A widely used rule-of-thumb to obtain this property suggests to rescale the matrix $\mathbf{W}_r^r$ in order to have $\rho(\mathbf{W}_r^r) < 1$, where $\rho(\cdot)$ denotes the spectral radius.
However, several theoretical approaches have been proposed in the literature to tune $\rho$ more accurately, depending on the problem at hand \cite{boedecker2012information, 7765110, verstraeten2009quantification, rnnmultiplex}.

On the other hand, the weight matrices $\mathbf{W}_{i}^{o}$ and $\mathbf{W}_{r}^{o}$ are optimized for the target task. 
To determine them, let us consider the training sequence of $T_{\text{tr}}$ desired input-outputs pairs given by:
\begin{equation}
  (\mathbf{x}[1], y^*[1]) \ldots, (\mathbf{x}[T_{\text{tr}}], y[T_{\text{tr}}]),
\end{equation}
where $T_{\text{tr}}$ is the length of the training sequence.
In the initial phase of training, called \emph{state harvesting}, the inputs are fed to the reservoir, producing a sequence of internal states $\mathbf{h}[1], \ldots, \mathbf{h}[T_{\text{tr}}]$, as defined in Eq. \eqref{eq:esn_state_update}. 
The states are stacked in a matrix $\mathbf{S} \in \mathbb{R}^{T_{\text{tr}} \times N_i + N_r}$ and the desired outputs in a vector $\mathbf{y}^* \in \mathbb{R}^{T_{\text{tr}}}$:
\begin{align}
  \mathbf{S} = & 
    \left[\begin{array}{c}
    \mathbf{x}^T[1], \,\, \mathbf{h}^T[1] \\
    \vdots \\
    \mathbf{x}^T[T_{\text{tr}}], \,\, \mathbf{h}^T[T_{\text{tr}}]
    \end{array}\right] \label{eq:esn_state_matrix} \,,\\
  \mathbf{y}^* =  &
    \left[\begin{array}{c}
    \y^*[1] \\
    \vdots \\
    \y^*[T_{\text{tr}}]
    \end{array}\right] \,.\label{eq:esn_output_vector}
\end{align}
The initial rows in $\mathbf{S}$ (and $\mathbf{y}^*$) are discarded, since they refer to a transient phase in the ESN's behavior.

The training of the readout consists in learning the weights in $\mathbf{W}_{i}^{o}$ and $\mathbf{W}_{r}^{o}$ so that the output of the ESN matches the desired output $\mathbf{y}^*$. 
This procedure is termed \emph{teacher forcing} and can be accomplished by solving a convex optimization problem, for which several closed form solution exist in the literature. 
The standard approach, originally proposed by \citet{jaeger2001echo}, consists in applying a least-square regression, defined by the following regularized least-square problem:
\begin{equation}
\boldsymbol{W}_{\text{ls}}^* = \argmin_{\boldsymbol{W} \in \mathbb{R}^{N_i+N_h}}  \frac{1}{2} \| \mathbf{S}\boldsymbol{W} - \mathbf{y}^* \|^2 + \frac{\lambda_2}{2} \|\boldsymbol{W}\|_2^2 \,,
\label{eq:esn_opt}
\end{equation}
where $ \boldsymbol{W} = \left[ \mathbf{W}_i^o \,, \mathbf{W}_r^o \right]^T$ and $\lambda_2 \in \mathbb{R}^+$ is the $\mathrm{L}_2$ regularization factor.

A solution to problem \eqref{eq:esn_opt} can be expressed in closed form as
\begin{equation}
\boldsymbol{W}_{\text{ls}}^* = \left( \mathbf{S}^T \mathbf{S} + \lambda_2 \mathbf{I} \right)^{-1}\mathbf{S}^T \mathbf{y}^* \,,
\label{eq:ridge_regression_minimizer}
\end{equation} 
which can be solved by computing the Moore-Penrose pseudo-inverse.
Whenever $N_h + N_i > T_{\text{tr}}$, Eq. \eqref{eq:ridge_regression_minimizer} can be computed more efficiently by rewriting
it as
\begin{equation}
\boldsymbol{W}_{\text{ls}}^* = \mathbf{S}^T\left( \mathbf{S}\mathbf{S}^T + \lambda_2 \mathbf{I} \right)^{-1} \mathbf{y}^* \,.
\label{eq:ridge_regression_minimizer2}
\end{equation}

\section{Synthetic time series}
\label{sec:synth_data}

We consider three different synthetically generated time series in order to provide controlled and easily replicable benchmarks for the architectures under analysis. 
The three forecasting exercises that we study have a different level of difficulty, given by the nature of the signal and the complexity of the task to be solved by the RNN. 
In order to obtain a prediction problem that is not too simple, it is reasonable to select as forecast horizon a time interval $t_f$ that guarantees the measurements in the time series to become decorrelated. 
Hence, we consider the first zero of the autocorrelation function of the time series.
Alternatively, the first minimum of the average mutual information \cite{fraser1986independent} or of the correlation sum \cite{liebert1989proper} could be chosen to select a $t_f$ where the signal shows a more-general form of independence.
All the time series introduced in the following consist of $15.000$ time steps. We use the first 60\% of the time series as training set, to learn the parameters of the RNN models. 
The next 20\% of the data are used as validation set and the prediction accuracy achieved by the RNNs on this second dataset is used to tune the hyperparameters of the models. 
The final model performance is evaluated on a test set, corresponding to the last 20\% of the values in the time series.

\paragraph{Mackey-Glass time series} 
The Mackey-Glass (MG) system is commonly used as benchmark for prediction of chaotic time series. The input signal is generated from the MG time-delay differential system, described by the following equation:
\begin{equation}
  \label{eq:MG_signal}
  \frac{dx}{dt} = \frac{\alpha x(t-\tau_{\mathrm{MG}})}{1+ x(t-\tau_{\mathrm{MG}})^{10}} - \beta x(t).
\end{equation}
For this prediction task, we set $\tau_{\mathrm{MG}} = 17, \alpha = 0.2, \beta = 0.1$, initial condition $x(0)=1.2$, 0.1 as integration step for (\ref{eq:MG_signal}) and the forecast horizon $t_f = 12$.  

\paragraph{NARMA signal} 
The Non-Linear Auto-Regressive Moving Average (NARMA) task, originally proposed by \citet{jaeger2002adaptive}, consists in modeling the output of the following $r$-order system:
\begin{equation}
  \label{eq:narma_signal}
  y(t + 1) = 0.3y(t) + 0.05y(t)\left[\sum \limits_{i=0}^{r} y(t - i)\right] + 1.5x(t - r)x(t) + 0.1. 
\end{equation}
The input to the system $x(t)$ is uniform random noise in [0, 1], and the model is trained to reproduce $y(t +1)$.
The NARMA task is known to require a memory of at least $r$ past time-steps, since the output is determined by input and outputs from the last $r$ time-steps. 
For this prediction task we set $r=10$ and the forecast step $t_f = 1$ in our experiments.

\paragraph{Multiple superimposed oscillator}
The prediction of a sinusoidal signal is a relatively simple task, which demands a minimum amount of memory to determine the next network output. 
However, superimposed sine waves with incommensurable frequencies are extremely difficult to predict, since the periodicity of the resulting signal is extremely long.
The time series we consider is the Multiple Superimposed Oscillator (MSO) introduced by \citet{jaeger2004harnessing}, and it is defined as
\begin{equation}
  \label{eq:mso_signal}
  y(t) = sin(0.2 t) + sin(0.311 t) + sin(0.42 t) + sin(0.51 t).
\end{equation}
This academic, yet important task, is particularly useful to test the memory capacity of a recurrent neural network and has been studied in detail by \citet{Xue2007365} in a dedicated work.
Indeed, to accurately predict the unseen values of the time series, the network requires a large amount of memory to simultaneously implement multiple decoupled internal dynamics \cite{wierstra2005modeling}. 
For this last prediction task, we chose a forecast step $t_f = 10$.

\section{Real-world load time series}
\label{sec:real_data}

In this section, we present three different real-world dataset, where the time series to be predicted contain measurements of electricity and telephonic activity load. 
Two of the dataset contain exogenous variables, which are used to provide additional context information to support the prediction task.
For each dataset, we perform a pre-analysis to study the nature of the time series and to find the most suitable data preprocessing.
In fact, forecast accuracy in several prediction models, among which neural networks, can be considerably improved by applying a meaningful preprocessing \cite{Zhang2005501}.

\subsection{Orange dataset -- telephonic activity load}
\label{sec:orange}

The first real-world dataset that we analyze is relative to the load of phone calls registered over a mobile network.
Data come from the Orange telephone dataset \cite{D4Dwebsite}, published in the Data for Development (D4D) challenge \cite{DBLP:journals/corr/abs-1210-0137}.
D4D is a collection of call data records, containing anonymized events of Orange's mobile phone users in Ivory Coast, in a period spanning from December 1, 2011 to April 28, 2012.
More detailed information on the data are available in Ref. \cite{Bianchi201649}. 
The time series we consider are relative to antenna-to-antenna traffic. 
In particular, we selected a specific antenna, retrieved all the records in the dataset relative to the telephone activity issued each hour in the area covered by the antenna and generated 6 time series:
\begin{itemize}
  \item \texttt{ts1}: number of incoming calls in the area covered by the antenna;
  \item \texttt{ts2}: volume in minutes of the incoming calls in the area covered by the antenna;
  \item \texttt{ts3}: number of outgoing calls in the area covered by the antenna;
  \item \texttt{ts4}: volume in minutes of the outgoing calls in the area covered by the antenna;
  \item \texttt{ts5}: hour when the telephonic activity is registered;
  \item \texttt{ts6}: day when the telephonic activity is registered.
\end{itemize}

\begin{figure}[!ht]
  \centering
  \subfigure[Load profile]{
  \includegraphics[width=0.45\textwidth,keepaspectratio]{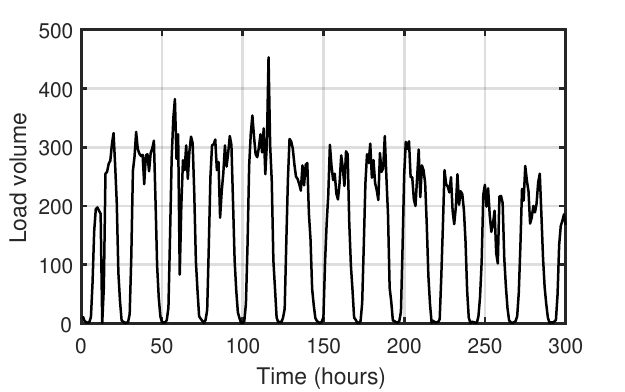}
  \label{fig:callVolume}}\hspace{0em}%
  ~
  \subfigure[Autocorrelation functions]{
  \includegraphics[width=0.45\textwidth]{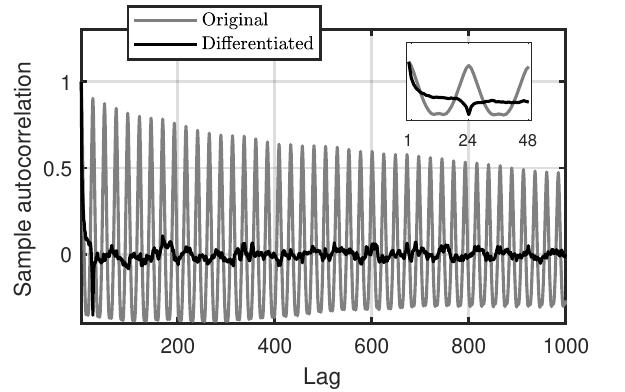}
  \label{fig:orange_autocorr}}\hspace{0em}%
  \caption{In (a), the load profile of \texttt{ts1}, the incoming calls volume, for 300 time intervals (hours). In (b), the autocorrelation functions of the time series \texttt{ts1} before (gray line) and after (black line) a seasonal differentiation. The original time series shows a strong seasonal pattern at lag 24, while after seasonal differencing, the time series does not show any strong correlation or trend. }
\end{figure}

In this work, we focus on predicting the volume (in minutes) of the incoming calls in \texttt{ts1} of the next day.
Due to the hourly resolution of the data, the STFL problem consists of a 24 step-ahead prediction.
The profile of \texttt{ts1} for 300 hours is depicted in Fig.~\ref{fig:callVolume}. 
%
%
The remaining time series are treated as exogenous variables and, according to a common practice in time series forecasting \cite{franses1991seasonality}, they are fed into the network to provide the model with additional information for improving the prediction of the target time series. 
Each time series contain $3336$ measurements, hourly sampled.
We used the first 70\% as training set, the successive 15\% as validation set and the remaining 15\% as test set.
The accuracy of each RNN model is evaluated on this last set.

In each time series there is a (small) fraction of missing values.
In fact, if in a given hour no activities are registered in the area covered by the considered antenna, the relative entries do not appear in the database. 
As we require the target time series and the exogenous ones to have same lengths and to contain a value in each time interval, we inserted an entry with value ``0'' in the dataset to fill the missing values. 
Another issue is the presence of corrupted data, marked by a ``-1'' in the dataset, which are relative to periods when the telephone activity is not registered correctly. 
To address this problem, we followed the procedure described by \citet{shen2005analysis} and we replaced the corrupted entries with the average value of the corresponding periods (same weekday and hour) from the two adjacent weeks. 
Contrarily to some other works on STLF \cite{ibrahim2013forecasting, shen2008interday,andrews1995ll}, we decided to not discard outliers, such as holidays or days with an anomalous number of calls, nor we modeled them as separate variables.

As next step in our pre-analysis, we identify the main seasonality in the data.
We analyze \texttt{ts1}, but similar considerations hold also for the remaining time series.
Through frequency analysis and by inspecting the autocorrelation function, depicted as a gray line in Fig.~\ref{fig:orange_autocorr}, it emerges a strong seasonal pattern every $24$ hours.
As expected, data experience regular and predictable daily changes, due to the nature of the telephonic traffic.
This cycle represents the main seasonality and we filter it out by applying a seasonal differencing with lag 24.
In this way, the RNNs focus on learning to predict the series of changes in each seasonal cycle.
The practice of removing the seasonal effect from the time series, demonstrated to improve the prediction accuracy of models based on neural networks \cite{4359174,doi:10.3846/20294913.2015.1070772}.
The black line in Fig.~\ref{fig:orange_autocorr} depicts the autocorrelation of the time series after seasonal differentiation. 
Except from the high anticorrelation at lag $24$, introduced by the differentiation, the time series appears to be uncorrelated elsewhere and, therefore, we can exclude the presence of a second, less obvious seasonality.

\begin{SCfigure}[1.2][!ht]
  \centering
  \includegraphics[width=0.4\textwidth,keepaspectratio]{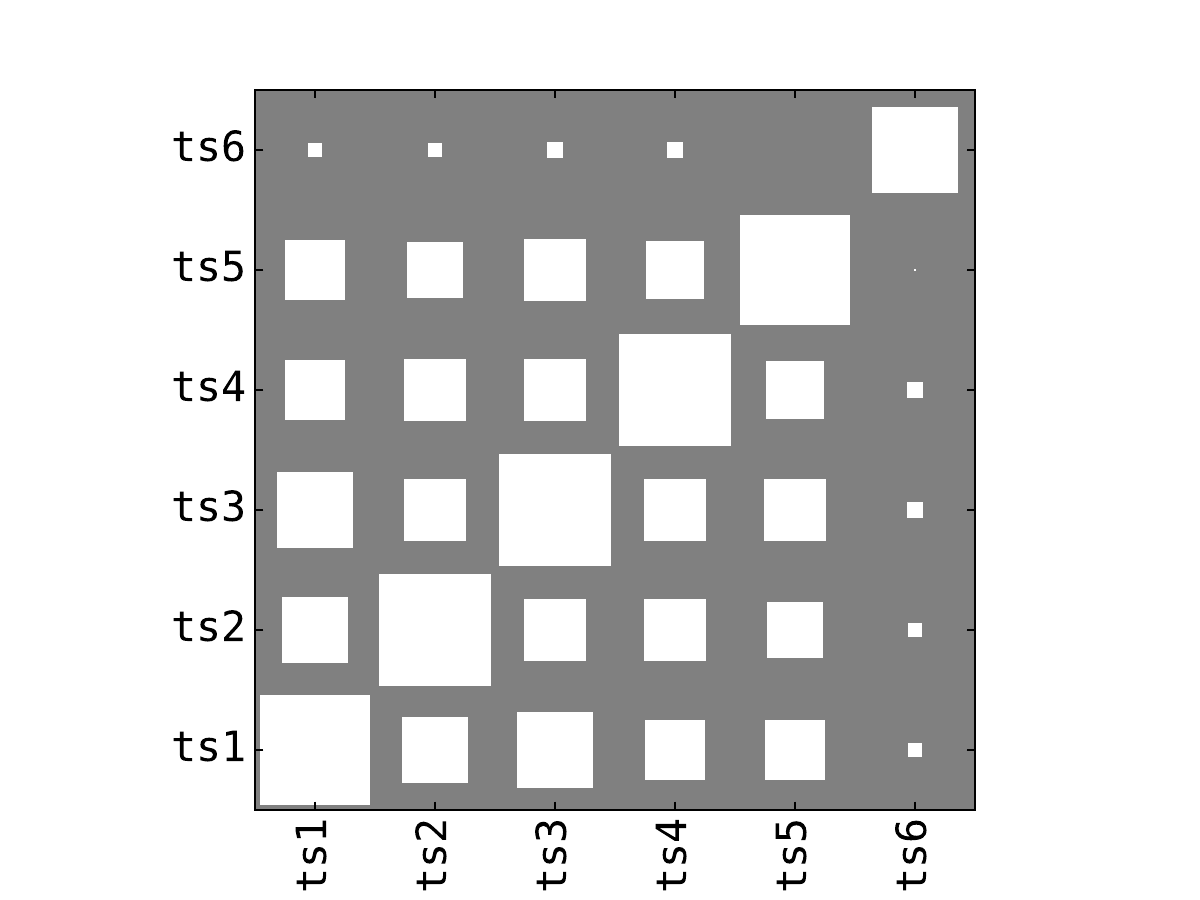}
  \caption{Hinton diagram of the mutual information between the time series in the Orange dataset. 
	The size of each block is proportional to the degree of mutual information among the time series.
	The measurements indicates a strong relationship between the load time series and the daily hours (\texttt{ts5}), while the dependency with the day of the week (\texttt{ts6}) is low.}
  \label{fig:hinton_orange}
\end{SCfigure}

Due to the nature of the seasonality in the data, we expect a strong relationship between the time series of the loads (\texttt{ts1} - \texttt{ts4}) and \texttt{ts5}, which is relative to the hour of the day.
On the other hand, we envisage a lower dependency of the loads with \texttt{ts6}, the time series of the week days, since we did not notice the presence of a second seasonal cycle after the differentiation at lag $24$.
To confirm our hypothesis, we computed the mutual information between the time series, which are reported in the Hinton diagram in Fig.~\ref{fig:hinton_orange}.
%
%
The size of the blocks is proportional to the degree of mutual information among the time series. 
Due to absence of strong relationships, we decided to discard \texttt{ts6} to reduce the complexity of the model by excluding a variable with potentially low impact in the prediction task.
We also discarded \texttt{ts5} because the presence of the cyclic daily pattern is already accounted by doing the seasonal differencing at lag $24$. Therefore, there is not need to provide daily hours as an additional exogenous input. 

Beside differentiation, a common practice in STLF is to apply some form of normalization to the data.
We applied a standardization (z-score), but rescaling into the interval $[-1,1]$ or $[0,1]$ are other viable options.
Additionally, a nonlinear transformation of the data by means of a non-linear function (e.g., square-root or logarithm) can remove some kinds of trend and stabilize the variance in the data, without altering too much their underlying structure \cite{weinberg2007bayesian, shen2008interday, ibrahim2013forecasting}. 
In particular, a log-transform is suitable for a set of random variables characterized by a high variability in their statistical dispersion (heteroscedasticity), or for a process whose fluctuation of the variance is larger than the fluctuation of the mean (overdispersion).
To check those properties, we analyze the mean and the variance of the telephonic traffic within the main seasonal cycle across the whole dataset.
%
\begin{figure*}[!ht]
\centering
  \subfigure[Raw data]{
  \includegraphics[width=0.45\textwidth]{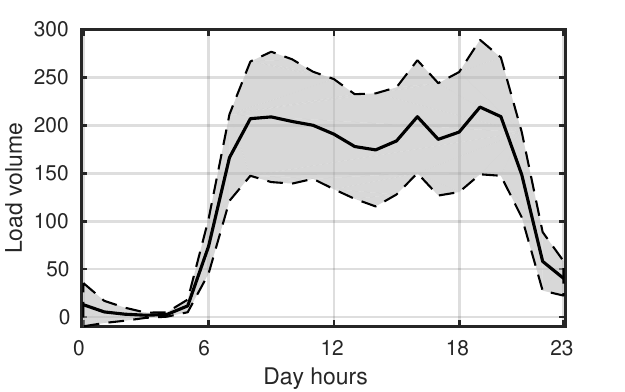}
  \label{fig:orange_var}}
  ~
  \subfigure[Log-transformed data]{
  \includegraphics[width=0.45\textwidth]{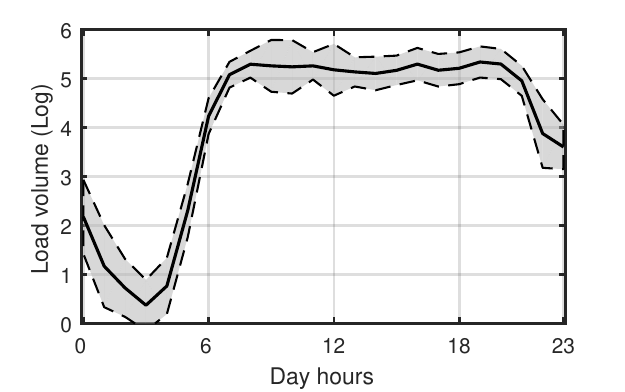}
  \label{fig:orange_var_log}}
\caption{Average weekly load (solid black line) and the standard deviation (shaded gray area) of the telephonic activity in the whole dataset.}
\end{figure*}
%
The solid black line in Fig.~\ref{fig:orange_var}, represents the mean load of \texttt{ts1}, while the shaded gray area illustrates the variance.
As we can see, the data are not characterized by overdispersion, since the fluctuations of the mean are greater than the ones of the variance.
However, we notice the presence of heteroscedasticity, since the amount of variance changes in different hours of the day.
In fact, the central hours where the amount of telephonic activity is higher, are characterized by a greater standard deviation in the load.
In Fig.~\ref{fig:orange_var_log}, we observe that by applying a log-transform we significantly reduce the amount of variance in the periods characterized by a larger traffic load.
However, after the log-transformation the mean value of the load become more flattened and the variance relative to periods with lower telephonic activity is enhanced.
This could cause issues during the training of the RNN, hence in the experiments we evaluate the prediction accuracy both with and without applying the log-transformation to the data.

Preprocessing transformations are applied in this order: (i) log-transform, (ii) seasonal differencing at lag 24, (iii) standardization.
Each preprocessing operation is successively reversed to evaluate the forecast produced by each RNN.

\subsection{ACEA dataset -- electricity load}

The second time series we analyze is relative to the electricity consumption registered by ACEA (Azienda Comunale Energia e Ambiente), the company which provides the electricity to Rome and some neighbouring regions. The ACEA power grid in Rome consists of 10.490 km of medium voltage lines, while the low voltage section covers 11.120 km. 
The distribution network is constituted of backbones of uniform section, exerting radially and with the possibility of counter-supply if a branch is out of order. 
Each backbone is fed by two distinct primary stations and each half-line is protected against faults through the breakers. Additional details can be found in Ref. \cite{DeSantis2015368}. 
The time series we consider concerns the amount of supplied electricity, measured on a medium voltage feeder from the distribution network of Rome. 
Data are collected every $10$ minutes for $954$ days of activity (almost 3 years), spanning from 2009 to 2011, for a total of $137444$ measurements.
Also in this case, we train the RNNs to predict the electricity load 24h ahead, which corresponds to 144 time step ahead prediction.
For this forecast task we do not provide any exogenous time series to the RNNs.
In the hyperparameter optimization, we use the load relative to the first 3 months as training set and the load of the 4\textsuperscript{th} month as validation set.
Once the best hyperparameter configuration is identified, we fine-tune each RNN on the first 4 months and we use the 5\textsuperscript{th} month as test set to evaluate and to compare the accuracy of each network.

\begin{figure}[!ht]
  \centering
  \subfigure[Load profile]{
  \includegraphics[width=0.45\textwidth,keepaspectratio]{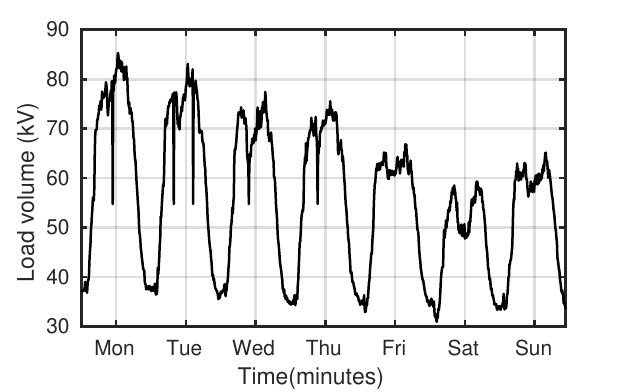}
  \label{fig:aceaVolume}}
  ~
  \subfigure[Autocorrelation functions]{
  \includegraphics[width=0.45\textwidth,keepaspectratio]{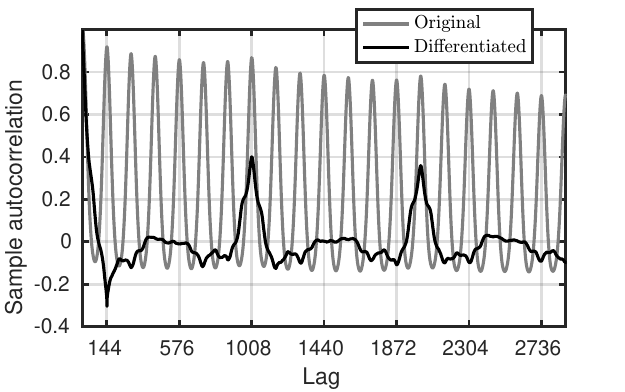}
  \label{fig:acea_autocorr}}
  \caption{In (a), the load profile in kiloVolts (kV) of the electricity consumption registered over one week. The sampling time is 10 minutes.
  In (b), the autocorrelation functions of the ACEA time series before (gray line) and after (black line) a seasonal differentiation at lag 144. 
  The original time series shows a strong seasonal pattern at lag 144, which corresponds to a daily cycle. 
  After seasonal differencing, a previously hidden pattern is revealed at lag 1008, which corresponds to a weekly cycle.}
\end{figure}

A profile of the electric consumption over one week (1008 measurements), is depicted in Fig.~\ref{fig:aceaVolume}.
%

In the ACEA time series there are no missing values, but 742 measurements (which represent 0.54\% of the whole dataset) are corrupted.
The consumption profile is more irregular in this time series, with respect to the telephonic data from the Orange dataset. 
Therefore, rather than replacing the corrupted values with an average load, we used a form of imputation with a less strong bias.
Specifically, we first fit a cubic spline to the whole dataset and then we replaced the corrupted entries with the corresponding values from the fitted spline.
In this way, the imputation better accounts for the local variations of the load.

Also in this case, we perform a preemptive analysis in order to understand the nature of the seasonality, to detect the presence of hidden cyclic patterns, and to evaluate the amount of variance in the time series.
By computing the autocorrelation function up to a sufficient number of lags, depicted as a gray line in Fig.~\ref{fig:acea_autocorr}, it emerges a strong seasonality pattern every 144 time intervals.
As expected, this corresponds exactly to the number of measurements in one day.
By differencing the time series at lag 144, we remove the main seasonal pattern and the trend.
Also in this case, the negative peak at lag 144 is introduced by the differentiation.
If we observe the autocorrelation plot of the time series after seasonal differencing (black line in \ref{fig:acea_autocorr}), a second strong correlation appears each 1008 lags.
This second seasonal pattern represents a weekly cycle, that was not clearly visible before the differentiation.
Due to the long periodicity of the time cycle, to account this second seasonality a predictive model would require a large amount of memory to store information for a longer time interval.
While a second differentiation can remove this second seasonal pattern, we would have to discard the values relative to the last week of measurements.
Most importantly, the models we train could not learn the similarities in consecutive days at a particular time, since they would be trained on the residuals of the load at the same time and day in two consecutive weeks.
Therefore, we decided to apply only the seasonal differentiation at lag 144.

\begin{figure*}[!ht]
\centering
  \subfigure[Whole dataset]{
  \includegraphics[width=0.3\textwidth]{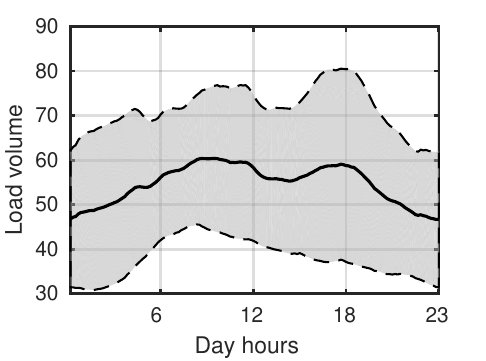}
  \label{fig:acea_var}}
  ~
  \subfigure[January]{
  \includegraphics[width=0.3\textwidth]{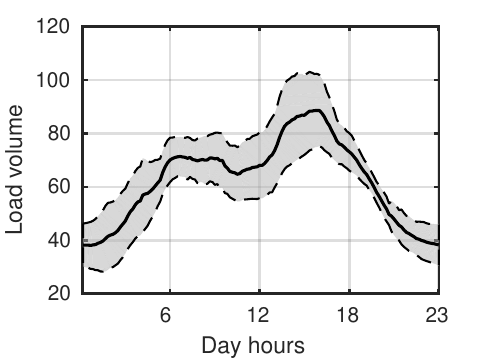}
  \label{fig:acea_var2}}
  ~
  \subfigure[June]{
  \includegraphics[width=0.3\textwidth]{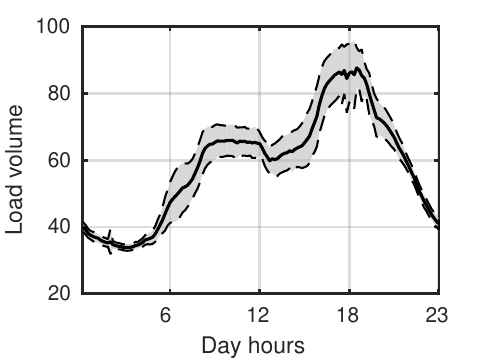}
  \label{fig:acea_var1}}
\caption{In (a) we report the mean load (black line) and the standard deviation (gray area) of the electricity consumption in a week, accounting the measurements from all the dataset. In (b) and (c), the measurements are relative only to one month of activity, which are January and June respectively. }
\label{fig:acea_vartot}
\end{figure*}

To study the variance in the time series, we consider the average daily load over the main seasonal cycle of 144 time intervals.
As we can see from Fig.~\ref{fig:acea_var}, data appear to be affected by overdispersion, as the standard deviation (gray shaded areas) fluctuates more than the mean.
Furthermore, the mean load value (black solid line) seems to not change much across the different hours, while it is reasonable to expect significant differences in the load between night and day. 
However, we remind that the Acea time series spans a long time lapse (almost 3 years) and that the electric consumption is highly related to external factors such as temperature, daylight saving time, holidays and other seasonal events that change over time.
Therefore, in different periods the load profile may vary significantly.
For example, in Fig.~\ref{fig:acea_var2} we report the load profile relative to the month of January, when temperatures are lower and there is a high consumption of electricity, also in the evening, due to the usage of heating.
In June instead (Fig.~\ref{fig:acea_var1}), the overall electricity consumption is lower and mainly concentrated on the central hours of the day.
Also, it is possible to notice that the load profile is shifted due to the daylight saving time.
As we can see, the daily averages within a single month are characterized by a much lower standard deviation (especially in the summer months, with lower overall load consumption) and the mean consumption is less flat.
Henceforth, a non-linear transformation for stabilizing the variance is not required and, also in this case, standardization is suitable for normalizing the values in the time series. 
Since we focus on a short term forecast, having a high variance in loads relative to very distant periods is not an issue, since the model prediction will depends mostly on the most recently seen values.

To summarize, as preprocessing operation we apply: (i) seasonal differencing at lag 144, (ii) standardization.
As before, the transformations are reverted to estimate the forecast.

\subsection{GEFCom2012 dataset -- electricity load}

The last real world dataset that we study is the time series of electricity consumption from the Global Energy Forecasting Competition 2012 (GEF-Com2012) \cite{GEFCom2012}.
The GEFCom 2012 dataset consists of 4 years (2004 -- 2007) of hourly electricity load collected from a US energy supplier.
The dataset comprehends time series of consumption measurements, from 20 different feeders in the same geographical area. 
The values in each time series represent the average hourly load, which varies from $10.000$kWh to $200.000$kWh.
The dataset also includes time series of the temperatures registered in the area where the electricity consumption is measured.

\begin{figure}[!ht]
  \centering
  \subfigure[Load profile]{
  \includegraphics[width=0.45\textwidth,keepaspectratio]{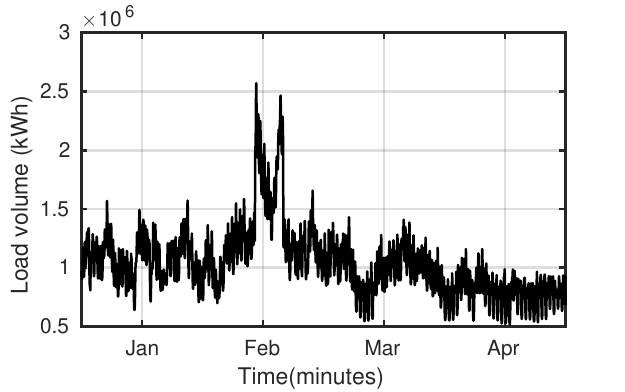}
  \label{fig:gefcomVolume}}
  ~
  \subfigure[Autocorrelation functions]{
  \includegraphics[width=0.45\textwidth,keepaspectratio]{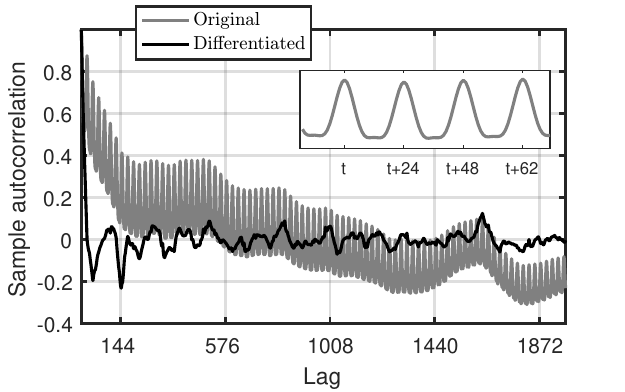}
  \label{fig:gefcom_autocorr}}
  \caption{In (a), the load profile in kilowatt-hour (kWh) of the aggregated electricity consumption registered in the first 4 months of activity in 2006, from the GEFCom dataset. The sampling time in the time series is 1 hour.
  In (b), the autocorrelation functions of the GEFCom time series before (gray line) and after (black line) a seasonal differentiation at lag 24. 
  The small subplot on the top-right part of the figure reports a magnified version of the autocorrelation function before differentiation at lag $t=200$.}
\end{figure}

The forecast task that we tackle is the $24$ hours ahead prediction of the aggregated electricity consumption, which is the sum of the 20 different load time series in year 2006.
The measurements relative to the to first 10 months of the 2006 are used as training set, while the 11\textsuperscript{th} month is used as validation set for guiding the hyperparameters optimization.
The time series of the temperature in the area is also provided to the RNNs as an exogenous input.
The prediction accuracy of the optimized RNNs is then evaluated on the last month of the 2006.
A depiction of the load profile of the aggregated load time series is reported in Fig.~\ref{fig:gefcomVolume}.
%
%
We can observe a trend in the time series, which indicates a decrement in the energy demand over time.
This can be related to climate conditions since, as the temperature becomes warmer during the year, the electricity consumption for the heating decreases.

To study the seasonality in the aggregated time series, we evaluate the autocorrelation function, which is depicted as the gray line in Fig.~\ref{fig:gefcom_autocorr}.
From the small subplot in top-right part of the figure, relative to a small segment of the time series, it emerges a strong seasonal pattern every 24 hours.
By applying a seasonal differentiation with lag 24 the main seasonal pattern is removed, as we can see from the autocorrelation function of the differentiated time series, depicted as a black line in the figure.
After differentiation, the autocorrelation becomes close to zero after the first lags and, therefore, we can exclude the presence of a second, strong seasonal pattern (e.g. a weekly pattern).

\begin{figure*}[!ht]
\centering
  \subfigure[January]{
  \includegraphics[width=0.35\textwidth]{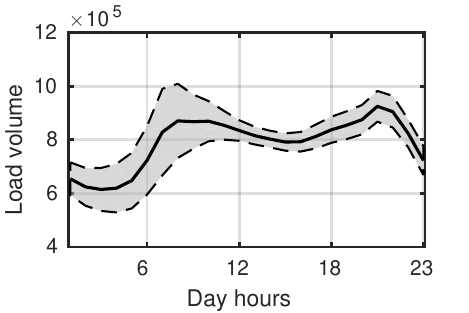}
  \label{fig:gefcom_var1}}
  ~
  \subfigure[May]{
  \includegraphics[width=0.35\textwidth]{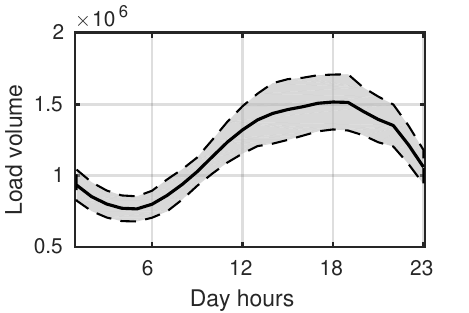}
  \label{fig:gefcom_var2}}
\caption{In (a) the average load (solid black line) and the standard deviation (shaded gray area) of the electricity consumption during one week, in the month of January. In (b), we report the measurements relative to the month of June.}
\label{fig:gefcom_vartot}
\end{figure*}

Similarly to what we did previously, we analyze the average load of the electricity consumption during one week.
As for the ACEA dataset, rather than considering the whole dataset, we analyze separately the load in one month of winter and one month in summer.
In Fig.~\ref{fig:gefcom_var1}, we report the mean load (black line) and standard deviation (gray area) in January.
Fig.~\ref{fig:gefcom_var2} instead, depicts the measurements for May. 
It is possible to notice a decrement of the load during the spring period, due to the reduced usage of heating.
It is also possible to observe a shift in the consumption profile to later hours in the day, due to the time change.
By analyzing the amount of variance and the fluctuations of the mean load, we can exclude the presence of overdispersion and heteroscedasticity phenomena in the data.

\begin{figure}[!ht]
  \centering
  \includegraphics[width=0.7\textwidth,keepaspectratio]{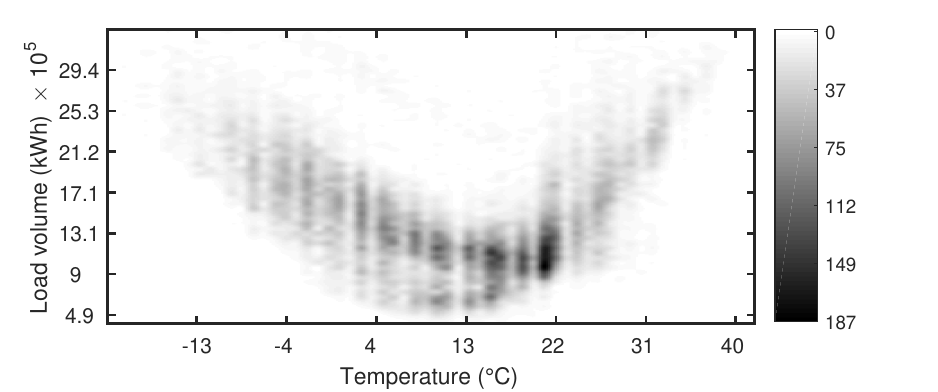}
  \caption{2-dimensional histogram of the aggregated electricity load and temperature in GEFCom dataset. Darker areas represent more populated bins. The bar on the right indicates the number of elements in each bin.
  The characteristic V-shape of the resulting pattern is because of the increased use of heating and cooling devices in presence of hot and cold temperatures.
  }
  \label{fig:gefcom_hist}
\end{figure}

To improve the forecasting accuracy of the electricity consumption, a common practice is to provide to the prediction system the time series of the temperature as an exogenous variable.
In general, the load and the temperature are highly related, since both in the coldest and in the warmest months electricity demand increases, due to the usage of heating and air conditioning, respectively.
However, the relationship between temperature and load cannot be captured by the linear correlation, since the consumption increases both when temperatures are too low or too high.
Indeed, the estimated correlation between the aggregated load time series of interest and the time series of the temperature in the area yields only a value of 0.2.
However, their relationship is evidenced by computing a 2-dimensional histogram of the two variables, proportional to their estimated joint distribution, which is reported in Fig \ref{fig:gefcom_hist}.
The V-shape, denotes an increment of the electricity consumption for low and high temperatures with respect to a mean value of about $22^{\circ}$ C.

The preprocessing operations we apply on the GEFCom dataset are: (i) seasonal differencing at lag 24, (ii) standardization.
Also in this case, these transformations are reverted to estimate the forecast.

\section{Experiments}
\label{sec:experiments}

In this section we compare the prediction performance achieved by the network architectures presented in Sec. \ref{sec:rnn_architectures} and \ref{sec:other_architectures} on different time series.
For each architecture, we describe the validation procedure we follow to tune the hyperparameters and to find an optimal learning strategy for training the weights.
During the validation phase, different configurations are randomly selected from admissible intervals and, once the training is over, their optimality is evaluated as the prediction accuracy achieved on the validation set.
We opted for a random search as it can find more accurate results than a grid search, when the same number of configurations are evaluated \cite{bergstra2012random}.
Once the (sub)optimal configuration is identified, we train each model $10$ times on the training and validation data, using random and independent initializations of the network parameters, and we report the highest prediction accuracy obtained on the unseen values of the test set.

To compare the forecast capability of each model, we evaluate the prediction accuracy $\psi$ as $\psi=1-\mathrm{NRMSE}$.
NRMSE is the Normalized Root Mean Squared Error that reads
\begin{equation}
  \label{eq:nrmse}
  \mathrm{NRMSE}\left( \mathcal{Y}, \mathcal{Y}^* \right) = \sqrt{\frac{\langle \lVert \mathbf{\mathcal{Y}} - \mathbf{\mathcal{Y}^*} \rVert^2 \rangle}{\langle \lVert \mathcal{Y} - \langle \mathcal{Y}^* \rangle \rVert^2 \rangle}},
\end{equation}
where $\langle \cdot \rangle$ computes the mean, $\mathcal{Y}$ are the RNN outputs and $\mathcal{Y}^*$ are the ground-truth values. 

In the following, we present two types of experiments. 
The first experiment consists in the prediction of the synthetic time series presented in Sec. \ref{sec:synth_data}, commonly considered as benchmarks in forecast applications, and the results are discussed in Sec. \ref{sec:synth_res}.
In the second experiment we forecast the real-world telephonic and electricity load time series, presented in Sec. \ref{sec:real_data}. 
The results of this second experiment are discussed in Sec. \ref{sec:real_res}.

\subsection{Experimental settings}

\paragraph{ERNN, LSTM and GRU}

The three RNNs described in Sec. \ref{sec:rnn_architectures} have been implemented in Python, using Keras library with Theano \cite{2016arXiv160502688short} as backend\footnote{Keras library is available at \url{https://github.com/fchollet/keras}. Theano library is available at \url{http://deeplearning.net/software/theano/}}.

To identify an optimal configuration for the specific task at hand, we evaluate for each RNN different values of the hyperparameters and training procedures.
The configurations are selected randomly and their performances are evaluated on the validation set, after having trained the network for $400$ epochs.
To get rid of the initial transient phase, we drop the first $50$ outputs of the network.
A total of $500$ random configurations for each RNN are evaluated and, once the optimal configuration is found, we compute the prediction accuracy on the test set.
In the test phase, each network is trained for $2000$ epochs.

The optimization is performed by assigning to each hyperparameter a value uniformly sampled from a given interval, which can be continuous or discrete.
The gradient descent strategies are selected from a set of possible alternatives, which are SGD, Nesterov momentum and Adam.
For SGD and Nesterov, we anneal the learning rate with a step decay of $10^{-6}$ in each epoch.
The learning rate $\eta$ is sampled from different intervals, depending on the strategy selected. 
Specifically, for SGD we set $\eta = 10^{c}$, with $c$ uniformly sampled in $[-3, -1]$. 
For Nesterov and Adam, since they benefit from a smaller initial value of the learning rate, we sample $c$ uniformly in $[-4,-2]$.
The remaining hyperparameters used in the optimization strategies are kept fixed to their default values (see Sec. \ref{sec:strategies}).
Regarding the number $N_h$ of hidden units in the recurrent hidden layer, we randomly chose for each architecture four possible configurations that yield an amount of trainable parameters approximately equal to $1800$, $3900$, $6800$, and $10000$. 
This corresponds to $N_h = \{40,60,80,100\}$ in ERNN, $N_h = \{20,30,40,50\}$ in LSTM and $N_h = \{23,35,46,58\}$ in GRU. 
For each RNNs, $N_h$ is randomly selected from these sets. 
To deal with the problem of vanishing gradient discussed in Sec. \ref{sec:vanishing}, we initialize the RNN weights by sampling them from an uniform distribution in $[0,1]$ and then rescaling their values by $1/\sqrt{N_h}$.
For the $\mathrm{L}_1$ and $\mathrm{L}_2$ regularization terms, we sample independently $\lambda_{1}$ and $\lambda_{2}$ from $[0, 0.1]$, an interval containing values commonly assigned to these hyperparameters in RNNs \cite{DBLP:journals/corr/ZeyerDVSN16}.
We apply the same regularization to input, recurrent and output weights.
As suggested by \citet{2015arXiv151205287G}, we drop the same input and recurrent connections at each time step in the BPTT, with a dropout probability $p_{\text{drop}}$ drawn from $\{0, 0.1, 0.2, 0.3, 0.5 \}$, which are commonly used values \cite{6981034}. 
If $p_{\text{drop}} \neq 0$, we also apply a $\mathrm{L}_2$ regularization. 
This combination usually yields a lowest generalization error than dropout alone \cite{srivastava2014dropout}.
Note that another possible approach combines dropout with the max-norm constraint, where the $\mathrm{L}_2$ norm of the weights is clipped whenever it grows beyond a given constant, which, however, introduces another hyperparameter.

For the training we consider the backpropagation through time procedure $\mathrm{BPTT}(\tau_b,\tau_f)$ with $\tau_b = 2 \tau_f$.
The parameter $\tau_f$ is randomly selected from the set $\{ 10, 15, 20, 25, 30 \}$. 
As we discussed in Sec. \ref{sec:bppt}, this procedure differs from both the \textit{true} BPTT and the \textit{epochwise} BPTT \cite{williams1995gradient}, which is implemented as default by popular deep learning libraries such as TensorFlow \cite{tensorflow2015-whitepaper}.

\paragraph{NARX}

This RNN is implemented using the Matlab Neural Network toolbox\footnote{\url{https://se.mathworks.com/help/nnet/ref/narxnet.html}}.
We configured NARX network with an equal number of input and output lags on the TDLs ($d_x = d_y$) and with the same number of neurons $N_h$ in each one of the $N_l$ hidden layers.
Parameters relative to weight matrices and bias values $\Theta = \left\{ \theta_, \theta_o, \theta_{h_1}, \dots, \theta_{h_{N_l}} \right\}$ are trained with a variant of the quasi Newton search, called Levenberg-Marquardt optimization algorithm. 
This is an algorithm for error backpropagation that provides a good tradeoff between the speed of the Newton algorithm and the stability of the steepest descent method \cite{battiti1992first}. 
The loss function to be minimized is defined in Eq. \ref{eq:narx_loss}. 

NARX requires the specification of $5$ hyperparameters, which are uniformly drawn from different intervals. 
Specifically, TDL lags are drawn from $\{ 2, 3, \dots, 10 \}$; 
the number of hidden layers $N_l$ is drawn from $\{1, 2, \dots, 5 \}$; 
the number of neurons $N_h$ in each layer is drawn from $\{ 5, 6, \dots, 20 \}$; 
the regularization hyperparameter $\lambda_2$ in the loss function is randomly selected from $\{ 2^{-1}, 2^{-2}, \ldots, 2^{-10}\}$; 
the initial value $\eta$ of learning rate is randomly selected from $\{ 2^{-5}, 2^{-6}, \ldots, 2^{-25}\}$.

A total of $500$ random configurations for NARX are evaluated and, for each hyperparameters setting, the network is trained for $1000$ epochs in the validation.
In the test phase, the network configured with the optimal hyperparameters is trained for $2000$ epochs. 
Also in this case, we discard the first $50$ network outputs to get rid of the initial transient phase of the network.

\paragraph{ESN}

For the ESN, we used a modified version of the Python implementation\footnote{\url{https://github.com/siloekse/PythonESN}}, provided by \citet{Lokse2017}.
Learning in ESN is fast, as the readout is trained by means of a linear regression.
However, the training does not influence the internal dynamics of the random reservoir, which can be controlled only through the ESN hyperparameters.
This means that a more accurate (and computationally intensive) search of the optimal hyperparametyers is required with respect to the other RNN architectures.
In RNNs, the precise, yet slow gradient-based training procedure is mainly responsible for learning the necessary dynamics and it can compensate a suboptimal choice of the hyperparameters.

Therefore, in the ESN validation phase we evaluate a larger number of configurations ($5000$), by uniformly drawing $8$ different hyperparameters from specific intervals.
In particular, the number of neurons in the reservoir, $N_h$, is drawn from $\{ 400, 450, \dots, 900 \}$; 
the reservoir spectral radius, $\rho$, is drawn in the interval $[0.5, 1.8]$; 
the reservoir connectivity $R_c$ is drawn from $[0.15, 0.45]$; 
the noise term $\xi$ in Eq. \eqref{eq:esn_state_update} comes from a Gaussian distribution with zero mean and variance drawn from $[0, 0.1]$; 
scaling of input signal $\omega_i$ and desired response $\omega_o$ are drawn from $[0.1, 1]$; 
scaling of output feedback $\omega_f$ is drawn from $[0, 0.5]$; 
the linear regression regularization parameter $\lambda_2$ is drawn from $[0.001, 0.4]$. 
Also in this case, we discarded the first $50$ ESN outputs relative to the initial transient phase.

\subsection{Results on synthetic dataset}
\label{sec:synth_res}

\begin{SCfigure}[1.0][!ht]
  \centering
  \includegraphics[width=0.5\textwidth,keepaspectratio]{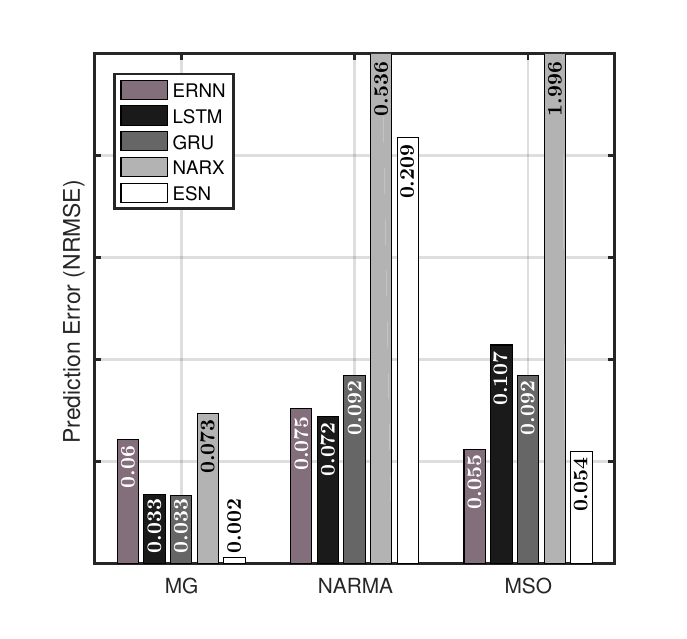}
  \caption{NRMSE values achieved on the test sets by each RNN architecture on the three synthetic prediction tasks.}
  \label{fig:synth_res}
\end{SCfigure}

In Fig.~\ref{fig:synth_res} we report the prediction accuracy obtained by the RNNs on the test set of the three synthetic problems.
The best configurations of the architectures identified for each task through random search are reported in Tab.~\ref{tab:rnn_synth}.

\bgroup
\def\arraystretch{.7} 
\setlength\tabcolsep{.5em} 
\begin{center}
\begin{table}[ht]\small
\centering
  \caption{Optimal RNNs configurations for solving the three synthetic prediction tasks, MG, NARMA and MSO. 
  The acronyms in the table are: 
  $N_h$ -- number of nodes in the hidden layer; 
  $N_l$ -- number of hidden layers; 
  TDL -- number of lags on the tapped delay lines;
  $\eta$ -- learning rate;
  $\lambda_1$ -- $\mathrm{L}_1$ regularization parameter;
  $\lambda_2$ -- $\mathrm{L}_2$ regularization parameter;
  OPT -- gradient descent strategy;
  $\tau_\text{f}$ -- number of new time steps processed before computing the BPTT;
  $\tau_\text{b}$ -- number of time step the gradient is propagated back in BPTT;
  $p_\text{drop}$ -- dropout probability;
  $\rho$ -- spectral radius of ESN reservoir;
  $R_c$ -- percentage of sparsity in ESN reservoir;
  $\xi$ -- noise in ESN state update; 
  $\omega_i$, $\omega_o$, $\omega_f$ -- scaling of input, teacher and feedback weights.}
\vspace{0.2cm}
\begin{tabular}{crcccccccc}
\cmidrule[1pt]{1-10}
\multicolumn{1}{c}{\textbf{Network}} & \multicolumn{1}{r}{\textbf{Task}} & \multicolumn{8}{c}{\textbf{RNN Configuration}} \\
\cmidrule[1pt]{1-10}
\multirow{4}{*}{\vspace{-1.0em}Narx} & & \multicolumn{2}{c}{$\boldsymbol{N_h}$} & \multicolumn{2}{c}{$\boldsymbol{N_l}$} & \multicolumn{2}{c}{\textbf{TDL}} & \multicolumn{1}{c}{$\boldsymbol{\eta}$} & \multicolumn{1}{c}{$\boldsymbol{\lambda_2}$} \\
\cmidrule[.5pt]{3-10}
& MG    & \multicolumn{2}{c}{15} & \multicolumn{2}{c}{2} & \multicolumn{2}{c}{6} & \multicolumn{1}{c}{$\text{3.8E-6}$} & \multicolumn{1}{c}{$0.0209$} \\
& NARMA & \multicolumn{2}{c}{17} & \multicolumn{2}{c}{2} & \multicolumn{2}{c}{10} & \multicolumn{1}{c}{$\text{2.4E-4}$} & \multicolumn{1}{c}{$0.4367$} \\
& MSO   & \multicolumn{2}{c}{12} & \multicolumn{2}{c}{5} & \multicolumn{2}{c}{2} & \multicolumn{1}{c}{$0.002$} & \multicolumn{1}{c}{$0.446$} \\
\cmidrule[1pt]{1-10}
\multirow{4}{*}{\vspace{-1.0em}ERNN} &  & $\boldsymbol{\tau}_{\textbf{b}}$ & $\boldsymbol{\tau}_{\textbf{f}}$ & \multicolumn{1}{c}{$\boldsymbol{N_h}$} & \multicolumn{1}{c}{\textbf{OPT}} & \multicolumn{1}{c}{$\boldsymbol{\eta}$} & \multicolumn{1}{c}{$\boldsymbol{p_\textbf{drop}}$} & $\boldsymbol{\lambda_1}$ & $\boldsymbol{\lambda_2}$ \\
\cmidrule[.5pt]{3-10}
& MG    & 20 & 10 & 80 & Adam & 0.00026 & 0 & 0 & 0.00037 \\
& NARMA & 50 & 25 & 80 & Nesterov & 0.00056 & 0 & 0 & 1E-5 \\
& MSO   & 50 & 25 & 60 & Adam & 0.00041 & 0 & 0 & 0.00258 \\
\cmidrule[1pt]{1-10}
\multirow{4}{*}{\vspace{-1.0em}LSTM} &  & $\boldsymbol{\tau}_{\textbf{b}}$ & $\boldsymbol{\tau}_{\textbf{f}}$ & \multicolumn{1}{c}{$\boldsymbol{N_h}$} & \multicolumn{1}{c}{\textbf{OPT}} & \multicolumn{1}{c}{$\boldsymbol{\eta}$} & \multicolumn{1}{c}{$\boldsymbol{p_\textbf{drop}}$} & $\boldsymbol{\lambda_1}$ & $\boldsymbol{\lambda_2}$ \\
\cmidrule[.5pt]{3-10}
& MG    & 50 & 25 & 40 & Adam & 0.00051 & 0 & 0 & 0.00065 \\
& NARMA & 40 & 20 & 40 & Adam & 0.00719 & 0 & 0 & 0.00087 \\
& MSO   & 50 & 25 & 20 & Adam & 0.00091 & 0 & 0 & 0.0012 \\
\cmidrule[1pt]{1-10}
\multirow{4}{*}{\vspace{-1.0em}GRU} &  & $\boldsymbol{\tau}_{\textbf{b}}$ & $\boldsymbol{\tau}_{\textbf{f}}$ & \multicolumn{1}{c}{$\boldsymbol{N_h}$} & \multicolumn{1}{c}{\textbf{OPT}} & \multicolumn{1}{c}{$\boldsymbol{\eta}$} & \multicolumn{1}{c}{$\boldsymbol{p_\textbf{drop}}$} & $\boldsymbol{\lambda_1}$ & $\boldsymbol{\lambda_2}$ \\
\cmidrule[.5pt]{3-10}
& MG    & 40 & 20 & 46 & SGD  & 0.02253 & 0 & 0 & 6.88E-6 \\
& NARMA & 40 & 20 & 46 & Adam & 0.00025 & 0 & 0 & 0.00378 \\
& MSO   & 50 & 25 & 35 & Adam & 0.00333 & 0 & 0 & 0.00126 \\
\cmidrule[1pt]{1-10}
\multirow{4}{*}{\vspace{-1.0em}ESN} & & $\mathbf{N_h}$ & $\boldsymbol{\rho}$ & $\mathbf{R_c}$ & $\boldsymbol{\xi}$ & $\boldsymbol{\omega_i}$ & $\boldsymbol{\omega_o}$ & $\boldsymbol{\omega_f}$ & $\boldsymbol{\lambda_2}$ \\
\cmidrule[.5pt]{3-10}
& MG    & 800 & 1.334 & 0.234 & 0.001 & 0.597 & 0.969 & 0.260 & 0.066 \\
& NARMA & 700 & 0.932 & 0.322 & 0.013 & 0.464 & 0.115 & 0.045 & 0.343 \\
& MSO   & 600 & 1.061 & 0.231 & 0.002 & 0.112 & 0.720 & 0.002 & 0.177 \\
\cmidrule[1pt]{1-10}
\end{tabular}
\label{tab:rnn_synth}
\end{table}
\end{center}
\egroup

First of all, we observe that the best performing RNN is different in each task.
In the MG task, ESN outperforms the other networks.
This result confirms the excellent and well-known capability of the ESN in predicting chaotic time series \cite{6177672, jaeger2004harnessing}.
In particular, ESN demonstrated to be the most accurate architecture for the prediction of the MG system \cite{shi2007support}. 
The ESN achieves the best results also in the MSO task, immediately followed by ERNN.
On the NARMA task instead, ESN performs poorly, while the LSTM is the RNN that predicts the target signal with the highest accuracy.

In each test, NARX struggles in reaching performance comparable with the other architectures.
In particular, in NARMA and MSO task the NRMSE prediction error of NARX is $0.53$ and $1.99$, respectively (note that we cut the y-axis to better show the remaining bars).
Note that, since the NRMSE is normalized by the variance of the target signal, an error greater than 1 means that the performance is worse than a constant predictor, with value equal to the mean of the target signal.

It is also interesting to notice that in MSO, ERNN achieves a prediction accuracy higher than GRU and LSTM. 
Despite the fact that the MSO task demands a large amount of memory, due to the extremely long periodicity of the target signal, the two gated architectures (LSTM and GRU) are not able to outperform the ERNN.
We can also notice that for MSO the optimal number of hidden nodes ($N_h$) is lower than in the other tasks.
A network with a limited complexity is less prone to overfit on the training data, but it is also characterized by an inferior modeling capability.
Such a high modeling capability is not needed to solve the MSO task, given that the network manages to learn correctly the frequencies of the superimposed sinusoidal signals.

Finally, we observe that LSTM and GRU performs similarly on the each task, but there is not a clear winner.
This finding is in agreement with previous studies, which, after several empirical evaluations, concluded that it is difficult to choose in advance the most suitable gated RNN to solve a specific problem \cite{DBLP:journals/corr/ChungGCB14}. 

Regarding the gradient descent strategies used to train the parameters in RNN, LSTM and GRU, we observe in Tab.~\ref{tab:rnn_synth} that Adam is often identified as the optimal strategy.
The standard SGD is selected only for GRU in the MG task.
This is probably a consequence of the lower convergence rate of the SGD minimization, which struggles to discover a configuration that achieves a good prediction accuracy on the validation set in the limited amount ($400$) of training epochs.
Also, the Nesterov approach seldom results to be as the optimal strategy and a possible explanation is its high sensitivity to the (randomly selected) learning rate. 
In fact, if the latter is too high, the gradient may build up too much momentum and bring the weights into a configuration where the loss function is very large. 
This results in even greater gradient updates, which leads to rough oscillations of the weights that can reach very large values. 

From the optimal configurations in Tab.~\ref{tab:rnn_synth}, another striking behavior about the optimal regularization procedures emerges.
In fact, we observe that in each RNN and for each task, only the L\textsubscript{2} norm of the weights is the optimal regularizer.
On the other hand, the parameters $\lambda_1$ and $p_\text{drop}$ relative to the L\textsubscript{1} norm and the dropout are always zero.
This indicates that, to successfully solve the synthetic prediction tasks, it is sufficient to train the networks with small weights in order to prevent the overfitting.

Finally, we notice that the best results are often found using network with a high level of complexity, in terms of number of neurons and long windows in BPTT or TDL, for Narx.
In fact, in most cases the validation procedure identifies the optimal values for these variables to be close to the upper limit of their admissible intervals.
This is somehow expected, since a more complex model can achieve higher modeling capabilities, if equipped with a suitable regularization procedure to prevent overfitting during training.
However, the tradeoff in terms of computational resources for training more complex models is often very high and small increments in the performance are obtained at the cost of much longer training times.

\subsection{Results on real-world dataset}
\label{sec:real_res}

\begin{SCfigure}[1.0][!ht]
  \centering
  \includegraphics[width=0.55\textwidth,keepaspectratio]{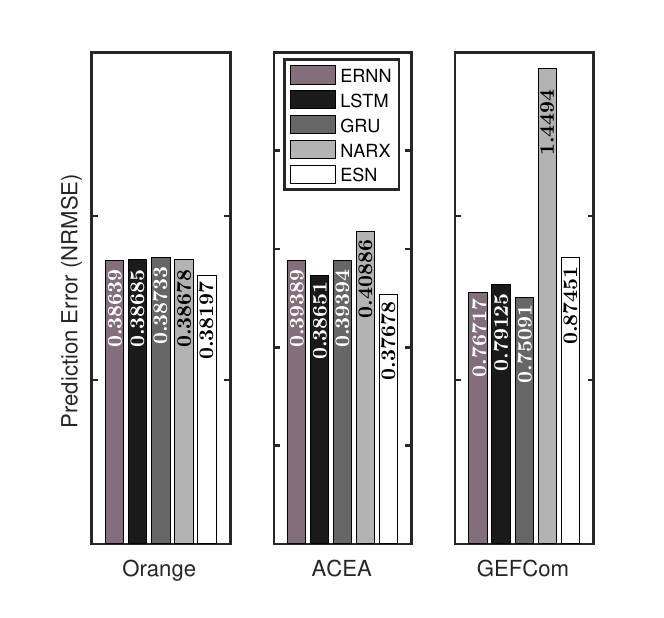}
  \caption{NRMSE values achieved on the test sets by each RNN architecture on the three real-world STLF problems. 
  Note that scales are different for each dataset}
  \label{fig:real_res}
\end{SCfigure}

The highest prediction accuracies obtained by the RNNs on the test set (unseen data) of the real-world load time series, are reported in Fig.~\ref{fig:real_res}.
As before, in Tab.~\ref{tab:rnn_real} we report the optimal configuration of each RNN for the different tasks.

\bgroup
\def\arraystretch{.7} 
\setlength\tabcolsep{.5em} 
\begin{center}
\begin{table}[ht]\small
\centering
  \caption{Optimal RNNs configurations adopted in the three real-world STLF problems.
  Refer to Tab.~\ref{tab:rnn_synth} for the definition of the acronyms in this table.}
\vspace{0.2cm}
\begin{tabular}{crcccccccc}

\cmidrule[1pt]{1-10}
\multicolumn{1}{c}{\textbf{Network}} & \multicolumn{1}{r}{\textbf{Task}} & \multicolumn{8}{c}{\textbf{RNN Configuration}} \\
\cmidrule[1pt]{1-10}
\multirow{4}{*}{\vspace{-1.0em}Narx} & & \multicolumn{2}{c}{$\boldsymbol{N_h}$} & \multicolumn{2}{c}{$\boldsymbol{N_l}$} & \multicolumn{2}{c}{\textbf{TDL}} & \multicolumn{1}{c}{$\boldsymbol{\eta}$} & \multicolumn{1}{c}{$\boldsymbol{\lambda_2}$} \\
\cmidrule[.5pt]{3-10}
& Orange	& \multicolumn{2}{c}{11} & \multicolumn{2}{c}{4} & \multicolumn{2}{c}{2} & \multicolumn{1}{c}{$1.9E-6$} & \multicolumn{1}{c}{$0.082$} \\
& ACEA		& \multicolumn{2}{c}{11} & \multicolumn{2}{c}{3} & \multicolumn{2}{c}{2} & \multicolumn{1}{c}{$1.9E-6$} & \multicolumn{1}{c}{$0.0327$} \\
& GEFCom	& \multicolumn{2}{c}{18} & \multicolumn{2}{c}{4} & \multicolumn{2}{c}{9} & \multicolumn{1}{c}{$6.1E-5$} & \multicolumn{1}{c}{$0.3136$} \\

\cmidrule[1pt]{1-10}
\multirow{4}{*}{\vspace{-1.0em}ERNN} &  & $\boldsymbol{\tau}_{\textbf{b}}$ & $\boldsymbol{\tau}_{\textbf{f}}$ & \multicolumn{1}{c}{$\boldsymbol{N_h}$} & \multicolumn{1}{c}{\textbf{OPT}} & \multicolumn{1}{c}{$\boldsymbol{\eta}$} & \multicolumn{1}{c}{$\boldsymbol{p_\textbf{drop}}$} & $\boldsymbol{\lambda_1}$ & $\boldsymbol{\lambda_2}$ \\
\cmidrule[.5pt]{3-10}
& Orange	& 30 & 15 & 100 & SGD & 0.011 & 0 & 0 & 0.0081 \\
& ACEA		& 60 & 30 & 80 & Nesterov & 0.00036 & 0 & 0 & 0.0015 \\
& GEFCom   	& 50 & 25 & 60 & Adam & 0.0002 & 0 & 0 & 0.0023 \\

\cmidrule[1pt]{1-10}
\multirow{4}{*}{\vspace{-1.0em}LSTM} &  & $\boldsymbol{\tau}_{\textbf{b}}$ & $\boldsymbol{\tau}_{\textbf{f}}$ & \multicolumn{1}{c}{$\boldsymbol{N_h}$} & \multicolumn{1}{c}{\textbf{OPT}} & \multicolumn{1}{c}{$\boldsymbol{\eta}$} & \multicolumn{1}{c}{$\boldsymbol{p_\textbf{drop}}$} & $\boldsymbol{\lambda_1}$ & $\boldsymbol{\lambda_2}$ \\
\cmidrule[.5pt]{3-10}
& Orange    & 40 & 20 & 50 & Adam & 0.0013 & 0   & 0 & 0.0036 \\
& ACEA 		& 50 & 25 & 40 & Adam & 0.0010 & 0.1 & 0 & 0.0012 \\
& GEFCom   	& 50 & 25 & 20 & SGD & 0.0881 & 0 & 0 & 0.0017 \\

\cmidrule[1pt]{1-10}
\multirow{4}{*}{\vspace{-1.0em}GRU} &  & $\boldsymbol{\tau}_{\textbf{b}}$ & $\boldsymbol{\tau}_{\textbf{f}}$ & \multicolumn{1}{c}{$\boldsymbol{N_h}$} & \multicolumn{1}{c}{\textbf{OPT}} & \multicolumn{1}{c}{$\boldsymbol{\eta}$} & \multicolumn{1}{c}{$\boldsymbol{p_\textbf{drop}}$} & $\boldsymbol{\lambda_1}$ & $\boldsymbol{\lambda_2}$ \\
\cmidrule[.5pt]{3-10}
& Orange   	& 40 & 20 & 46 & SGD  & 0.0783 & 0 & 0.0133 & 0.0004 \\
& ACEA 		& 40 & 20 & 35 & Adam & 0.0033 & 0 & 0 & 0.0013 \\
& GEFCom   	& 60 & 30 & 23 & Adam & 0.0005 & 0 & 0 & 0.0043 \\

\cmidrule[1pt]{1-10}
\multirow{4}{*}{\vspace{-1.0em}ESN} & & $\mathbf{N_h}$ & $\boldsymbol{\rho}$ & $\mathbf{R_c}$ & $\boldsymbol{\xi}$ & $\boldsymbol{\omega_i}$ & $\boldsymbol{\omega_o}$ & $\boldsymbol{\omega_f}$ & $\boldsymbol{\lambda_2}$ \\
\cmidrule[.5pt]{3-10}
& Orange    & 400 & 0.5006 & 0.3596 & 0.0261 & 0.2022 & 0.4787 & 0.1328 & 0.3240 \\
& ACEA 		& 800 & 0.7901 & 0.4099 & 0.0025 & 0.1447 & 0.5306 & 0.0604 & 0.1297 \\
& GEFCom   	& 500 & 1.7787 & 0.4283 & 0.0489 & 0.7974 & 0.9932 & 0.0033 & 0.2721 \\

\cmidrule[1pt]{1-10}

\end{tabular}
\label{tab:rnn_real}
\end{table}
\end{center}
\egroup

\begin{figure}[!ht]
  \centering
  \includegraphics[width=\textwidth,keepaspectratio]{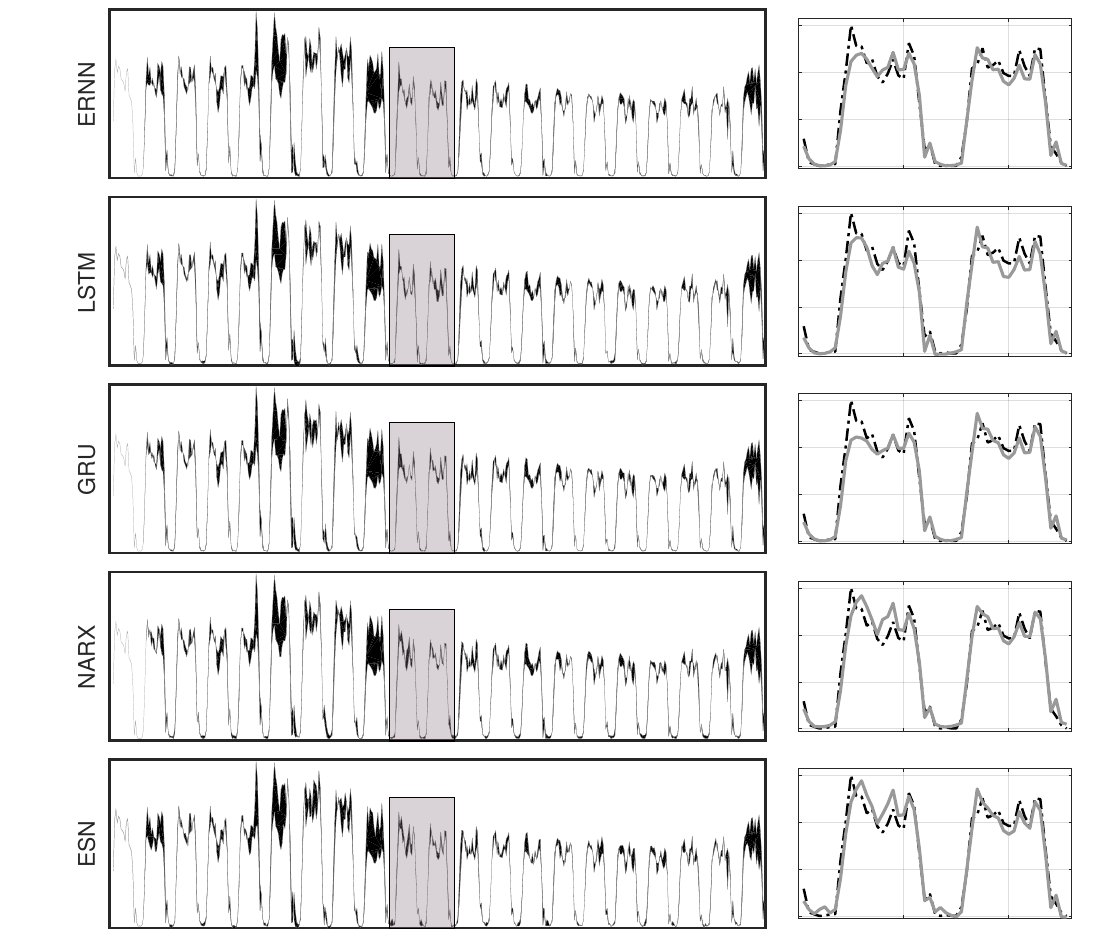}
  \caption{Orange dataset -- The plots on the left show the residuals of predictions of each RNN with respect to the ground truth; black areas indicate the errors in the predictions. The plots on right depict a magnification of the area in the gray boxes from the left graphics; the dashed black line is the ground truth, the solid gray line is the prediction of each RNN.}
  \label{fig:orange_pred}
\end{figure}

\paragraph{Orange}
All the RNNs achieve very similar prediction accuracy on this dataset, as it is possible to see from the first bar plot in Fig.~\ref{fig:real_res}.
In Fig.~\ref{fig:orange_pred} we report the residuals, depicted as black areas, between the target time series and the forecasts of each RNN. 
The figure gives immediately a visual quantification of the accuracy, as the larger the black areas, the greater the prediction error in that parts of the time series.
In particular, we observe that the values which the RNNs fail to predict are often relative to the same interval.
Those values represent fluctuations that are particularly hard to forecast, since they correspond to unusual increments (or decrements) of load, which differ significantly from the trend observed in the past. 
For example, the error increases when the load suddenly grows in the last seasonal cycle in Fig.~\ref{fig:orange_pred}.

In the Orange experiment we evaluate the results with or without applying a log transform to the data. 
We observed sometime log-transform yields slightly worse result ($\sim 0.1\%$), but in most cases the results are equal. 

For ERNN SGD is found as optimal, which is a slower yet more precise update strategy and is more suitable for gradient descent if the problem is difficult. 
ERNN takes into account a limited amount of past information, as the window in the BPTT procedure is set to a relatively small value.

Like ERNN, also for GRU the validation procedure identified SGD as the optimal gradient descent strategy. 
Interestingly, L\textsubscript{1} regularization is used, while in all the other cases it is not considered. 
On the other hand, the L\textsubscript{2} regularization parameter is much smaller.

In the optimal NARX configuration, TDL is set to a very small value. 
In particular, since the regression is performed only on the last 2 time intervals, the current output depends only on the most recent inputs and estimated outputs. 
From the number of hidden nodes and layers, we observe that the optimal size of the network is relatively small.

Relatively to the ESN configuration, we notice a very small spectral radius. This means that, also in this case, the network is configured with a small amount of memory.
This results in reservoir dynamics that are more responsive and, consequently, in outputs that mostly depend on the recent inputs.
As a consequence, the value of input scaling is small, since there is no necessity of quickly saturating the neurons activation.

\begin{figure}[!ht]
  \centering
  \includegraphics[width=\textwidth,keepaspectratio]{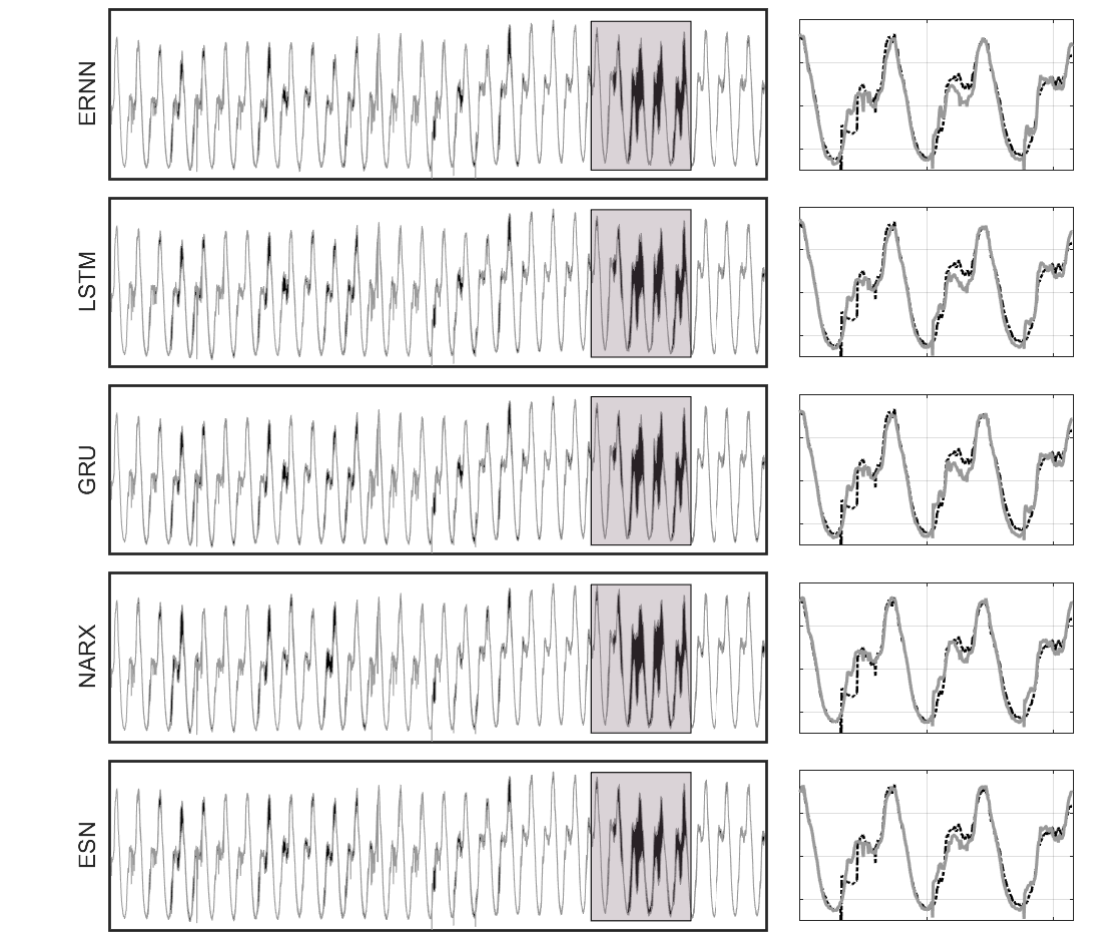}
  \caption{ACEA dataset -- The plots on the left show the residuals of predictions of each RNN with respect to the ground truth; black areas indicate the errors in the predictions. The plots on right depict a magnification of the area in the gray boxes from the left graphics; the dashed black line is the ground truth, the solid gray line is the prediction of each RNN.}
  \label{fig:acea_pred}
\end{figure}

\paragraph{ACEA}
The time series of the electricity load is quite regular except for few, erratic fluctuations. 
As for the Orange dataset, RNN predictions are inaccurate mainly in correspondence of such fluctuations, while they output a correct prediction elsewhere. 
This behavior is outlined by the plots in Fig.~\ref{fig:acea_pred}, where we observe that the residuals are small and, in each RNN prediction, they are mostly localized in common time intervals.
From the NRMSE values in Fig.~\ref{fig:real_res}, we see that ESN performs better than the other networks.
The worst performance is obtained by NARX, while the gradient-based RNNs yield better results, which are very similar to each other.

In ERNN and GRU, the optimal regularization found is the L\textsubscript{2} norm, whose coefficient assumes a small value.
In LSTM instead, beside the L\textsubscript{2} regularization term, the optimal configuration includes also a dropout regularization with a small probability.
The BPTT windows have comparable size in all the gradient-based networks.

The optimal NARX configuration for ACEA is very similar to the one identified for Orange and is characterized by a low complexity in terms of number of hidden nodes and layers. 
Also in this case the TDLs are very short.

Similarly to the optimal configuration for Orange, the ESN spectral radius assumes a small value, meaning that the network is equipped with a short-term memory and it captures only short temporal correlations in the data. 
The reservoir is configured with a high connectivity, which yields more homogeneous internal dynamics.

\begin{figure}[!ht]
  \centering
  \includegraphics[width=\textwidth,keepaspectratio]{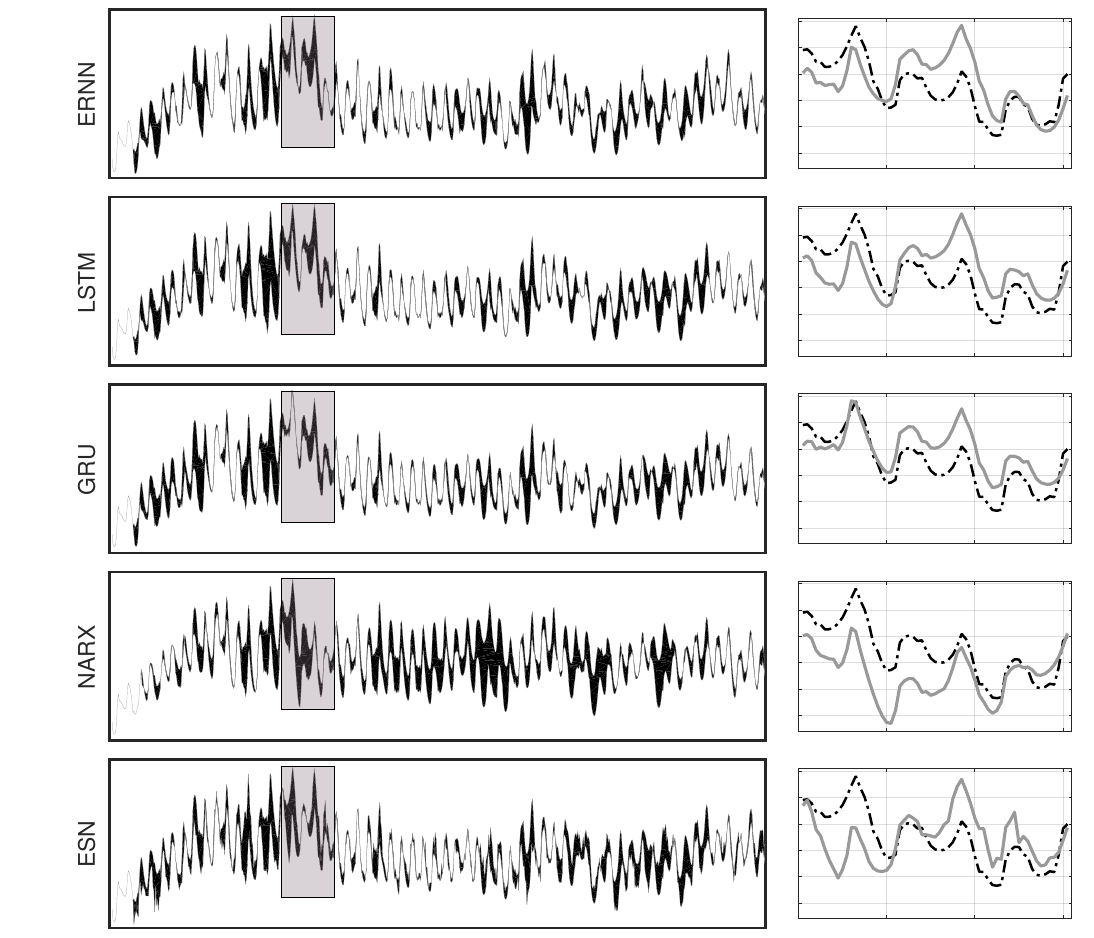}
  \caption{GEFCom dataset -- The plots on the left show the residuals of predictions of each RNN with respect to the ground truth; black areas indicate the errors in the predictions. The plots on right depict a magnification of the area in the gray boxes from the left graphics; the dashed black line is the ground truth, the solid gray line is the prediction of each RNN.}
  \label{fig:gefcom_pred}
\end{figure}

\paragraph{GEFCom}
This time series is more irregular than the previous ones, as it shows a more noisy behavior that is harder to be predicted.
From Fig.~\ref{fig:gefcom_pred} we see that the extent of the black areas of the residual is much larger than in the other datasets, denoting a higher prediction error.
From the third panel in Fig.~\ref{fig:real_res} we observe larger differences in the results with respect to the previous cases.
In this dataset, the exogenous time series of temperature plays a key role in the prediction, as it conveys information that are particularly helpful to yield a high accuracy. 
The main reason of the discrepancy in the results for the different networks may be in their capability of correctly leveraging this exogenous information for building an accurate forecast model. 

From the results, we observe that the gradient-based RNNs yield the best prediction accuracy. 
In particular, ERNN and GRU generate a prediction with the lowest NRMSE with respect to the target signal.
ESN, instead, obtains considerably lower performance.
Like for the syntethic datasets NARMA and MSO, NARX produces a very inaccurate prediction, scoring a NRMSE which is above 1.

The optimal ERNN configuration consists of only 60 nodes.

For LSTM, the optimal configuration includes only 20 hidden units, which is the lowest amount admitted in the validation search and SGD is the best as optimizer.

The optimal configuration for GRU is characterized by a large BPTT window, which assumes the maximum value allowed. This means that the network benefits from considering a large amount of past values to compute the prediction. 
As in LSTM, the number of processing units is very low. 
The best optimizer is Adam initialized with a particularly small learning rate, which yields a slower but more precise gradient update. 

The optimal configuration of NARX network is characterized by a quite large number of hidden nodes and layers, which denote a network of higher complexity with respect to the ones identified in the other tasks. 
This can be related to the TDL larger values, which require to be processed by a network with greater modeling capabilities.

For ESN, we notice an extremely large spectral radius, close to the maximum value admitted in the random search. Consequently, also the value of the input scaling is set to a high number, to increase the amount of nonlinerarity in the processing units. 
The output scaling is set close to 1, meaning that the teacher signal is almost unchanged when fed into the training procedure. 
A feedback scaling close to zero means that the feedback is almost disabled and it is not used by the ESN to update its internal state.

\section{Conclusions}
\label{sec:conclusions}

In this paper we studied the application of recurrent neural networks to time series prediction, focusing on the problem of short term load forecasting.
We reviewed five different architectures, ERNN, LSTM, GRU, NARX, and ESN, explaining their internal mechanisms, discussing their properties and the procedures for the training.
We performed a comparative analysis of the prediction performance obtained by the different networks on several time series, considering both synthetic benchmarks and real-world short term forecast problems.
For each network, we outlined the scheme we followed for the optimization of its hyperparameters.
Relative to the real-world problems, we discussed how to preprocess the data according to a detailed analysis of the time series.
We completed our analysis by comparing the performance of the RNNs on each task and discussing their optimal configurations.

From our experiments we can draw the following important conclusions.

There is not a specific RNN model that outperforms the others in every prediction problem. 
The choice of the most suitable architecture depends on the specific task at hand and it is important to consider more training strategies and configurations for each RNN.
On average, the NARX network achieved the lowest performance, especially on synthetic problems NARMA and MSO, and on the GEFCom dataset.

The training of gradient-based networks (ERNN, LSTM and GRU) is slower and in general more complex, due to the unfolding and backpropagation through time procedure. 
However, while some precautions need to be taken in the design of these networks, satisfactory results can be obtained with minimal fine-tuning and by selecting default hyperparameters. 
This implies that a strong expertise on the data domain is not always necessary.

The results obtained by the ESN are competitive in most tasks and the simplicity of its implementation makes it an appealing instrument for time series prediction.
ESN is characterized by a faster training procedure, but the performance heavily depends on the hyperparameters. 
Therefore, to identify the optimal configuration in the validation phase, ESN requires a search procedure of the hyperparameters that is more accurate than in gradient-based models. 

Another important aspect highlighted by our results is that the gated RNNs (LSTM and GRU) did not perform particularly better than an ERNN, whose architecture is much simpler, as well as its training.
While LSTM and GRU achieve outstanding results in many sequence learning problems, the additional complexity of the complicated gated mechanisms seems to be unnecessary in many time series predictions tasks.

We hypothesize as a possible explanation that in sequence learning problems, such as the ones of Natural Language Processing \cite{pmlr-v48-kumar16}, the temporal dependencies are more irregular than in the dynamical systems underlying the load time series.
In natural language for example, the dependency from a past word can persist for a long time period and then terminate abruptly when a sentence ends.
Moreover, there could exist relations between very localized chunks of the sequence. 
In this case, the RNN should focus on a specific temporal segment.

LSTM and GRU can efficiently model these highly nonlinear statistical dependencies, since their gating mechanisms allow to quickly modify the memory content of the cells and the internal dynamics. 
On the other hand, traditional RNNs implement smoother transfer functions and they would require a much larger complexity (number of units) to approximate such nonlinearities.
However, in dynamical systems with dependencies that decay smoothly over time, the features of the gates may not be necessary and a simple RNN could be more suitable for the task. 

Therefore, we conclude by arguing that ERNN and ESN may represent the most convenient choice in time series prediction problems, both in terms of performance and simplicity of their implementation and training.

\clearpage
\bibliographystyle{elsarticle-num-names}
\bibliography{Biblio}

\begin{thebibliography}{182}
\providecommand{\natexlab}[1]{#1}
\providecommand{\url}[1]{\texttt{#1}}
\providecommand{\urlprefix}{URL }
\expandafter\ifx\csname urlstyle\endcsname\relax
  \providecommand{\doi}[1]{doi:\discretionary{}{}{}#1}\else
  \providecommand{\doi}[1]{doi:\discretionary{}{}{}\begingroup
  \urlstyle{rm}\url{#1}\endgroup}\fi
\providecommand{\bibinfo}[2]{#2}

\bibitem[{Gooijer and Hyndman(2006)}]{DeGooijer2006443}
\bibinfo{author}{J.~G.~D. Gooijer}, \bibinfo{author}{R.~J. Hyndman},
  \bibinfo{title}{25 years of time series forecasting},
  \bibinfo{journal}{International Journal of Forecasting}
  \bibinfo{volume}{22}~(\bibinfo{number}{3}) (\bibinfo{year}{2006})
  \bibinfo{pages}{443 -- 473}, ISSN \bibinfo{issn}{0169-2070},
  \urlprefix\url{http://doi.org/10.1016/j.ijforecast.2006.01.001},
  \bibinfo{note}{twenty five years of forecasting}.

\bibitem[{Simchi-Levi et~al.(1999)Simchi-Levi, Simchi-Levi, and
  Kaminsky}]{simchi1999designing}
\bibinfo{author}{D.~Simchi-Levi}, \bibinfo{author}{E.~Simchi-Levi},
  \bibinfo{author}{P.~Kaminsky}, \bibinfo{title}{Designing and managing the
  supply chain: Concepts, strategies, and cases},
  \bibinfo{publisher}{McGraw-Hill New York}, \bibinfo{year}{1999}.

\bibitem[{Bunn(2000)}]{bunn2000forecasting}
\bibinfo{author}{D.~W. Bunn}, \bibinfo{title}{{F}orecasting loads and prices in
  competitive power markets}, \bibinfo{journal}{{P}roceedings of the {IEEE}}
  \bibinfo{volume}{88}~(\bibinfo{number}{2}).

\bibitem[{Ruiz and Gross(2008)}]{ruiz2008short}
\bibinfo{author}{P.~A. Ruiz}, \bibinfo{author}{G.~Gross},
  \bibinfo{title}{{S}hort-term resource adequacy in electricity market design},
  \bibinfo{journal}{{IEEE} {T}ransactions on {P}ower {S}ystems}
  \bibinfo{volume}{23}~(\bibinfo{number}{3}) (\bibinfo{year}{2008})
  \bibinfo{pages}{916--926}.

\bibitem[{Deihimi and Showkati(2012)}]{deihimi2012application}
\bibinfo{author}{A.~Deihimi}, \bibinfo{author}{H.~Showkati},
  \bibinfo{title}{{A}pplication of echo state networks in short-term electric
  load forecasting}, \bibinfo{journal}{{E}nergy}
  \bibinfo{volume}{39}~(\bibinfo{number}{1}) (\bibinfo{year}{2012})
  \bibinfo{pages}{327--340}.

\bibitem[{Peng et~al.(2014)Peng, Lei, Li, and Peng}]{peng2014novel}
\bibinfo{author}{Y.~Peng}, \bibinfo{author}{M.~Lei}, \bibinfo{author}{J.-B.
  Li}, \bibinfo{author}{X.-Y. Peng}, \bibinfo{title}{{A} novel hybridization of
  echo state networks and multiplicative seasonal {ARIMA} model for mobile
  communication traffic series forecasting}, \bibinfo{journal}{{N}eural
  {C}omputing and {A}pplications} \bibinfo{volume}{24}~(\bibinfo{number}{3-4})
  (\bibinfo{year}{2014}) \bibinfo{pages}{883--890}.

\bibitem[{Shen and Huang(2008)}]{shen2008interday}
\bibinfo{author}{H.~Shen}, \bibinfo{author}{J.~Z. Huang},
  \bibinfo{title}{{I}nterday forecasting and intraday updating of call center
  arrivals}, \bibinfo{journal}{{M}anufacturing \& {S}ervice {O}perations
  {M}anagement} \bibinfo{volume}{10}~(\bibinfo{number}{3})
  (\bibinfo{year}{2008}) \bibinfo{pages}{391--410}.

\bibitem[{Bianchi et~al.(2015{\natexlab{a}})Bianchi, Scardapane, Uncini, Rizzi,
  and Sadeghian}]{bianchi2015prediction}
\bibinfo{author}{F.~M. Bianchi}, \bibinfo{author}{S.~Scardapane},
  \bibinfo{author}{A.~Uncini}, \bibinfo{author}{A.~Rizzi},
  \bibinfo{author}{A.~Sadeghian}, \bibinfo{title}{Prediction of telephone calls
  load using {E}cho {S}tate {N}etwork with exogenous variables},
  \bibinfo{journal}{Neural Networks} \bibinfo{volume}{71}
  (\bibinfo{year}{2015}{\natexlab{a}}) \bibinfo{pages}{204--213},
  \urlprefix\url{10.1016/j.neunet.2015.08.010}.

\bibitem[{Bianchi et~al.(2015{\natexlab{b}})Bianchi, De~Santis, Rizzi, and
  Sadeghian}]{7286732}
\bibinfo{author}{F.~M. Bianchi}, \bibinfo{author}{E.~De~Santis},
  \bibinfo{author}{A.~Rizzi}, \bibinfo{author}{A.~Sadeghian},
  \bibinfo{title}{Short-term electric load forecasting using echo state
  networks and {PCA} decomposition}, \bibinfo{journal}{IEEE Access}
  \bibinfo{volume}{3} (\bibinfo{year}{2015}{\natexlab{b}})
  \bibinfo{pages}{1931--1943}, ISSN \bibinfo{issn}{2169-3536},
  \urlprefix\url{10.1109/ACCESS.2015.2485943}.

\bibitem[{Deihimi et~al.(2013)Deihimi, Orang, and Showkati}]{deihimi2013short}
\bibinfo{author}{A.~Deihimi}, \bibinfo{author}{O.~Orang},
  \bibinfo{author}{H.~Showkati}, \bibinfo{title}{Short-term electric load and
  temperature forecasting using wavelet echo state networks with neural
  reconstruction}, \bibinfo{journal}{Energy} \bibinfo{volume}{57}
  (\bibinfo{year}{2013}) \bibinfo{pages}{382--401}.

\bibitem[{Jan~van Oldenborgh et~al.(2005)Jan~van Oldenborgh, Balmaseda,
  Ferranti, Stockdale, and Anderson}]{jan2005did}
\bibinfo{author}{G.~Jan~van Oldenborgh}, \bibinfo{author}{M.~A. Balmaseda},
  \bibinfo{author}{L.~Ferranti}, \bibinfo{author}{T.~N. Stockdale},
  \bibinfo{author}{D.~L. Anderson}, \bibinfo{title}{Did the ECMWF seasonal
  forecast model outperform statistical ENSO forecast models over the last 15
  years?}, \bibinfo{journal}{Journal of climate}
  \bibinfo{volume}{18}~(\bibinfo{number}{16}) (\bibinfo{year}{2005})
  \bibinfo{pages}{3240--3249}.

\bibitem[{{Dang-Ha} et~al.(2017){Dang-Ha}, {Bianchi}, and
  {Olsson}}]{2017arXiv170208025D}
\bibinfo{author}{T.-H. {Dang-Ha}}, \bibinfo{author}{F.~M. {Bianchi}},
  \bibinfo{author}{R.~{Olsson}}, \bibinfo{title}{{Local Short Term Electricity
  Load Forecasting: Automatic Approaches}}, \bibinfo{journal}{ArXiv e-prints} .

\bibitem[{Hyndman et~al.(2008)Hyndman, Koehler, Ord, and
  Snyder}]{hyndman2008forecasting}
\bibinfo{author}{R.~Hyndman}, \bibinfo{author}{A.~B. Koehler},
  \bibinfo{author}{J.~K. Ord}, \bibinfo{author}{R.~D. Snyder},
  \bibinfo{title}{Forecasting with exponential smoothing: the state space
  approach}, \bibinfo{publisher}{Springer Science \& Business Media}, ISBN
  \bibinfo{isbn}{9783540719182}, \bibinfo{year}{2008}.

\bibitem[{Box et~al.(2011)Box, Jenkins, and Reinsel}]{box2011time}
\bibinfo{author}{G.~E. Box}, \bibinfo{author}{G.~M. Jenkins},
  \bibinfo{author}{G.~C. Reinsel}, \bibinfo{title}{Time series analysis:
  forecasting and control}, vol. \bibinfo{volume}{734},
  \bibinfo{publisher}{John Wiley \& Sons}, \bibinfo{year}{2011}.

\bibitem[{Box and Cox(1964)}]{box1964analysis}
\bibinfo{author}{G.~E. Box}, \bibinfo{author}{D.~R. Cox}, \bibinfo{title}{An
  analysis of transformations}, \bibinfo{journal}{Journal of the Royal
  Statistical Society. Series B (Methodological)}  (\bibinfo{year}{1964})
  \bibinfo{pages}{211--252}.

\bibitem[{Taylor(2008)}]{taylor2008comparison}
\bibinfo{author}{J.~W. Taylor}, \bibinfo{title}{{A} comparison of univariate
  time series methods for forecasting intraday arrivals at a call center},
  \bibinfo{journal}{{M}anagement {S}cience}
  \bibinfo{volume}{54}~(\bibinfo{number}{2}) (\bibinfo{year}{2008})
  \bibinfo{pages}{253--265}.

\bibitem[{Takens(1981)}]{takens1981detecting}
\bibinfo{author}{F.~Takens}, \bibinfo{title}{Detecting strange attractors in
  turbulence}, \bibinfo{publisher}{Springer}, \bibinfo{year}{1981}.

\bibitem[{Sapankevych and Sankar(2009)}]{sapankevych2009time}
\bibinfo{author}{N.~I. Sapankevych}, \bibinfo{author}{R.~Sankar},
  \bibinfo{title}{Time series prediction using support vector machines: a
  survey}, \bibinfo{journal}{Computational Intelligence Magazine, IEEE}
  \bibinfo{volume}{4}~(\bibinfo{number}{2}) (\bibinfo{year}{2009})
  \bibinfo{pages}{24--38}.

\bibitem[{Hornik et~al.(1989)Hornik, Stinchcombe, and White}]{Hornik1989359}
\bibinfo{author}{K.~Hornik}, \bibinfo{author}{M.~Stinchcombe},
  \bibinfo{author}{H.~White}, \bibinfo{title}{Multilayer feedforward networks
  are universal approximators}, \bibinfo{journal}{Neural Networks}
  \bibinfo{volume}{2}~(\bibinfo{number}{5}) (\bibinfo{year}{1989})
  \bibinfo{pages}{359 -- 366}, ISSN \bibinfo{issn}{0893-6080},
  \urlprefix\url{http://dx.doi.org/10.1016/0893-6080(89)90020-8}.

\bibitem[{Jang(1993)}]{jang1993anfis}
\bibinfo{author}{J.-S.~R. Jang}, \bibinfo{title}{ANFIS: adaptive-network-based
  fuzzy inference system}, \bibinfo{journal}{Systems, Man and Cybernetics, IEEE
  Transactions on} \bibinfo{volume}{23}~(\bibinfo{number}{3})
  (\bibinfo{year}{1993}) \bibinfo{pages}{665--685}.

\bibitem[{Zhang et~al.(1998)Zhang, Patuwo, and Hu}]{Zhang199835}
\bibinfo{author}{G.~Zhang}, \bibinfo{author}{B.~E. Patuwo},
  \bibinfo{author}{M.~Y. Hu}, \bibinfo{title}{Forecasting with artificial
  neural networks:: The state of the art}, \bibinfo{journal}{International
  Journal of Forecasting} \bibinfo{volume}{14}~(\bibinfo{number}{1})
  (\bibinfo{year}{1998}) \bibinfo{pages}{35 -- 62}, ISSN
  \bibinfo{issn}{0169-2070},
  \urlprefix\url{http://doi.org/10.1016/S0169-2070(97)00044-7}.

\bibitem[{Hippert et~al.(2001)Hippert, Pedreira, and Souza}]{Hippert2001}
\bibinfo{author}{H.~Hippert}, \bibinfo{author}{C.~Pedreira},
  \bibinfo{author}{R.~Souza}, \bibinfo{title}{{Neural networks for short-term
  load forecasting: a review and evaluation}}, \bibinfo{journal}{IEEE
  Transactions on Power Systems} \bibinfo{volume}{16}~(\bibinfo{number}{1})
  (\bibinfo{year}{2001}) \bibinfo{pages}{44--55}, ISSN
  \bibinfo{issn}{08858950}, \urlprefix\url{10.1109/59.910780}.

\bibitem[{Law(2000)}]{Law2000331}
\bibinfo{author}{R.~Law}, \bibinfo{title}{Back-propagation learning in
  improving the accuracy of neural network-based tourism demand forecasting},
  \bibinfo{journal}{Tourism Management}
  \bibinfo{volume}{21}~(\bibinfo{number}{4}) (\bibinfo{year}{2000})
  \bibinfo{pages}{331 -- 340}, ISSN \bibinfo{issn}{0261-5177},
  \urlprefix\url{http://doi.org/10.1016/S0261-5177(99)00067-9}.

\bibitem[{Tsaur et~al.(2002)Tsaur, Chiu, and Huang}]{Tsaur2002397}
\bibinfo{author}{S.-H. Tsaur}, \bibinfo{author}{Y.-C. Chiu},
  \bibinfo{author}{C.-H. Huang}, \bibinfo{title}{Determinants of guest loyalty
  to international tourist hotels—a neural network approach},
  \bibinfo{journal}{Tourism Management}
  \bibinfo{volume}{23}~(\bibinfo{number}{4}) (\bibinfo{year}{2002})
  \bibinfo{pages}{397 -- 405}, ISSN \bibinfo{issn}{0261-5177},
  \urlprefix\url{http://doi.org/10.1016/S0261-5177(01)00097-8}.

\bibitem[{Kon and Turner(2005)}]{kon2005neural}
\bibinfo{author}{S.~C. Kon}, \bibinfo{author}{L.~W. Turner},
  \bibinfo{title}{Neural network forecasting of tourism demand},
  \bibinfo{journal}{Tourism Economics}
  \bibinfo{volume}{11}~(\bibinfo{number}{3}) (\bibinfo{year}{2005})
  \bibinfo{pages}{301--328},
  \urlprefix\url{http://dx.doi.org/10.5367/000000005774353006}.

\bibitem[{Palmer et~al.(2006)Palmer, Montaño, and Sesé}]{Palmer2006781}
\bibinfo{author}{A.~Palmer}, \bibinfo{author}{J.~J. Montaño},
  \bibinfo{author}{A.~Sesé}, \bibinfo{title}{Designing an artificial neural
  network for forecasting tourism time series}, \bibinfo{journal}{Tourism
  Management} \bibinfo{volume}{27}~(\bibinfo{number}{5}) (\bibinfo{year}{2006})
  \bibinfo{pages}{781 -- 790}, ISSN \bibinfo{issn}{0261-5177},
  \urlprefix\url{http://doi.org/10.1016/j.tourman.2005.05.006}.

\bibitem[{Claveria and Torra(2014)}]{Claveria2014220}
\bibinfo{author}{O.~Claveria}, \bibinfo{author}{S.~Torra},
  \bibinfo{title}{Forecasting tourism demand to Catalonia: Neural networks vs.
  time series models}, \bibinfo{journal}{Economic Modelling}
  \bibinfo{volume}{36} (\bibinfo{year}{2014}) \bibinfo{pages}{220 -- 228}, ISSN
  \bibinfo{issn}{0264-9993},
  \urlprefix\url{http://doi.org/10.1016/j.econmod.2013.09.024}.

\bibitem[{Kourentzes(2013)}]{Kourentzes2013198}
\bibinfo{author}{N.~Kourentzes}, \bibinfo{title}{Intermittent demand forecasts
  with neural networks}, \bibinfo{journal}{International Journal of Production
  Economics} \bibinfo{volume}{143}~(\bibinfo{number}{1}) (\bibinfo{year}{2013})
  \bibinfo{pages}{198 -- 206}, ISSN \bibinfo{issn}{0925-5273},
  \urlprefix\url{http://doi.org/10.1016/j.ijpe.2013.01.009}.

\bibitem[{Díaz-Robles et~al.(2008)Díaz-Robles, Ortega, Fu, Reed, Chow,
  Watson, and Moncada-Herrera}]{DiazRobles20088331}
\bibinfo{author}{L.~A. Díaz-Robles}, \bibinfo{author}{J.~C. Ortega},
  \bibinfo{author}{J.~S. Fu}, \bibinfo{author}{G.~D. Reed},
  \bibinfo{author}{J.~C. Chow}, \bibinfo{author}{J.~G. Watson},
  \bibinfo{author}{J.~A. Moncada-Herrera}, \bibinfo{title}{A hybrid \{ARIMA\}
  and artificial neural networks model to forecast particulate matter in urban
  areas: The case of Temuco, Chile}, \bibinfo{journal}{Atmospheric Environment}
  \bibinfo{volume}{42}~(\bibinfo{number}{35}) (\bibinfo{year}{2008})
  \bibinfo{pages}{8331 -- 8340}, ISSN \bibinfo{issn}{1352-2310},
  \urlprefix\url{http://doi.org/10.1016/j.atmosenv.2008.07.020}.

\bibitem[{Plummer(2000)}]{plummer2000time}
\bibinfo{author}{E.~Plummer}, \bibinfo{title}{Time series forecasting with
  feed-forward neural networks: guidelines and limitations},
  \bibinfo{journal}{Neural Networks} \bibinfo{volume}{1} (\bibinfo{year}{2000})
  \bibinfo{pages}{1}.

\bibitem[{Teixeira and Fernandes(2012)}]{TEIXEIRA2012445}
\bibinfo{author}{J.~P. Teixeira}, \bibinfo{author}{P.~O. Fernandes},
  \bibinfo{title}{Tourism Time Series Forecast -Different ANN Architectures
  with Time Index Input}, \bibinfo{journal}{Procedia Technology}
  \bibinfo{volume}{5} (\bibinfo{year}{2012}) \bibinfo{pages}{445 -- 454}, ISSN
  \bibinfo{issn}{2212-0173},
  \urlprefix\url{http://dx.doi.org/10.1016/j.protcy.2012.09.049}.

\bibitem[{Claveria et~al.(2015)Claveria, Monte, and Torra}]{JTR:JTR2016}
\bibinfo{author}{O.~Claveria}, \bibinfo{author}{E.~Monte},
  \bibinfo{author}{S.~Torra}, \bibinfo{title}{Tourism Demand Forecasting with
  Neural Network Models: Different Ways of Treating Information},
  \bibinfo{journal}{International Journal of Tourism Research}
  \bibinfo{volume}{17}~(\bibinfo{number}{5}) (\bibinfo{year}{2015})
  \bibinfo{pages}{492--500}, ISSN \bibinfo{issn}{1522-1970},
  \urlprefix\url{10.1002/jtr.2016}, \bibinfo{note}{jTR-13-0416.R2}.

\bibitem[{Sch{\"a}fer and Zimmermann(2007)}]{schafer2007approximators}
\bibinfo{author}{A.~M. Sch{\"a}fer}, \bibinfo{author}{H.-G. Zimmermann},
  \bibinfo{title}{Recurrent Neural Networks are Universal Approximators},
  \bibinfo{journal}{International Journal of Neural Systems}
  \bibinfo{volume}{17}~(\bibinfo{number}{04}) (\bibinfo{year}{2007})
  \bibinfo{pages}{253--263}, \urlprefix\url{10.1142/S0129065707001111}.

\bibitem[{Bianchi et~al.(2017)Bianchi, Kampffmeyer, Maiorino, and
  Jenssen}]{bianchi2017temporal}
\bibinfo{author}{F.~M. Bianchi}, \bibinfo{author}{M.~Kampffmeyer},
  \bibinfo{author}{E.~Maiorino}, \bibinfo{author}{R.~Jenssen},
  \bibinfo{title}{Temporal Overdrive Recurrent Neural Network},
  \bibinfo{journal}{arXiv preprint arXiv:1701.05159} .

\bibitem[{Sch\"{a}fer and Zimmermann(2007)}]{Schafer2007}
\bibinfo{author}{A.~M. Sch\"{a}fer}, \bibinfo{author}{H.-G. Zimmermann},
  \bibinfo{title}{{Recurrent Neural Networks are universal approximators.}},
  \bibinfo{journal}{International journal of neural systems}
  \bibinfo{volume}{17}~(\bibinfo{number}{4}) (\bibinfo{year}{2007})
  \bibinfo{pages}{253--263}, ISSN \bibinfo{issn}{0129-0657},
  \urlprefix\url{10.1142/S0129065707001111}.

\bibitem[{Graves(2012)}]{graves2012sequence}
\bibinfo{author}{A.~Graves}, \bibinfo{title}{Sequence transduction with
  recurrent neural networks}, \bibinfo{journal}{arXiv preprint arXiv:1211.3711}
  .

\bibitem[{Graves(2013)}]{Graves2013}
\bibinfo{author}{A.~Graves}, \bibinfo{title}{{Generating sequences with
  recurrent neural networks}}, \bibinfo{journal}{arXiv preprint
  arXiv:1308.0850}  (\bibinfo{year}{2013}) \bibinfo{pages}{1--43}.

\bibitem[{Pascanu et~al.(2013{\natexlab{a}})Pascanu, Mikolov, and
  Bengio}]{Pascanu_Mikolov_Bengio}
\bibinfo{author}{R.~Pascanu}, \bibinfo{author}{T.~Mikolov},
  \bibinfo{author}{Y.~Bengio}, \bibinfo{title}{On the Difficulty of Training
  Recurrent Neural Networks}, in: \bibinfo{booktitle}{Proceedings of the 30th
  International Conference on International Conference on Machine Learning -
  Volume 28}, ICML'13, \bibinfo{publisher}{JMLR.org},
  \bibinfo{pages}{III--1310--III--1318},
  \urlprefix\url{http://dl.acm.org/citation.cfm?id=3042817.3043083},
  \bibinfo{year}{2013}{\natexlab{a}}.

\bibitem[{Mikolov(2012)}]{tomavs2012statistical}
\bibinfo{author}{T.~Mikolov}, \bibinfo{title}{Statistical language models based
  on neural networks}, Ph.D. thesis, \bibinfo{school}{PhD thesis, Brno
  University of Technology. 2012.[PDF]}, \bibinfo{year}{2012}.

\bibitem[{Sutskever et~al.(2011)Sutskever, Martens, and
  Hinton}]{sutskever2011generating}
\bibinfo{author}{I.~Sutskever}, \bibinfo{author}{J.~Martens},
  \bibinfo{author}{G.~E. Hinton}, \bibinfo{title}{Generating text with
  recurrent neural networks}, in: \bibinfo{booktitle}{Proceedings of the 28th
  International Conference on Machine Learning (ICML-11)},
  \bibinfo{pages}{1017--1024}, \bibinfo{year}{2011}.

\bibitem[{Graves(2011)}]{graves2011practical}
\bibinfo{author}{A.~Graves}, \bibinfo{title}{Practical variational inference
  for neural networks}, in: \bibinfo{booktitle}{Advances in Neural Information
  Processing Systems}, \bibinfo{pages}{2348--2356}, \bibinfo{year}{2011}.

\bibitem[{Mikolov et~al.(2013)Mikolov, Sutskever, Chen, Corrado, and
  Dean}]{mikolov2013distributed}
\bibinfo{author}{T.~Mikolov}, \bibinfo{author}{I.~Sutskever},
  \bibinfo{author}{K.~Chen}, \bibinfo{author}{G.~S. Corrado},
  \bibinfo{author}{J.~Dean}, \bibinfo{title}{Distributed representations of
  words and phrases and their compositionality}, in:
  \bibinfo{booktitle}{Advances in neural information processing systems},
  \bibinfo{pages}{3111--3119}, \bibinfo{year}{2013}.

\bibitem[{Oord et~al.(2016)Oord, Dieleman, Zen, Simonyan, Vinyals, Graves,
  Kalchbrenner, Senior, and Kavukcuoglu}]{oord2016wavenet}
\bibinfo{author}{A.~v.~d. Oord}, \bibinfo{author}{S.~Dieleman},
  \bibinfo{author}{H.~Zen}, \bibinfo{author}{K.~Simonyan},
  \bibinfo{author}{O.~Vinyals}, \bibinfo{author}{A.~Graves},
  \bibinfo{author}{N.~Kalchbrenner}, \bibinfo{author}{A.~Senior},
  \bibinfo{author}{K.~Kavukcuoglu}, \bibinfo{title}{Wavenet: A generative model
  for raw audio}, \bibinfo{journal}{arXiv preprint arXiv:1609.03499} .

\bibitem[{Graves and Schmidhuber(2009)}]{graves2009offline}
\bibinfo{author}{A.~Graves}, \bibinfo{author}{J.~Schmidhuber},
  \bibinfo{title}{Offline handwriting recognition with multidimensional
  recurrent neural networks}, in: \bibinfo{booktitle}{Advances in neural
  information processing systems}, \bibinfo{pages}{545--552},
  \bibinfo{year}{2009}.

\bibitem[{Graves et~al.(2008)Graves, Liwicki, Bunke, Schmidhuber, and
  Fern{\'a}ndez}]{graves2008unconstrained}
\bibinfo{author}{A.~Graves}, \bibinfo{author}{M.~Liwicki},
  \bibinfo{author}{H.~Bunke}, \bibinfo{author}{J.~Schmidhuber},
  \bibinfo{author}{S.~Fern{\'a}ndez}, \bibinfo{title}{Unconstrained on-line
  handwriting recognition with recurrent neural networks}, in:
  \bibinfo{booktitle}{Advances in Neural Information Processing Systems},
  \bibinfo{pages}{577--584}, \bibinfo{year}{2008}.

\bibitem[{Gregor et~al.(2015)Gregor, Danihelka, Graves, Rezende, and
  Wierstra}]{gregor2015draw}
\bibinfo{author}{K.~Gregor}, \bibinfo{author}{I.~Danihelka},
  \bibinfo{author}{A.~Graves}, \bibinfo{author}{D.~J. Rezende},
  \bibinfo{author}{D.~Wierstra}, \bibinfo{title}{DRAW: A recurrent neural
  network for image generation}, \bibinfo{journal}{arXiv preprint
  arXiv:1502.04623} .

\bibitem[{Hochreiter and Schmidhuber(1997)}]{hochreiter1997long}
\bibinfo{author}{S.~Hochreiter}, \bibinfo{author}{J.~Schmidhuber},
  \bibinfo{title}{Long short-term memory}, \bibinfo{journal}{Neural
  computation} \bibinfo{volume}{9}~(\bibinfo{number}{8}) (\bibinfo{year}{1997})
  \bibinfo{pages}{1735--1780}.

\bibitem[{Weston et~al.(2014)Weston, Chopra, and
  Bordes}]{DBLP:journals/corr/WestonCB14}
\bibinfo{author}{J.~Weston}, \bibinfo{author}{S.~Chopra},
  \bibinfo{author}{A.~Bordes}, \bibinfo{title}{Memory Networks},
  \bibinfo{journal}{CoRR} \bibinfo{volume}{abs/1410.3916}.

\bibitem[{Graves et~al.(2014)Graves, Wayne, and
  Danihelka}]{DBLP:journals/corr/GravesWD14}
\bibinfo{author}{A.~Graves}, \bibinfo{author}{G.~Wayne},
  \bibinfo{author}{I.~Danihelka}, \bibinfo{title}{Neural Turing Machines},
  \bibinfo{journal}{CoRR} \bibinfo{volume}{abs/1410.5401},
  \urlprefix\url{http://arxiv.org/abs/1410.5401}.

\bibitem[{Gers et~al.(2001)Gers, Eck, and Schmidhuber}]{Gers2001}
\bibinfo{author}{F.~A. Gers}, \bibinfo{author}{D.~Eck},
  \bibinfo{author}{J.~Schmidhuber}, \bibinfo{title}{Applying LSTM to Time
  Series Predictable through Time-Window Approaches}, in:
  \bibinfo{editor}{G.~Dorffner}, \bibinfo{editor}{H.~Bischof},
  \bibinfo{editor}{K.~Hornik} (Eds.), \bibinfo{booktitle}{Artificial Neural
  Networks --- ICANN 2001: International Conference Vienna, Austria, August
  21--25, 2001 Proceedings}, \bibinfo{publisher}{Springer Berlin Heidelberg},
  \bibinfo{address}{Berlin, Heidelberg}, ISBN
  \bibinfo{isbn}{978-3-540-44668-2}, \bibinfo{pages}{669--676},
  \urlprefix\url{10.1007/3-540-44668-0_93}, \bibinfo{year}{2001}.

\bibitem[{{Flunkert} et~al.(2017){Flunkert}, {Salinas}, and
  {Gasthaus}}]{2017arXiv170404110F}
\bibinfo{author}{V.~{Flunkert}}, \bibinfo{author}{D.~{Salinas}},
  \bibinfo{author}{J.~{Gasthaus}}, \bibinfo{title}{{DeepAR: Probabilistic
  Forecasting with Autoregressive Recurrent Networks}}, \bibinfo{journal}{ArXiv
  e-prints} .

\bibitem[{Greff et~al.(2015)Greff, Srivastava, Koutn{\'\i}k, Steunebrink, and
  Schmidhuber}]{greff2015lstm}
\bibinfo{author}{K.~Greff}, \bibinfo{author}{R.~K. Srivastava},
  \bibinfo{author}{J.~Koutn{\'\i}k}, \bibinfo{author}{B.~R. Steunebrink},
  \bibinfo{author}{J.~Schmidhuber}, \bibinfo{title}{LSTM: A search space
  odyssey}, \bibinfo{journal}{arXiv preprint arXiv:1503.04069} .

\bibitem[{Malhotra et~al.(2015)Malhotra, Vig, Shroff, and
  Agarwal}]{malhotra2015long}
\bibinfo{author}{P.~Malhotra}, \bibinfo{author}{L.~Vig},
  \bibinfo{author}{G.~Shroff}, \bibinfo{author}{P.~Agarwal},
  \bibinfo{title}{Long short term memory networks for anomaly detection in time
  series}, in: \bibinfo{booktitle}{Proceedings}, \bibinfo{organization}{Presses
  universitaires de Louvain}, \bibinfo{pages}{89}, \bibinfo{year}{2015}.

\bibitem[{Maass et~al.(2007)Maass, Joshi, and Sontag}]{maass2007computational}
\bibinfo{author}{W.~Maass}, \bibinfo{author}{P.~Joshi}, \bibinfo{author}{E.~D.
  Sontag}, \bibinfo{title}{Computational aspects of feedback in neural
  circuits}, \bibinfo{journal}{PLoS Computational Biology}
  \bibinfo{volume}{3}~(\bibinfo{number}{1}) (\bibinfo{year}{2007})
  \bibinfo{pages}{e165}, \urlprefix\url{10.1371/journal.pcbi.0020165.eor}.

\bibitem[{Siegelmann and Sontag(1991)}]{siegelmann1991turing}
\bibinfo{author}{H.~T. Siegelmann}, \bibinfo{author}{E.~D. Sontag},
  \bibinfo{title}{Turing computability with neural nets},
  \bibinfo{journal}{Applied Mathematics Letters}
  \bibinfo{volume}{4}~(\bibinfo{number}{6}) (\bibinfo{year}{1991})
  \bibinfo{pages}{77--80}.

\bibitem[{Schmidhuber et~al.(2007)Schmidhuber, Wierstra, Gagliolo, and
  Gomez}]{schmidhuber2007training}
\bibinfo{author}{J.~Schmidhuber}, \bibinfo{author}{D.~Wierstra},
  \bibinfo{author}{M.~Gagliolo}, \bibinfo{author}{F.~Gomez},
  \bibinfo{title}{Training recurrent networks by evolino},
  \bibinfo{journal}{Neural computation}
  \bibinfo{volume}{19}~(\bibinfo{number}{3}) (\bibinfo{year}{2007})
  \bibinfo{pages}{757--779}.

\bibitem[{Jaeger(2001)}]{jaeger2001echo}
\bibinfo{author}{H.~Jaeger}, \bibinfo{title}{The ``echo state'' approach to
  analysing and training recurrent neural networks-with an erratum note},
  \bibinfo{journal}{Bonn, Germany: German National Research Center for
  Information Technology GMD Technical Report} \bibinfo{volume}{148}
  (\bibinfo{year}{2001}) \bibinfo{pages}{34}.

\bibitem[{Zhang et~al.(2016)Zhang, Wu, Che, Lin, Memisevic, Salakhutdinov, and
  Bengio}]{NIPS2016_6303}
\bibinfo{author}{S.~Zhang}, \bibinfo{author}{Y.~Wu}, \bibinfo{author}{T.~Che},
  \bibinfo{author}{Z.~Lin}, \bibinfo{author}{R.~Memisevic},
  \bibinfo{author}{R.~R. Salakhutdinov}, \bibinfo{author}{Y.~Bengio},
  \bibinfo{title}{Architectural Complexity Measures of Recurrent Neural
  Networks}, in: \bibinfo{editor}{D.~D. Lee}, \bibinfo{editor}{M.~Sugiyama},
  \bibinfo{editor}{U.~V. Luxburg}, \bibinfo{editor}{I.~Guyon},
  \bibinfo{editor}{R.~Garnett} (Eds.), \bibinfo{booktitle}{Advances in Neural
  Information Processing Systems 29}, \bibinfo{publisher}{Curran Associates,
  Inc.}, \bibinfo{pages}{1822--1830}, \bibinfo{year}{2016}.

\bibitem[{Koutn{\'{\i}}k et~al.(2014)Koutn{\'{\i}}k, Greff, Gomez, and
  Schmidhuber}]{DBLP:journals/corr/KoutnikGGS14}
\bibinfo{author}{J.~Koutn{\'{\i}}k}, \bibinfo{author}{K.~Greff},
  \bibinfo{author}{F.~J. Gomez}, \bibinfo{author}{J.~Schmidhuber},
  \bibinfo{title}{A Clockwork {RNN}}, \bibinfo{journal}{CoRR}
  \bibinfo{volume}{abs/1402.3511}.

\bibitem[{Sutskever and Hinton(2010)}]{sutskever2010temporal}
\bibinfo{author}{I.~Sutskever}, \bibinfo{author}{G.~Hinton},
  \bibinfo{title}{Temporal-kernel recurrent neural networks},
  \bibinfo{journal}{Neural Networks} \bibinfo{volume}{23}~(\bibinfo{number}{2})
  (\bibinfo{year}{2010}) \bibinfo{pages}{239--243}.

\bibitem[{Schuster and Paliwal(1997)}]{schuster1997bidirectional}
\bibinfo{author}{M.~Schuster}, \bibinfo{author}{K.~K. Paliwal},
  \bibinfo{title}{Bidirectional recurrent neural networks},
  \bibinfo{journal}{Signal Processing, IEEE Transactions on}
  \bibinfo{volume}{45}~(\bibinfo{number}{11}) (\bibinfo{year}{1997})
  \bibinfo{pages}{2673--2681}.

\bibitem[{{Schoenholz} et~al.(2016){Schoenholz}, {Gilmer}, {Ganguli}, and
  {Sohl-Dickstein}}]{2016arXiv161101232S}
\bibinfo{author}{S.~S. {Schoenholz}}, \bibinfo{author}{J.~{Gilmer}},
  \bibinfo{author}{S.~{Ganguli}}, \bibinfo{author}{J.~{Sohl-Dickstein}},
  \bibinfo{title}{{Deep Information Propagation}}, \bibinfo{journal}{ArXiv
  e-prints} .

\bibitem[{Lipton(2015)}]{DBLP:journals/corr/Lipton15}
\bibinfo{author}{Z.~C. Lipton}, \bibinfo{title}{A Critical Review of Recurrent
  Neural Networks for Sequence Learning}, \bibinfo{journal}{CoRR}
  \bibinfo{volume}{abs/1506.00019},
  \urlprefix\url{http://arxiv.org/abs/1506.00019}.

\bibitem[{Montavon et~al.(2012)Montavon, Orr, and M\"{u}ller}]{Montavon2012}
\bibinfo{author}{G.~Montavon}, \bibinfo{author}{G.~Orr}, \bibinfo{author}{K.-R.
  M\"{u}ller}, \bibinfo{title}{{Neural networks-tricks of the trade second
  edition}}, \bibinfo{publisher}{Springer},
  \urlprefix\url{10.1007/978-3-642-35289-8}, \bibinfo{year}{2012}.

\bibitem[{Scardapane et~al.(2017)Scardapane, Comminiello, Hussain, and
  Uncini}]{Scardapane201781}
\bibinfo{author}{S.~Scardapane}, \bibinfo{author}{D.~Comminiello},
  \bibinfo{author}{A.~Hussain}, \bibinfo{author}{A.~Uncini},
  \bibinfo{title}{Group sparse regularization for deep neural networks},
  \bibinfo{journal}{Neurocomputing} \bibinfo{volume}{241}
  (\bibinfo{year}{2017}) \bibinfo{pages}{81 -- 89}, ISSN
  \bibinfo{issn}{0925-2312},
  \urlprefix\url{https://doi.org/10.1016/j.neucom.2017.02.029}.

\bibitem[{Scardapane and Wang(2017)}]{WIDM:WIDM1200}
\bibinfo{author}{S.~Scardapane}, \bibinfo{author}{D.~Wang},
  \bibinfo{title}{Randomness in neural networks: an overview},
  \bibinfo{journal}{Wiley Interdisciplinary Reviews: Data Mining and Knowledge
  Discovery} \bibinfo{volume}{7}~(\bibinfo{number}{2}) (\bibinfo{year}{2017})
  \bibinfo{pages}{e1200--n/a}, ISSN \bibinfo{issn}{1942-4795},
  \urlprefix\url{10.1002/widm.1200}, \bibinfo{note}{e1200}.

\bibitem[{Jaeger(2002{\natexlab{a}})}]{jaeger2002tutorial}
\bibinfo{author}{H.~Jaeger}, \bibinfo{title}{Tutorial on training recurrent
  neural networks, covering BPPT, RTRL, EKF and the" echo state network"
  approach}, vol.~\bibinfo{volume}{5},
  \bibinfo{publisher}{GMD-Forschungszentrum Informationstechnik},
  \bibinfo{year}{2002}{\natexlab{a}}.

\bibitem[{Williams and Zipser(1989)}]{williams1989learning}
\bibinfo{author}{R.~J. Williams}, \bibinfo{author}{D.~Zipser},
  \bibinfo{title}{A learning algorithm for continually running fully recurrent
  neural networks}, \bibinfo{journal}{Neural computation}
  \bibinfo{volume}{1}~(\bibinfo{number}{2}) (\bibinfo{year}{1989})
  \bibinfo{pages}{270--280}.

\bibitem[{Haykin et~al.(2001)Haykin, Haykin, and Haykin}]{haykin2001kalman}
\bibinfo{author}{S.~S. Haykin}, \bibinfo{author}{S.~S. Haykin},
  \bibinfo{author}{S.~S. Haykin}, \bibinfo{title}{Kalman filtering and neural
  networks}, \bibinfo{publisher}{Wiley Online Library}, \bibinfo{year}{2001}.

\bibitem[{John(1992)}]{john1992holland}
\bibinfo{author}{H.~John}, \bibinfo{title}{Holland, Adaptation in natural and
  artificial systems}, \bibinfo{year}{1992}.

\bibitem[{Luko{\v{s}}evi{\v{c}}ius and
  Jaeger(2009)}]{lukovsevivcius2009reservoir}
\bibinfo{author}{M.~Luko{\v{s}}evi{\v{c}}ius}, \bibinfo{author}{H.~Jaeger},
  \bibinfo{title}{Reservoir computing approaches to recurrent neural network
  training}, \bibinfo{journal}{Computer Science Review}
  \bibinfo{volume}{3}~(\bibinfo{number}{3}) (\bibinfo{year}{2009})
  \bibinfo{pages}{127--149}, \urlprefix\url{10.1016/j.cosrev.2009.03.005}.

\bibitem[{Rumelhart et~al.(1985)Rumelhart, Hinton, and
  Williams}]{rumelhart1985learning}
\bibinfo{author}{D.~E. Rumelhart}, \bibinfo{author}{G.~E. Hinton},
  \bibinfo{author}{R.~J. Williams}, \bibinfo{title}{Learning internal
  representations by error propagation}, \bibinfo{type}{Tech. Rep.},
  \bibinfo{institution}{DTIC Document}, \bibinfo{year}{1985}.

\bibitem[{Williams and Peng(1990)}]{williams1990efficient}
\bibinfo{author}{R.~J. Williams}, \bibinfo{author}{J.~Peng}, \bibinfo{title}{An
  efficient gradient-based algorithm for on-line training of recurrent network
  trajectories}, \bibinfo{journal}{Neural computation}
  \bibinfo{volume}{2}~(\bibinfo{number}{4}) (\bibinfo{year}{1990})
  \bibinfo{pages}{490--501}, \urlprefix\url{10.1162/neco.1990.2.4.490}.

\bibitem[{Sutskever(2013)}]{sutskever2013training}
\bibinfo{author}{I.~Sutskever}, \bibinfo{title}{Training recurrent neural
  networks}, Ph.D. thesis, \bibinfo{school}{University of Toronto},
  \bibinfo{year}{2013}.

\bibitem[{Williams and Zipser(1995)}]{williams1995gradient}
\bibinfo{author}{R.~J. Williams}, \bibinfo{author}{D.~Zipser},
  \bibinfo{title}{Gradient-based learning algorithms for recurrent networks and
  their computational complexity}, \bibinfo{journal}{Backpropagation: Theory,
  architectures, and applications} \bibinfo{volume}{1} (\bibinfo{year}{1995})
  \bibinfo{pages}{433--486}.

\bibitem[{Pham et~al.(2014{\natexlab{a}})Pham, Bluche, Kermorvant, and
  Louradour}]{pham2014dropout}
\bibinfo{author}{V.~Pham}, \bibinfo{author}{T.~Bluche},
  \bibinfo{author}{C.~Kermorvant}, \bibinfo{author}{J.~Louradour},
  \bibinfo{title}{Dropout improves recurrent neural networks for handwriting
  recognition}, in: \bibinfo{booktitle}{Frontiers in Handwriting Recognition
  (ICFHR), 2014 14th International Conference on},
  \bibinfo{organization}{IEEE}, \bibinfo{pages}{285--290},
  \bibinfo{year}{2014}{\natexlab{a}}.

\bibitem[{{Gal} and {Ghahramani}(2015)}]{2015arXiv151205287G}
\bibinfo{author}{Y.~{Gal}}, \bibinfo{author}{Z.~{Ghahramani}},
  \bibinfo{title}{{A Theoretically Grounded Application of Dropout in Recurrent
  Neural Networks}}, \bibinfo{journal}{ArXiv e-prints} .

\bibitem[{{Zilly} et~al.(2016){Zilly}, {Srivastava}, {Koutn{\'{\i}}k}, and
  {Schmidhuber}}]{2016arXiv160703474Z}
\bibinfo{author}{J.~G. {Zilly}}, \bibinfo{author}{R.~K. {Srivastava}},
  \bibinfo{author}{J.~{Koutn{\'{\i}}k}}, \bibinfo{author}{J.~{Schmidhuber}},
  \bibinfo{title}{{Recurrent Highway Networks}}, \bibinfo{journal}{ArXiv
  e-prints} .

\bibitem[{Che et~al.(2016)Che, Purushotham, Cho, Sontag, and
  Liu}]{DBLP:journals/corr/ChePCSL16}
\bibinfo{author}{Z.~Che}, \bibinfo{author}{S.~Purushotham},
  \bibinfo{author}{K.~Cho}, \bibinfo{author}{D.~Sontag},
  \bibinfo{author}{Y.~Liu}, \bibinfo{title}{Recurrent Neural Networks for
  Multivariate Time Series with Missing Values}, \bibinfo{journal}{CoRR}
  \bibinfo{volume}{abs/1606.01865}.

\bibitem[{Neelakantan et~al.(2015)Neelakantan, Vilnis, Le, Sutskever, Kaiser,
  Kurach, and Martens}]{neelakantan2015adding}
\bibinfo{author}{A.~Neelakantan}, \bibinfo{author}{L.~Vilnis},
  \bibinfo{author}{Q.~V. Le}, \bibinfo{author}{I.~Sutskever},
  \bibinfo{author}{L.~Kaiser}, \bibinfo{author}{K.~Kurach},
  \bibinfo{author}{J.~Martens}, \bibinfo{title}{Adding Gradient Noise Improves
  Learning for Very Deep Networks}, \bibinfo{journal}{arXiv preprint
  arXiv:1511.06807} .

\bibitem[{Lee et~al.(2010)Lee, Recht, Srebro, Tropp, and
  Salakhutdinov}]{lee2010practical}
\bibinfo{author}{J.~D. Lee}, \bibinfo{author}{B.~Recht},
  \bibinfo{author}{N.~Srebro}, \bibinfo{author}{J.~Tropp},
  \bibinfo{author}{R.~R. Salakhutdinov}, \bibinfo{title}{Practical large-scale
  optimization for max-norm regularization}, in: \bibinfo{booktitle}{Advances
  in Neural Information Processing Systems}, \bibinfo{pages}{1297--1305},
  \bibinfo{year}{2010}.

\bibitem[{Bengio(2012)}]{Bengio2012}
\bibinfo{author}{Y.~Bengio}, \bibinfo{title}{Practical Recommendations for
  Gradient-Based Training of Deep Architectures}, in:
  \bibinfo{editor}{G.~Montavon}, \bibinfo{editor}{G.~B. Orr},
  \bibinfo{editor}{K.-R. M{\"u}ller} (Eds.), \bibinfo{booktitle}{Neural
  Networks: Tricks of the Trade: Second Edition}, \bibinfo{publisher}{Springer
  Berlin Heidelberg}, \bibinfo{address}{Berlin, Heidelberg}, ISBN
  \bibinfo{isbn}{978-3-642-35289-8}, \bibinfo{pages}{437--478},
  \urlprefix\url{10.1007/978-3-642-35289-8_26}, \bibinfo{year}{2012}.

\bibitem[{Bottou(2004)}]{bottou-mlss-2004}
\bibinfo{author}{L.~Bottou}, \bibinfo{title}{Stochastic Learning}, in:
  \bibinfo{editor}{O.~Bousquet}, \bibinfo{editor}{U.~von Luxburg} (Eds.),
  \bibinfo{booktitle}{Advanced Lectures on Machine Learning}, Lecture Notes in
  Artificial Intelligence, LNAI~3176, \bibinfo{publisher}{Springer Verlag},
  \bibinfo{address}{Berlin}, \bibinfo{pages}{146--168},
  \urlprefix\url{http://leon.bottou.org/papers/bottou-mlss-2004},
  \bibinfo{year}{2004}.

\bibitem[{Bottou(2012{\natexlab{a}})}]{bottou2012stochastic}
\bibinfo{author}{L.~Bottou}, \bibinfo{title}{Stochastic gradient descent
  tricks}, in: \bibinfo{booktitle}{Neural Networks: Tricks of the Trade},
  \bibinfo{publisher}{Springer}, \bibinfo{pages}{421--436},
  \bibinfo{year}{2012}{\natexlab{a}}.

\bibitem[{Bottou(2012{\natexlab{b}})}]{Bottou2012}
\bibinfo{author}{L.~Bottou}, \bibinfo{title}{Stochastic Gradient Descent
  Tricks}, in: \bibinfo{editor}{G.~Montavon}, \bibinfo{editor}{G.~B. Orr},
  \bibinfo{editor}{K.-R. M{\"u}ller} (Eds.), \bibinfo{booktitle}{Neural
  Networks: Tricks of the Trade: Second Edition}, \bibinfo{publisher}{Springer
  Berlin Heidelberg}, \bibinfo{address}{Berlin, Heidelberg}, ISBN
  \bibinfo{isbn}{978-3-642-35289-8}, \bibinfo{pages}{421--436},
  \urlprefix\url{10.1007/978-3-642-35289-8_25},
  \bibinfo{year}{2012}{\natexlab{b}}.

\bibitem[{Dauphin et~al.(2014)Dauphin, Pascanu, Gulcehre, Cho, Ganguli, and
  Bengio}]{NIPS2014_5486}
\bibinfo{author}{Y.~N. Dauphin}, \bibinfo{author}{R.~Pascanu},
  \bibinfo{author}{C.~Gulcehre}, \bibinfo{author}{K.~Cho},
  \bibinfo{author}{S.~Ganguli}, \bibinfo{author}{Y.~Bengio},
  \bibinfo{title}{Identifying and attacking the saddle point problem in
  high-dimensional non-convex optimization}, in:
  \bibinfo{editor}{Z.~Ghahramani}, \bibinfo{editor}{M.~Welling},
  \bibinfo{editor}{C.~Cortes}, \bibinfo{editor}{N.~D. Lawrence},
  \bibinfo{editor}{K.~Q. Weinberger} (Eds.), \bibinfo{booktitle}{Advances in
  Neural Information Processing Systems 27}, \bibinfo{publisher}{Curran
  Associates, Inc.}, \bibinfo{pages}{2933--2941}, \bibinfo{year}{2014}.

\bibitem[{Nesterov(1983)}]{nesterov1983method}
\bibinfo{author}{Y.~Nesterov}, \bibinfo{title}{{A method of solving a convex
  programming problem with convergence rate O(1/sqrt(k))}},
  \bibinfo{journal}{Soviet Mathematics Doklady} \bibinfo{volume}{27}
  (\bibinfo{year}{1983}) \bibinfo{pages}{372--376}.

\bibitem[{Duchi et~al.(2011)Duchi, Hazan, and Singer}]{duchi2011adaptive}
\bibinfo{author}{J.~Duchi}, \bibinfo{author}{E.~Hazan},
  \bibinfo{author}{Y.~Singer}, \bibinfo{title}{Adaptive subgradient methods for
  online learning and stochastic optimization}, \bibinfo{journal}{The Journal
  of Machine Learning Research} \bibinfo{volume}{12} (\bibinfo{year}{2011})
  \bibinfo{pages}{2121--2159}.

\bibitem[{Tieleman and Hinton(2012)}]{tieleman2012lecture}
\bibinfo{author}{T.~Tieleman}, \bibinfo{author}{G.~Hinton},
  \bibinfo{title}{Lecture 6.5-rmsprop: Divide the gradient by a running average
  of its recent magnitude}, \bibinfo{journal}{COURSERA: Neural Networks for
  Machine Learning} \bibinfo{volume}{4} (\bibinfo{year}{2012})
  \bibinfo{pages}{2}.

\bibitem[{Kingma and Ba(2014)}]{kingma2014adam}
\bibinfo{author}{D.~Kingma}, \bibinfo{author}{J.~Ba}, \bibinfo{title}{Adam: A
  method for stochastic optimization}, \bibinfo{journal}{arXiv preprint
  arXiv:1412.6980} .

\bibitem[{Martens(2010)}]{martens2010deep}
\bibinfo{author}{J.~Martens}, \bibinfo{title}{Deep learning via Hessian-free
  optimization}, in: \bibinfo{booktitle}{Proceedings of the 27th International
  Conference on Machine Learning (ICML-10)}, \bibinfo{pages}{735--742},
  \bibinfo{year}{2010}.

\bibitem[{Sutskever et~al.(2013)Sutskever, Martens, Dahl, and
  Hinton}]{sutskever2013importance}
\bibinfo{author}{I.~Sutskever}, \bibinfo{author}{J.~Martens},
  \bibinfo{author}{G.~E. Dahl}, \bibinfo{author}{G.~E. Hinton},
  \bibinfo{title}{On the importance of initialization and momentum in deep
  learning.}, \bibinfo{journal}{ICML (3)} \bibinfo{volume}{28}
  (\bibinfo{year}{2013}) \bibinfo{pages}{1139--1147}.

\bibitem[{Pascanu et~al.(2013{\natexlab{b}})Pascanu, G{\"{u}}l{\c{c}}ehre, Cho,
  and Bengio}]{DBLP:journals/corr/PascanuGCB13}
\bibinfo{author}{R.~Pascanu}, \bibinfo{author}{{\c{C}}.~G{\"{u}}l{\c{c}}ehre},
  \bibinfo{author}{K.~Cho}, \bibinfo{author}{Y.~Bengio}, \bibinfo{title}{How to
  Construct Deep Recurrent Neural Networks}, \bibinfo{journal}{CoRR}
  \bibinfo{volume}{abs/1312.6026}.

\bibitem[{Sutskever et~al.(2014)Sutskever, Vinyals, and
  Le}]{sutskever2014sequence}
\bibinfo{author}{I.~Sutskever}, \bibinfo{author}{O.~Vinyals},
  \bibinfo{author}{Q.~V. Le}, \bibinfo{title}{Sequence to sequence learning
  with neural networks}, in: \bibinfo{booktitle}{Advances in neural information
  processing systems}, \bibinfo{pages}{3104--3112}, \bibinfo{year}{2014}.

\bibitem[{El~Hihi and Bengio(1995)}]{el1995hierarchical}
\bibinfo{author}{S.~El~Hihi}, \bibinfo{author}{Y.~Bengio},
  \bibinfo{title}{Hierarchical Recurrent Neural Networks for Long-term
  Dependencies}, in: \bibinfo{booktitle}{Proceedings of the 8th International
  Conference on Neural Information Processing Systems}, NIPS'95,
  \bibinfo{publisher}{MIT Press}, \bibinfo{address}{Cambridge, MA, USA},
  \bibinfo{pages}{493--499},
  \urlprefix\url{http://dl.acm.org/citation.cfm?id=2998828.2998898},
  \bibinfo{year}{1995}.

\bibitem[{Hochreiter et~al.(2001)Hochreiter, Bengio, Frasconi, and
  Schmidhuber}]{hochreiter2001gradient}
\bibinfo{author}{S.~Hochreiter}, \bibinfo{author}{Y.~Bengio},
  \bibinfo{author}{P.~Frasconi}, \bibinfo{author}{J.~Schmidhuber},
  \bibinfo{title}{Gradient flow in recurrent nets: the difficulty of learning
  long-term dependencies}, \bibinfo{year}{2001}.

\bibitem[{Bianchi et~al.(2016{\natexlab{a}})Bianchi, Livi, and
  Alippi}]{7765110}
\bibinfo{author}{F.~M. Bianchi}, \bibinfo{author}{L.~Livi},
  \bibinfo{author}{C.~Alippi}, \bibinfo{title}{Investigating Echo-State
  Networks Dynamics by Means of Recurrence Analysis}, \bibinfo{journal}{IEEE
  Transactions on Neural Networks and Learning Systems}
  \bibinfo{volume}{PP}~(\bibinfo{number}{99})
  (\bibinfo{year}{2016}{\natexlab{a}}) \bibinfo{pages}{1--13}, ISSN
  \bibinfo{issn}{2162-237X}, \urlprefix\url{10.1109/TNNLS.2016.2630802}.

\bibitem[{Livi et~al.(2017)Livi, Bianchi, and Alippi}]{7817870}
\bibinfo{author}{L.~Livi}, \bibinfo{author}{F.~M. Bianchi},
  \bibinfo{author}{C.~Alippi}, \bibinfo{title}{Determination of the Edge of
  Criticality in Echo State Networks Through Fisher Information Maximization},
  \bibinfo{journal}{IEEE Transactions on Neural Networks and Learning Systems}
  \bibinfo{volume}{PP}~(\bibinfo{number}{99}) (\bibinfo{year}{2017})
  \bibinfo{pages}{1--12}, ISSN \bibinfo{issn}{2162-237X},
  \urlprefix\url{10.1109/TNNLS.2016.2644268}.

\bibitem[{Pascanu et~al.(2012)Pascanu, Mikolov, and
  Bengio}]{pascanu2012understanding}
\bibinfo{author}{R.~Pascanu}, \bibinfo{author}{T.~Mikolov},
  \bibinfo{author}{Y.~Bengio}, \bibinfo{title}{Understanding the exploding
  gradient problem}, \bibinfo{journal}{Computing Research Repository (CoRR)
  abs/1211.5063} .

\bibitem[{Glorot and Bengio(2010)}]{glorot2010understanding}
\bibinfo{author}{X.~Glorot}, \bibinfo{author}{Y.~Bengio},
  \bibinfo{title}{Understanding the difficulty of training deep feedforward
  neural networks}, in: \bibinfo{booktitle}{International conference on
  artificial intelligence and statistics}, \bibinfo{pages}{249--256},
  \bibinfo{year}{2010}.

\bibitem[{He et~al.(2015)He, Zhang, Ren, and Sun}]{he2015delving}
\bibinfo{author}{K.~He}, \bibinfo{author}{X.~Zhang}, \bibinfo{author}{S.~Ren},
  \bibinfo{author}{J.~Sun}, \bibinfo{title}{Delving deep into rectifiers:
  Surpassing human-level performance on imagenet classification}, in:
  \bibinfo{booktitle}{Proceedings of the IEEE International Conference on
  Computer Vision}, \bibinfo{pages}{1026--1034}, \bibinfo{year}{2015}.

\bibitem[{Nair and Hinton(2010)}]{DBLP:conf/icml/NairH10}
\bibinfo{author}{V.~Nair}, \bibinfo{author}{G.~E. Hinton},
  \bibinfo{title}{Rectified Linear Units Improve Restricted Boltzmann
  Machines}, in: \bibinfo{booktitle}{Proceedings of the 27th International
  Conference on Machine Learning (ICML-10), June 21-24, 2010, Haifa, Israel},
  \bibinfo{pages}{807--814}, \bibinfo{year}{2010}.

\bibitem[{Srivastava et~al.(2015)Srivastava, Greff, and
  Schmidhuber}]{NIPS2015_5850}
\bibinfo{author}{R.~K. Srivastava}, \bibinfo{author}{K.~Greff},
  \bibinfo{author}{J.~Schmidhuber}, \bibinfo{title}{Training Very Deep
  Networks}, in: \bibinfo{editor}{C.~Cortes}, \bibinfo{editor}{N.~D. Lawrence},
  \bibinfo{editor}{D.~D. Lee}, \bibinfo{editor}{M.~Sugiyama},
  \bibinfo{editor}{R.~Garnett} (Eds.), \bibinfo{booktitle}{Advances in Neural
  Information Processing Systems 28}, \bibinfo{publisher}{Curran Associates,
  Inc.}, \bibinfo{pages}{2377--2385}, \bibinfo{year}{2015}.

\bibitem[{{He} et~al.(2015){He}, {Zhang}, {Ren}, and
  {Sun}}]{2015arXiv151203385H}
\bibinfo{author}{K.~{He}}, \bibinfo{author}{X.~{Zhang}},
  \bibinfo{author}{S.~{Ren}}, \bibinfo{author}{J.~{Sun}}, \bibinfo{title}{{Deep
  Residual Learning for Image Recognition}}, \bibinfo{journal}{ArXiv e-prints}
  .

\bibitem[{Gomez and Miikkulainen(2003)}]{gomez2003robust}
\bibinfo{author}{F.~J. Gomez}, \bibinfo{author}{R.~Miikkulainen},
  \bibinfo{title}{Robust non-linear control through neuroevolution},
  \bibinfo{publisher}{Computer Science Department, University of Texas at
  Austin}, \bibinfo{year}{2003}.

\bibitem[{Elman(1995)}]{elman1995language}
\bibinfo{author}{J.~L. Elman}, \bibinfo{title}{Language as a dynamical system},
  \bibinfo{journal}{Mind as motion: Explorations in the dynamics of cognition}
  (\bibinfo{year}{1995}) \bibinfo{pages}{195--223}.

\bibitem[{Ogata et~al.(2007)Ogata, Murase, Tani, Komatani, and
  Okuno}]{ogata2007two}
\bibinfo{author}{T.~Ogata}, \bibinfo{author}{M.~Murase},
  \bibinfo{author}{J.~Tani}, \bibinfo{author}{K.~Komatani},
  \bibinfo{author}{H.~G. Okuno}, \bibinfo{title}{Two-way translation of
  compound sentences and arm motions by recurrent neural networks}, in:
  \bibinfo{booktitle}{Intelligent Robots and Systems, 2007. IROS 2007. IEEE/RSJ
  International Conference on}, \bibinfo{organization}{IEEE},
  \bibinfo{pages}{1858--1863}, \bibinfo{year}{2007}.

\bibitem[{Mori and Ogasawara(1993)}]{Mori1993}
\bibinfo{author}{H.~M.~H. Mori}, \bibinfo{author}{T.~O.~T. Ogasawara},
  \bibinfo{title}{{A recurrent neural network for short-term load
  forecasting}}, \bibinfo{journal}{1993 Proceedings of the Second International
  Forum on Applications of Neural Networks to Power Systems}
  \bibinfo{volume}{31} (\bibinfo{year}{1993}) \bibinfo{pages}{276--281},
  \urlprefix\url{10.1109/ANN.1993.264315}.

\bibitem[{Cai et~al.(2007)Cai, Zhang, Venayagamoorthy, and Wunsch}]{Cai2007}
\bibinfo{author}{X.~Cai}, \bibinfo{author}{N.~Zhang}, \bibinfo{author}{G.~K.
  Venayagamoorthy}, \bibinfo{author}{D.~C. Wunsch}, \bibinfo{title}{{Time
  series prediction with recurrent neural networks trained by a hybrid PSO-EA
  algorithm}}, \bibinfo{journal}{Neurocomputing}
  \bibinfo{volume}{70}~(\bibinfo{number}{13-15}) (\bibinfo{year}{2007})
  \bibinfo{pages}{2342--2353}, ISSN \bibinfo{issn}{09252312},
  \urlprefix\url{10.1016/j.neucom.2005.12.138}.

\bibitem[{Cho(2003)}]{Cho2003323}
\bibinfo{author}{V.~Cho}, \bibinfo{title}{A comparison of three different
  approaches to tourist arrival forecasting}, \bibinfo{journal}{Tourism
  Management} \bibinfo{volume}{24}~(\bibinfo{number}{3}) (\bibinfo{year}{2003})
  \bibinfo{pages}{323 -- 330}, ISSN \bibinfo{issn}{0261-5177},
  \urlprefix\url{http://doi.org/10.1016/S0261-5177(02)00068-7}.

\bibitem[{Mandal et~al.(2006)Mandal, Senjyu, Urasaki, and
  Funabashi}]{Mandal2006}
\bibinfo{author}{P.~Mandal}, \bibinfo{author}{T.~Senjyu},
  \bibinfo{author}{N.~Urasaki}, \bibinfo{author}{T.~Funabashi},
  \bibinfo{title}{{A neural network based several-hour-ahead electric load
  forecasting using similar days approach}}, \bibinfo{journal}{International
  Journal of Electrical Power and Energy Systems}
  \bibinfo{volume}{28}~(\bibinfo{number}{6}) (\bibinfo{year}{2006})
  \bibinfo{pages}{367--373}, ISSN \bibinfo{issn}{01420615},
  \urlprefix\url{10.1016/j.ijepes.2005.12.007}.

\bibitem[{Chitsaz et~al.(2015)Chitsaz, Shaker, Zareipour, Wood, and
  Amjady}]{Chitsaz2015}
\bibinfo{author}{H.~Chitsaz}, \bibinfo{author}{H.~Shaker},
  \bibinfo{author}{H.~Zareipour}, \bibinfo{author}{D.~Wood},
  \bibinfo{author}{N.~Amjady}, \bibinfo{title}{{Short-term electricity load
  forecasting of buildings in microgrids}}, \bibinfo{journal}{Energy and
  Buildings} \bibinfo{volume}{99} (\bibinfo{year}{2015})
  \bibinfo{pages}{50--60}, ISSN \bibinfo{issn}{03787788},
  \urlprefix\url{10.1016/j.enbuild.2015.04.011}.

\bibitem[{Graves et~al.(2013)Graves, Mohamed, and Hinton}]{graves2013speech}
\bibinfo{author}{A.~Graves}, \bibinfo{author}{A.-r. Mohamed},
  \bibinfo{author}{G.~Hinton}, \bibinfo{title}{Speech recognition with deep
  recurrent neural networks}, in: \bibinfo{booktitle}{Acoustics, Speech and
  Signal Processing (ICASSP), 2013 IEEE International Conference on},
  \bibinfo{organization}{IEEE}, \bibinfo{pages}{6645--6649},
  \bibinfo{year}{2013}.

\bibitem[{Eck and Schmidhuber(2002)}]{eck2002finding}
\bibinfo{author}{D.~Eck}, \bibinfo{author}{J.~Schmidhuber},
  \bibinfo{title}{Finding temporal structure in music: Blues improvisation with
  LSTM recurrent networks}, in: \bibinfo{booktitle}{Neural Networks for Signal
  Processing, 2002. Proceedings of the 2002 12th IEEE Workshop on},
  \bibinfo{organization}{IEEE}, \bibinfo{pages}{747--756},
  \bibinfo{year}{2002}.

\bibitem[{Gers and Schmidhuber(2001)}]{gers2001lstm}
\bibinfo{author}{F.~A. Gers}, \bibinfo{author}{J.~Schmidhuber},
  \bibinfo{title}{LSTM recurrent networks learn simple context-free and
  context-sensitive languages}, \bibinfo{journal}{Neural Networks, IEEE
  Transactions on} \bibinfo{volume}{12}~(\bibinfo{number}{6})
  (\bibinfo{year}{2001}) \bibinfo{pages}{1333--1340}.

\bibitem[{Vinyals et~al.(2017)Vinyals, Toshev, Bengio, and
  Erhan}]{Vinyals_2015_CVPR}
\bibinfo{author}{O.~Vinyals}, \bibinfo{author}{A.~Toshev},
  \bibinfo{author}{S.~Bengio}, \bibinfo{author}{D.~Erhan}, \bibinfo{title}{Show
  and Tell: Lessons Learned from the 2015 {MSCOCO} Image Captioning Challenge},
  \bibinfo{journal}{{IEEE} Trans. Pattern Anal. Mach. Intell.}
  \bibinfo{volume}{39}~(\bibinfo{number}{4}) (\bibinfo{year}{2017})
  \bibinfo{pages}{652--663}, \urlprefix\url{10.1109/TPAMI.2016.2587640}.

\bibitem[{Ma et~al.(2015)Ma, Tao, Wang, Yu, and Wang}]{Ma2015}
\bibinfo{author}{X.~Ma}, \bibinfo{author}{Z.~Tao}, \bibinfo{author}{Y.~Wang},
  \bibinfo{author}{H.~Yu}, \bibinfo{author}{Y.~Wang}, \bibinfo{title}{{Long
  short-term memory neural network for traffic speed prediction using remote
  microwave sensor data}}, \bibinfo{journal}{Transportation Research Part C:
  Emerging Technologies} \bibinfo{volume}{54} (\bibinfo{year}{2015})
  \bibinfo{pages}{187--197}, ISSN \bibinfo{issn}{0968090X},
  \urlprefix\url{10.1016/j.trc.2015.03.014}.

\bibitem[{Pawlowski and Kurach(2015)}]{Pawlowski2015}
\bibinfo{author}{K.~Pawlowski}, \bibinfo{author}{K.~Kurach},
  \bibinfo{title}{Detecting Methane Outbreaks from Time Series Data with Deep
  Neural Networks}, in: \bibinfo{booktitle}{Rough Sets, Fuzzy Sets, Data
  Mining, and Granular Computing - 15th International Conference, RSFDGrC 2015,
  Tianjin, China, November 20-23, 2015, Proceedings}, vol.
  \bibinfo{volume}{9437}, ISBN \bibinfo{isbn}{978-3-319-25782-2}, ISSN
  \bibinfo{issn}{03029743}, \bibinfo{pages}{475--484},
  \urlprefix\url{10.1007/978-3-319-25783-9_42}, \bibinfo{year}{2015}.

\bibitem[{Felder et~al.(2010)Felder, Kaifel, and Graves}]{Felder2010}
\bibinfo{author}{M.~Felder}, \bibinfo{author}{A.~Kaifel},
  \bibinfo{author}{A.~Graves}, \bibinfo{title}{{Wind Power Prediction using
  Mixture Density Recurrent Neural Networks}}, in: \bibinfo{booktitle}{Poster
  P0.153}, \bibinfo{pages}{1--7}, \bibinfo{year}{2010}.

\bibitem[{Graves and Schmidhuber(2005)}]{Graves2005602}
\bibinfo{author}{A.~Graves}, \bibinfo{author}{J.~Schmidhuber},
  \bibinfo{title}{Framewise phoneme classification with bidirectional LSTM and
  other neural network architectures}, \bibinfo{journal}{Neural Networks}
  \bibinfo{volume}{18}~(\bibinfo{number}{5–6}) (\bibinfo{year}{2005})
  \bibinfo{pages}{602 -- 610}, ISSN \bibinfo{issn}{0893-6080},
  \urlprefix\url{10.1016/j.neunet.2005.06.042}, \bibinfo{note}{iJCNN 2005}.

\bibitem[{Cho et~al.(2014)Cho, Van~Merri{\"e}nboer, Gulcehre, Bahdanau,
  Bougares, Schwenk, and Bengio}]{cho2014learning}
\bibinfo{author}{K.~Cho}, \bibinfo{author}{B.~Van~Merri{\"e}nboer},
  \bibinfo{author}{C.~Gulcehre}, \bibinfo{author}{D.~Bahdanau},
  \bibinfo{author}{F.~Bougares}, \bibinfo{author}{H.~Schwenk},
  \bibinfo{author}{Y.~Bengio}, \bibinfo{title}{Learning phrase representations
  using RNN encoder-decoder for statistical machine translation},
  \bibinfo{journal}{arXiv preprint arXiv:1406.1078} .

\bibitem[{Chung et~al.(2014)Chung, G{\"{u}}l{\c{c}}ehre, Cho, and
  Bengio}]{DBLP:journals/corr/ChungGCB14}
\bibinfo{author}{J.~Chung}, \bibinfo{author}{{\c{C}}.~G{\"{u}}l{\c{c}}ehre},
  \bibinfo{author}{K.~Cho}, \bibinfo{author}{Y.~Bengio},
  \bibinfo{title}{Empirical Evaluation of Gated Recurrent Neural Networks on
  Sequence Modeling}, \bibinfo{journal}{CoRR} \bibinfo{volume}{abs/1412.3555}.

\bibitem[{Zaremba(2015)}]{zaremba2015empirical}
\bibinfo{author}{W.~Zaremba}, \bibinfo{title}{An empirical exploration of
  recurrent network architectures}, \bibinfo{journal}{Proceedings of the 32nd
  International Conference on Machine Learning, Lille, France} .

\bibitem[{Leontaritis and Billings(1985)}]{leontaritis1985input}
\bibinfo{author}{I.~Leontaritis}, \bibinfo{author}{S.~A. Billings},
  \bibinfo{title}{Input-output parametric models for non-linear systems Part I:
  deterministic non-linear systems}, \bibinfo{journal}{International journal of
  control} \bibinfo{volume}{41}~(\bibinfo{number}{2}) (\bibinfo{year}{1985})
  \bibinfo{pages}{303--328}.

\bibitem[{Diaconescu(2008)}]{diaconescu2008use}
\bibinfo{author}{E.~Diaconescu}, \bibinfo{title}{The use of NARX neural
  networks to predict chaotic time series}, \bibinfo{journal}{Wseas
  Transactions on computer research} \bibinfo{volume}{3}~(\bibinfo{number}{3})
  (\bibinfo{year}{2008}) \bibinfo{pages}{182--191}.

\bibitem[{Lin et~al.(1997)Lin, Giles, Horne, and Kung}]{lin1997delay}
\bibinfo{author}{T.-N. Lin}, \bibinfo{author}{C.~L. Giles},
  \bibinfo{author}{B.~G. Horne}, \bibinfo{author}{S.-Y. Kung},
  \bibinfo{title}{A delay damage model selection algorithm for NARX neural
  networks}, \bibinfo{journal}{Signal Processing, IEEE Transactions on}
  \bibinfo{volume}{45}~(\bibinfo{number}{11}) (\bibinfo{year}{1997})
  \bibinfo{pages}{2719--2730}.

\bibitem[{Menezes and Barreto(2008)}]{menezes2008}
\bibinfo{author}{J.~M.~P. Menezes}, \bibinfo{author}{G.~A. Barreto},
  \bibinfo{title}{Long-term time series prediction with the NARX network: an
  empirical evaluation}, \bibinfo{journal}{Neurocomputing}
  \bibinfo{volume}{71}~(\bibinfo{number}{16}) (\bibinfo{year}{2008})
  \bibinfo{pages}{3335--3343}.

\bibitem[{Xie et~al.(2009)Xie, Tang, and Liao}]{5212326}
\bibinfo{author}{H.~Xie}, \bibinfo{author}{H.~Tang}, \bibinfo{author}{Y.-H.
  Liao}, \bibinfo{title}{Time series prediction based on NARX neural networks:
  An advanced approach}, in: \bibinfo{booktitle}{Machine Learning and
  Cybernetics, 2009 International Conference on}, vol.~\bibinfo{volume}{3},
  \bibinfo{pages}{1275--1279}, \urlprefix\url{10.1109/ICMLC.2009.5212326},
  \bibinfo{year}{2009}.

\bibitem[{Napoli and Piroddi(2010)}]{napoli2010nonlinear}
\bibinfo{author}{R.~Napoli}, \bibinfo{author}{L.~Piroddi},
  \bibinfo{title}{Nonlinear active noise control with NARX models},
  \bibinfo{journal}{Audio, Speech, and Language Processing, IEEE Transactions
  on} \bibinfo{volume}{18}~(\bibinfo{number}{2}) (\bibinfo{year}{2010})
  \bibinfo{pages}{286--295}.

\bibitem[{Plett(2003)}]{plett2003adaptive}
\bibinfo{author}{G.~L. Plett}, \bibinfo{title}{Adaptive inverse control of
  linear and nonlinear systems using dynamic neural networks},
  \bibinfo{journal}{Neural Networks, IEEE Transactions on}
  \bibinfo{volume}{14}~(\bibinfo{number}{2}) (\bibinfo{year}{2003})
  \bibinfo{pages}{360--376}.

\bibitem[{Billings(2013)}]{billings2013nonlinear}
\bibinfo{author}{S.~A. Billings}, \bibinfo{title}{Nonlinear system
  identification: NARMAX methods in the time, frequency, and spatio-temporal
  domains}, \bibinfo{publisher}{John Wiley \& Sons}, \bibinfo{year}{2013}.

\bibitem[{Siegelmann et~al.(1997)Siegelmann, Horne, and
  Giles}]{siegelmann1997computational}
\bibinfo{author}{H.~T. Siegelmann}, \bibinfo{author}{B.~G. Horne},
  \bibinfo{author}{C.~L. Giles}, \bibinfo{title}{Computational capabilities of
  recurrent NARX neural networks}, \bibinfo{journal}{Systems, Man, and
  Cybernetics, Part B: Cybernetics, IEEE Transactions on}
  \bibinfo{volume}{27}~(\bibinfo{number}{2}) (\bibinfo{year}{1997})
  \bibinfo{pages}{208--215}.

\bibitem[{Lin et~al.(1996)Lin, Horne, Ti{\v{n}}o, and Giles}]{lin1996learning}
\bibinfo{author}{T.~Lin}, \bibinfo{author}{B.~G. Horne},
  \bibinfo{author}{P.~Ti{\v{n}}o}, \bibinfo{author}{C.~L. Giles},
  \bibinfo{title}{Learning long-term dependencies in NARX recurrent neural
  networks}, \bibinfo{journal}{Neural Networks, IEEE Transactions on}
  \bibinfo{volume}{7}~(\bibinfo{number}{6}) (\bibinfo{year}{1996})
  \bibinfo{pages}{1329--1338}.

\bibitem[{Huang et~al.(2005)Huang, Huang, and Wang}]{huang2005particle}
\bibinfo{author}{C.-M. Huang}, \bibinfo{author}{C.-J. Huang},
  \bibinfo{author}{M.-L. Wang}, \bibinfo{title}{A particle swarm optimization
  to identifying the ARMAX model for short-term load forecasting},
  \bibinfo{journal}{Power Systems, IEEE Transactions on}
  \bibinfo{volume}{20}~(\bibinfo{number}{2}) (\bibinfo{year}{2005})
  \bibinfo{pages}{1126--1133}.

\bibitem[{Cambria et~al.(2013)Cambria, Huang, Kasun, Zhou, Vong, Lin, Yin, Cai,
  Liu, Li et~al.}]{cambria2013extreme}
\bibinfo{author}{E.~Cambria}, \bibinfo{author}{G.-B. Huang},
  \bibinfo{author}{L.~L.~C. Kasun}, \bibinfo{author}{H.~Zhou},
  \bibinfo{author}{C.~M. Vong}, \bibinfo{author}{J.~Lin},
  \bibinfo{author}{J.~Yin}, \bibinfo{author}{Z.~Cai}, \bibinfo{author}{Q.~Liu},
  \bibinfo{author}{K.~Li}, et~al., \bibinfo{title}{Extreme learning machines
  [trends \& controversies]}, \bibinfo{journal}{Intelligent Systems, IEEE}
  \bibinfo{volume}{28}~(\bibinfo{number}{6}) (\bibinfo{year}{2013})
  \bibinfo{pages}{30--59}.

\bibitem[{Scardapane et~al.(2015)Scardapane, Comminiello, Scarpiniti, and
  Uncini}]{7000606}
\bibinfo{author}{S.~Scardapane}, \bibinfo{author}{D.~Comminiello},
  \bibinfo{author}{M.~Scarpiniti}, \bibinfo{author}{A.~Uncini},
  \bibinfo{title}{Online Sequential Extreme Learning Machine With Kernels},
  \bibinfo{journal}{IEEE Transactions on Neural Networks and Learning Systems}
  \bibinfo{volume}{26}~(\bibinfo{number}{9}) (\bibinfo{year}{2015})
  \bibinfo{pages}{2214--2220}, ISSN \bibinfo{issn}{2162-237X},
  \urlprefix\url{10.1109/TNNLS.2014.2382094}.

\bibitem[{Maass et~al.(2002)Maass, Natschl{\"a}ger, and
  Markram}]{maass2002real}
\bibinfo{author}{W.~Maass}, \bibinfo{author}{T.~Natschl{\"a}ger},
  \bibinfo{author}{H.~Markram}, \bibinfo{title}{Real-time computing without
  stable states: A new framework for neural computation based on
  perturbations}, \bibinfo{journal}{Neural computation}
  \bibinfo{volume}{14}~(\bibinfo{number}{11}) (\bibinfo{year}{2002})
  \bibinfo{pages}{2531--2560}, \urlprefix\url{10.1162/089976602760407955}.

\bibitem[{Alexandre et~al.(2009)Alexandre, Embrechts, and
  Linton}]{alexandre2009benchmarking}
\bibinfo{author}{L.~A. Alexandre}, \bibinfo{author}{M.~J. Embrechts},
  \bibinfo{author}{J.~Linton}, \bibinfo{title}{Benchmarking reservoir computing
  on time-independent classification tasks}, in: \bibinfo{booktitle}{Neural
  Networks, 2009. IJCNN 2009. International Joint Conference on},
  \bibinfo{organization}{IEEE}, \bibinfo{pages}{89--93}, \bibinfo{year}{2009}.

\bibitem[{Skowronski and Harris(2007)}]{skowronski2007automatic}
\bibinfo{author}{M.~D. Skowronski}, \bibinfo{author}{J.~G. Harris},
  \bibinfo{title}{Automatic speech recognition using a predictive echo state
  network classifier}, \bibinfo{journal}{Neural networks}
  \bibinfo{volume}{20}~(\bibinfo{number}{3}) (\bibinfo{year}{2007})
  \bibinfo{pages}{414--423}.

\bibitem[{Hai-yan et~al.(2005)Hai-yan, Wen-jiang, and Zhen-ya}]{1504645}
\bibinfo{author}{D.~Hai-yan}, \bibinfo{author}{P.~Wen-jiang},
  \bibinfo{author}{H.~Zhen-ya}, \bibinfo{title}{A multiple objective
  optimization based echo state network tree and application to intrusion
  detection}, in: \bibinfo{booktitle}{VLSI Design and Video Technology, 2005.
  Proceedings of 2005 IEEE International Workshop on},
  \bibinfo{pages}{443--446}, \urlprefix\url{10.1109/IWVDVT.2005.1504645},
  \bibinfo{year}{2005}.

\bibitem[{Han and Lee(2014{\natexlab{a}})}]{6480841}
\bibinfo{author}{S.~Han}, \bibinfo{author}{J.~Lee}, \bibinfo{title}{Fuzzy Echo
  State Neural Networks and Funnel Dynamic Surface Control for Prescribed
  Performance of a Nonlinear Dynamic System}, \bibinfo{journal}{Industrial
  Electronics, IEEE Transactions on} \bibinfo{volume}{61}~(\bibinfo{number}{2})
  (\bibinfo{year}{2014}{\natexlab{a}}) \bibinfo{pages}{1099--1112}, ISSN
  \bibinfo{issn}{0278-0046}, \urlprefix\url{10.1109/TIE.2013.2253072}.

\bibitem[{Maiorino et~al.(2017)Maiorino, Bianchi, Livi, Rizzi, and
  Sadeghian}]{desn}
\bibinfo{author}{E.~Maiorino}, \bibinfo{author}{F.~Bianchi},
  \bibinfo{author}{L.~Livi}, \bibinfo{author}{A.~Rizzi},
  \bibinfo{author}{A.~Sadeghian}, \bibinfo{title}{Data-driven detrending of
  nonstationary fractal time series with echo state networks},
  \bibinfo{journal}{Information Sciences} \bibinfo{volume}{382-383}
  (\bibinfo{year}{2017}) \bibinfo{pages}{359--373},
  \urlprefix\url{10.1016/j.ins.2016.12.015}.

\bibitem[{Mazumdar and Harley(2008)}]{4712533}
\bibinfo{author}{J.~Mazumdar}, \bibinfo{author}{R.~Harley},
  \bibinfo{title}{Utilization of Echo State Networks for Differentiating Source
  and Nonlinear Load Harmonics in the Utility Network}, \bibinfo{journal}{Power
  Electronics, IEEE Transactions on} \bibinfo{volume}{23}~(\bibinfo{number}{6})
  (\bibinfo{year}{2008}) \bibinfo{pages}{2738--2745}, ISSN
  \bibinfo{issn}{0885-8993}, \urlprefix\url{10.1109/TPEL.2008.2005097}.

\bibitem[{Han and Lee(2014{\natexlab{b}})}]{han2014fuzzy}
\bibinfo{author}{S.~I. Han}, \bibinfo{author}{J.~M. Lee}, \bibinfo{title}{Fuzzy
  echo state neural networks and funnel dynamic surface control for prescribed
  performance of a nonlinear dynamic system}, \bibinfo{journal}{Industrial
  Electronics, IEEE Transactions on} \bibinfo{volume}{61}~(\bibinfo{number}{2})
  (\bibinfo{year}{2014}{\natexlab{b}}) \bibinfo{pages}{1099--1112}.

\bibitem[{Niu et~al.(2012)Niu, Ji, Xing, and Wang}]{niu2012multi}
\bibinfo{author}{D.~Niu}, \bibinfo{author}{L.~Ji}, \bibinfo{author}{M.~Xing},
  \bibinfo{author}{J.~Wang}, \bibinfo{title}{{M}ulti-variable {E}cho {S}tate
  {N}etwork {O}ptimized by {B}ayesian {R}egulation for {D}aily {P}eak {L}oad
  {F}orecasting}, \bibinfo{journal}{{J}ournal of {N}etworks}
  \bibinfo{volume}{7}~(\bibinfo{number}{11}) (\bibinfo{year}{2012})
  \bibinfo{pages}{1790--1795}.

\bibitem[{L{\o}kse et~al.(2017)L{\o}kse, Bianchi, and Jenssen}]{Lokse2017}
\bibinfo{author}{S.~L{\o}kse}, \bibinfo{author}{F.~M. Bianchi},
  \bibinfo{author}{R.~Jenssen}, \bibinfo{title}{Training Echo State Networks
  with Regularization Through Dimensionality Reduction},
  \bibinfo{journal}{Cognitive Computation}  (\bibinfo{year}{2017})
  \bibinfo{pages}{1--15}ISSN \bibinfo{issn}{1866-9964},
  \urlprefix\url{10.1007/s12559-017-9450-z}.

\bibitem[{Varshney and Verma(2014)}]{varshneyhalf}
\bibinfo{author}{S.~Varshney}, \bibinfo{author}{T.~Verma},
  \bibinfo{title}{{H}alf {H}ourly {E}lectricity {L}oad {P}rediction using
  {E}cho {S}tate {N}etwork}, \bibinfo{journal}{{I}nternational {J}ournal of
  {S}cience and {R}esearch} \bibinfo{volume}{3}~(\bibinfo{number}{6})
  (\bibinfo{year}{2014}) \bibinfo{pages}{885--888}.

\bibitem[{Li et~al.(2012{\natexlab{a}})Li, Han, and Wang}]{li2012chaotic}
\bibinfo{author}{D.~Li}, \bibinfo{author}{M.~Han}, \bibinfo{author}{J.~Wang},
  \bibinfo{title}{{C}haotic time series prediction based on a novel robust echo
  state network}, \bibinfo{journal}{{IEEE} {T}ransactions on {N}eural
  {N}etworks and {L}earning {S}ystems}
  \bibinfo{volume}{23}~(\bibinfo{number}{5})
  (\bibinfo{year}{2012}{\natexlab{a}}) \bibinfo{pages}{787--799}.

\bibitem[{Jaeger and Haas(2004)}]{jaeger2004harnessing}
\bibinfo{author}{H.~Jaeger}, \bibinfo{author}{H.~Haas},
  \bibinfo{title}{Harnessing Nonlinearity: Predicting Chaotic Systems and
  Saving Energy in Wireless Communication}, \bibinfo{journal}{Science}
  \bibinfo{volume}{304}~(\bibinfo{number}{5667}) (\bibinfo{year}{2004})
  \bibinfo{pages}{78--80}, ISSN \bibinfo{issn}{0036-8075},
  \urlprefix\url{10.1126/science.1091277}.

\bibitem[{Rodan and Ti{\v{n}}o(2011)}]{rodan2011minimum}
\bibinfo{author}{A.~Rodan}, \bibinfo{author}{P.~Ti{\v{n}}o},
  \bibinfo{title}{Minimum complexity echo state network},
  \bibinfo{journal}{IEEE Transactions on Neural Networks}
  \bibinfo{volume}{22}~(\bibinfo{number}{1}) (\bibinfo{year}{2011})
  \bibinfo{pages}{131--144}, \urlprefix\url{10.1109/TNN.2010.2089641}.

\bibitem[{Appeltant et~al.(2011)Appeltant, Soriano, {Van der Sande}, Danckaert,
  Massar, Dambre, Schrauwen, Mirasso, and Fischer}]{appeltant2011information}
\bibinfo{author}{L.~Appeltant}, \bibinfo{author}{M.~C. Soriano},
  \bibinfo{author}{G.~{Van der Sande}}, \bibinfo{author}{J.~Danckaert},
  \bibinfo{author}{S.~Massar}, \bibinfo{author}{J.~Dambre},
  \bibinfo{author}{B.~Schrauwen}, \bibinfo{author}{C.~R. Mirasso},
  \bibinfo{author}{I.~Fischer}, \bibinfo{title}{Information processing using a
  single dynamical node as complex system}, \bibinfo{journal}{Nature
  Communications} \bibinfo{volume}{2} (\bibinfo{year}{2011})
  \bibinfo{pages}{468}, \urlprefix\url{10.1038/ncomms1476}.

\bibitem[{Boedecker et~al.(2012)Boedecker, Obst, Lizier, Mayer, and
  Asada}]{boedecker2012information}
\bibinfo{author}{J.~Boedecker}, \bibinfo{author}{O.~Obst},
  \bibinfo{author}{J.~T. Lizier}, \bibinfo{author}{N.~M. Mayer},
  \bibinfo{author}{M.~Asada}, \bibinfo{title}{Information processing in echo
  state networks at the edge of chaos}, \bibinfo{journal}{Theory in
  Biosciences} \bibinfo{volume}{131}~(\bibinfo{number}{3})
  (\bibinfo{year}{2012}) \bibinfo{pages}{205--213}.

\bibitem[{Verstraeten and Schrauwen(2009)}]{verstraeten2009quantification}
\bibinfo{author}{D.~Verstraeten}, \bibinfo{author}{B.~Schrauwen},
  \bibinfo{title}{On the Quantification of Dynamics in Reservoir Computing},
  in: \bibinfo{editor}{C.~Alippi}, \bibinfo{editor}{M.~Polycarpou},
  \bibinfo{editor}{C.~Panayiotou}, \bibinfo{editor}{G.~Ellinas} (Eds.),
  \bibinfo{booktitle}{Artificial Neural Networks -- ICANN 2009}, vol.
  \bibinfo{volume}{5768}, \bibinfo{publisher}{Springer Berlin, Heidelberg},
  ISBN \bibinfo{isbn}{978-3-642-04273-7}, \bibinfo{pages}{985--994},
  \urlprefix\url{10.1007/978-3-642-04274-4_101}, \bibinfo{year}{2009}.

\bibitem[{{Bianchi Filippo Maria} et~al.(2017){Bianchi Filippo Maria}, {Livi
  Lorenzo}, {Alippi Cesare}, and {Jenssen Robert}}]{rnnmultiplex}
\bibinfo{author}{{Bianchi Filippo Maria}}, \bibinfo{author}{{Livi Lorenzo}},
  \bibinfo{author}{{Alippi Cesare}}, \bibinfo{author}{{Jenssen Robert}},
  \bibinfo{title}{{Multiplex visibility graphs to investigate recurrent neural
  network dynamics}}, \bibinfo{journal}{Scientific Reports} \bibinfo{volume}{7}
  (\bibinfo{year}{2017}) \bibinfo{pages}{44037},
  \urlprefix\url{http://dx.doi.org/10.1038/srep44037 10.1038/srep44037}.

\bibitem[{Fraser and Swinney(1986)}]{fraser1986independent}
\bibinfo{author}{A.~M. Fraser}, \bibinfo{author}{H.~L. Swinney},
  \bibinfo{title}{Independent coordinates for strange attractors from mutual
  information}, \bibinfo{journal}{Physical review A}
  \bibinfo{volume}{33}~(\bibinfo{number}{2}) (\bibinfo{year}{1986})
  \bibinfo{pages}{1134}.

\bibitem[{Liebert and Schuster(1989)}]{liebert1989proper}
\bibinfo{author}{W.~Liebert}, \bibinfo{author}{H.~Schuster},
  \bibinfo{title}{Proper choice of the time delay for the analysis of chaotic
  time series}, \bibinfo{journal}{Physics Letters A}
  \bibinfo{volume}{142}~(\bibinfo{number}{2-3}) (\bibinfo{year}{1989})
  \bibinfo{pages}{107--111}.

\bibitem[{Jaeger(2002{\natexlab{b}})}]{jaeger2002adaptive}
\bibinfo{author}{H.~Jaeger}, \bibinfo{title}{Adaptive nonlinear system
  identification with echo state networks}, in: \bibinfo{booktitle}{Advances in
  neural information processing systems}, \bibinfo{pages}{593--600},
  \bibinfo{year}{2002}{\natexlab{b}}.

\bibitem[{Xue et~al.(2007)Xue, Yang, and Haykin}]{Xue2007365}
\bibinfo{author}{Y.~Xue}, \bibinfo{author}{L.~Yang},
  \bibinfo{author}{S.~Haykin}, \bibinfo{title}{Decoupled echo state networks
  with lateral inhibition}, \bibinfo{journal}{Neural Networks}
  \bibinfo{volume}{20}~(\bibinfo{number}{3}) (\bibinfo{year}{2007})
  \bibinfo{pages}{365 -- 376}, ISSN \bibinfo{issn}{0893-6080},
  \urlprefix\url{http://dx.doi.org/10.1016/j.neunet.2007.04.014},
  \bibinfo{note}{echo State Networks and Liquid State Machines}.

\bibitem[{Wierstra et~al.(2005)Wierstra, Gomez, and
  Schmidhuber}]{wierstra2005modeling}
\bibinfo{author}{D.~Wierstra}, \bibinfo{author}{F.~J. Gomez},
  \bibinfo{author}{J.~Schmidhuber}, \bibinfo{title}{Modeling systems with
  internal state using evolino}, in: \bibinfo{booktitle}{Proceedings of the 7th
  annual conference on Genetic and evolutionary computation},
  \bibinfo{organization}{ACM}, \bibinfo{pages}{1795--1802},
  \bibinfo{year}{2005}.

\bibitem[{Zhang and Qi(2005)}]{Zhang2005501}
\bibinfo{author}{G.~Zhang}, \bibinfo{author}{M.~Qi}, \bibinfo{title}{Neural
  network forecasting for seasonal and trend time series},
  \bibinfo{journal}{European Journal of Operational Research}
  \bibinfo{volume}{160}~(\bibinfo{number}{2}) (\bibinfo{year}{2005})
  \bibinfo{pages}{501 -- 514}, ISSN \bibinfo{issn}{0377-2217},
  \urlprefix\url{http://doi.org/10.1016/j.ejor.2003.08.037},
  \bibinfo{note}{decision Support Systems in the Internet Age}.

\bibitem[{Orange(2013)}]{D4Dwebsite}
\bibinfo{author}{Orange}, \bibinfo{title}{D4D challenge},
  \bibinfo{howpublished}{\url{http://http://www.d4d.orange.com/en/Accueil}},
  \bibinfo{note}{accessed: 2016-09-22}, \bibinfo{year}{2013}.

\bibitem[{Blondel et~al.(2012)Blondel, Esch, Chan, Cl{\'e}rot, Deville, Huens,
  Morlot, Smoreda, and Ziemlicki}]{DBLP:journals/corr/abs-1210-0137}
\bibinfo{author}{V.~D. Blondel}, \bibinfo{author}{M.~Esch},
  \bibinfo{author}{C.~Chan}, \bibinfo{author}{F.~Cl{\'e}rot},
  \bibinfo{author}{P.~Deville}, \bibinfo{author}{E.~Huens},
  \bibinfo{author}{F.~Morlot}, \bibinfo{author}{Z.~Smoreda},
  \bibinfo{author}{C.~Ziemlicki}, \bibinfo{title}{{D}ata for {D}evelopment: the
  {D4D} {C}hallenge on {M}obile {P}hone {D}ata}, \bibinfo{journal}{ArXiv
  preprint arXiv:1210.0137} .

\bibitem[{Bianchi et~al.(2016{\natexlab{b}})Bianchi, Rizzi, Sadeghian, and
  Moiso}]{Bianchi201649}
\bibinfo{author}{F.~M. Bianchi}, \bibinfo{author}{A.~Rizzi},
  \bibinfo{author}{A.~Sadeghian}, \bibinfo{author}{C.~Moiso},
  \bibinfo{title}{Identifying user habits through data mining on call data
  records}, \bibinfo{journal}{Engineering Applications of Artificial
  Intelligence} \bibinfo{volume}{54} (\bibinfo{year}{2016}{\natexlab{b}})
  \bibinfo{pages}{49 -- 61}, ISSN \bibinfo{issn}{0952-1976},
  \urlprefix\url{http://dx.doi.org/10.1016/j.engappai.2016.05.007}.

\bibitem[{Franses(1991)}]{franses1991seasonality}
\bibinfo{author}{P.~H. Franses}, \bibinfo{title}{Seasonality, non-stationarity
  and the forecasting of monthly time series}, \bibinfo{journal}{International
  Journal of Forecasting} \bibinfo{volume}{7}~(\bibinfo{number}{2})
  (\bibinfo{year}{1991}) \bibinfo{pages}{199--208}.

\bibitem[{Shen and Huang(2005)}]{shen2005analysis}
\bibinfo{author}{H.~Shen}, \bibinfo{author}{J.~Z. Huang},
  \bibinfo{title}{{A}nalysis of call centre arrival data using singular value
  decomposition}, \bibinfo{journal}{{A}pplied {S}tochastic {M}odels in
  {B}usiness and {I}ndustry} \bibinfo{volume}{21}~(\bibinfo{number}{3})
  (\bibinfo{year}{2005}) \bibinfo{pages}{251--263}.

\bibitem[{Ibrahim and L'Ecuyer(2013)}]{ibrahim2013forecasting}
\bibinfo{author}{R.~Ibrahim}, \bibinfo{author}{P.~L'Ecuyer},
  \bibinfo{title}{{F}orecasting call center arrivals: {F}ixed-effects,
  mixed-effects, and bivariate models}, \bibinfo{journal}{{M}anufacturing \&
  {S}ervice {O}perations {M}anagement}
  \bibinfo{volume}{15}~(\bibinfo{number}{1}) (\bibinfo{year}{2013})
  \bibinfo{pages}{72--85}.

\bibitem[{Andrews and Cunningham(1995)}]{andrews1995ll}
\bibinfo{author}{B.~H. Andrews}, \bibinfo{author}{S.~M. Cunningham},
  \bibinfo{title}{{LL} {B}ean improves call-center forecasting},
  \bibinfo{journal}{{I}nterfaces} \bibinfo{volume}{25}~(\bibinfo{number}{6})
  (\bibinfo{year}{1995}) \bibinfo{pages}{1--13}.

\bibitem[{Zhang and Kline(2007)}]{4359174}
\bibinfo{author}{G.~P. Zhang}, \bibinfo{author}{D.~M. Kline},
  \bibinfo{title}{Quarterly Time-Series Forecasting With Neural Networks},
  \bibinfo{journal}{IEEE Transactions on Neural Networks}
  \bibinfo{volume}{18}~(\bibinfo{number}{6}) (\bibinfo{year}{2007})
  \bibinfo{pages}{1800--1814}, ISSN \bibinfo{issn}{1045-9227},
  \urlprefix\url{10.1109/TNN.2007.896859}.

\bibitem[{Claveria et~al.(0)Claveria, Monte, and
  Torra}]{doi:10.3846/20294913.2015.1070772}
\bibinfo{author}{O.~Claveria}, \bibinfo{author}{E.~Monte},
  \bibinfo{author}{S.~Torra}, \bibinfo{title}{Data pre-processing for neural
  network-based forecasting: does it really matter?},
  \bibinfo{journal}{Technological and Economic Development of Economy}
  \bibinfo{volume}{0}~(\bibinfo{number}{0}) (\bibinfo{year}{0})
  \bibinfo{pages}{1--17}, \urlprefix\url{10.3846/20294913.2015.1070772}.

\bibitem[{Weinberg et~al.(2007)Weinberg, Brown, and
  Stroud}]{weinberg2007bayesian}
\bibinfo{author}{J.~Weinberg}, \bibinfo{author}{L.~D. Brown},
  \bibinfo{author}{J.~R. Stroud}, \bibinfo{title}{{B}ayesian forecasting of an
  inhomogeneous {P}oisson process with applications to call center data},
  \bibinfo{journal}{{J}ournal of the {A}merican {S}tatistical {A}ssociation}
  \bibinfo{volume}{102}~(\bibinfo{number}{480}) (\bibinfo{year}{2007})
  \bibinfo{pages}{1185--1198}.

\bibitem[{Santis et~al.(2015)Santis, Livi, Sadeghian, and
  Rizzi}]{DeSantis2015368}
\bibinfo{author}{E.~D. Santis}, \bibinfo{author}{L.~Livi},
  \bibinfo{author}{A.~Sadeghian}, \bibinfo{author}{A.~Rizzi},
  \bibinfo{title}{Modeling and recognition of smart grid faults by a combined
  approach of dissimilarity learning and one-class classification},
  \bibinfo{journal}{Neurocomputing} \bibinfo{volume}{170}
  (\bibinfo{year}{2015}) \bibinfo{pages}{368 -- 383}, ISSN
  \bibinfo{issn}{0925-2312},
  \urlprefix\url{http://dx.doi.org/10.1016/j.neucom.2015.05.112},
  \bibinfo{note}{advances on Biological Rhythmic Pattern Generation:
  Experiments, Algorithms and ApplicationsSelected Papers from the 2013
  International Conference on Intelligence Science and Big Data Engineering
  (IScIDE 2013)Computational Energy Management in Smart Grids}.

\bibitem[{Kaggle(2012)}]{GEFCom2012}
\bibinfo{author}{Kaggle}, \bibinfo{title}{{GEFCom 2012} Global Energy
  Forecasting Competition 2012},
  \bibinfo{howpublished}{\url{https://www.kaggle.com/c/global-energy-forecasting-competition-2012-load-forecasting}},
  \bibinfo{note}{accessed: 2017-04-26}, \bibinfo{year}{2012}.

\bibitem[{Bergstra and Bengio(2012)}]{bergstra2012random}
\bibinfo{author}{J.~Bergstra}, \bibinfo{author}{Y.~Bengio},
  \bibinfo{title}{Random search for hyper-parameter optimization},
  \bibinfo{journal}{The Journal of Machine Learning Research}
  \bibinfo{volume}{13}~(\bibinfo{number}{1}) (\bibinfo{year}{2012})
  \bibinfo{pages}{281--305}.

\bibitem[{{Theano Development Team}(2016)}]{2016arXiv160502688short}
\bibinfo{author}{{Theano Development Team}}, \bibinfo{title}{{Theano: A
  {Python} framework for fast computation of mathematical expressions}},
  \bibinfo{journal}{arXiv e-prints} \bibinfo{volume}{abs/1605.02688}.

\bibitem[{Zeyer et~al.(2016)Zeyer, Doetsch, Voigtlaender, Schl{\"{u}}ter, and
  Ney}]{DBLP:journals/corr/ZeyerDVSN16}
\bibinfo{author}{A.~Zeyer}, \bibinfo{author}{P.~Doetsch},
  \bibinfo{author}{P.~Voigtlaender}, \bibinfo{author}{R.~Schl{\"{u}}ter},
  \bibinfo{author}{H.~Ney}, \bibinfo{title}{A Comprehensive Study of Deep
  Bidirectional {LSTM} RNNs for Acoustic Modeling in Speech Recognition},
  \bibinfo{journal}{CoRR} \bibinfo{volume}{abs/1606.06871}.

\bibitem[{Pham et~al.(2014{\natexlab{b}})Pham, Bluche, Kermorvant, and
  Louradour}]{6981034}
\bibinfo{author}{V.~Pham}, \bibinfo{author}{T.~Bluche},
  \bibinfo{author}{C.~Kermorvant}, \bibinfo{author}{J.~Louradour},
  \bibinfo{title}{Dropout Improves Recurrent Neural Networks for Handwriting
  Recognition}, in: \bibinfo{booktitle}{2014 14th International Conference on
  Frontiers in Handwriting Recognition}, ISSN \bibinfo{issn}{2167-6445},
  \bibinfo{pages}{285--290}, \urlprefix\url{10.1109/ICFHR.2014.55},
  \bibinfo{year}{2014}{\natexlab{b}}.

\bibitem[{Srivastava et~al.(2014)Srivastava, Hinton, Krizhevsky, Sutskever, and
  Salakhutdinov}]{srivastava2014dropout}
\bibinfo{author}{N.~Srivastava}, \bibinfo{author}{G.~Hinton},
  \bibinfo{author}{A.~Krizhevsky}, \bibinfo{author}{I.~Sutskever},
  \bibinfo{author}{R.~Salakhutdinov}, \bibinfo{title}{Dropout: A simple way to
  prevent neural networks from overfitting}, \bibinfo{journal}{The Journal of
  Machine Learning Research} \bibinfo{volume}{15}~(\bibinfo{number}{1})
  (\bibinfo{year}{2014}) \bibinfo{pages}{1929--1958}.

\bibitem[{Abadi et~al.(2015)Abadi, Agarwal, Barham, Brevdo, Chen, Citro,
  Corrado, Davis, Dean, Devin, Ghemawat, Goodfellow, Harp, Irving, Isard, Jia,
  Jozefowicz, Kaiser, Kudlur, Levenberg, Man\'{e}, Moore, Murray, Olah,
  Schuster, Shlens, Steiner, Sutskever, Talwar, Tucker, Vanhoucke, Vasudevan,
  Vi\'{e}gas, Vinyals, Warden, Wattenberg, Wicke, Yu, and
  Zheng}]{tensorflow2015-whitepaper}
\bibinfo{author}{M.~Abadi}, \bibinfo{author}{A.~Agarwal},
  \bibinfo{author}{P.~Barham}, \bibinfo{author}{E.~Brevdo},
  \bibinfo{author}{Z.~Chen}, \bibinfo{author}{C.~Citro}, \bibinfo{author}{G.~S.
  Corrado}, \bibinfo{author}{A.~Davis}, \bibinfo{author}{J.~Dean},
  \bibinfo{author}{M.~Devin}, \bibinfo{author}{S.~Ghemawat},
  \bibinfo{author}{I.~Goodfellow}, \bibinfo{author}{A.~Harp},
  \bibinfo{author}{G.~Irving}, \bibinfo{author}{M.~Isard},
  \bibinfo{author}{Y.~Jia}, \bibinfo{author}{R.~Jozefowicz},
  \bibinfo{author}{L.~Kaiser}, \bibinfo{author}{M.~Kudlur},
  \bibinfo{author}{J.~Levenberg}, \bibinfo{author}{D.~Man\'{e}},
  \bibinfo{author}{D.~S. Moore}, \bibinfo{author}{D.~Murray},
  \bibinfo{author}{C.~Olah}, \bibinfo{author}{M.~Schuster},
  \bibinfo{author}{J.~Shlens}, \bibinfo{author}{B.~Steiner},
  \bibinfo{author}{I.~Sutskever}, \bibinfo{author}{K.~Talwar},
  \bibinfo{author}{P.~Tucker}, \bibinfo{author}{V.~Vanhoucke},
  \bibinfo{author}{V.~Vasudevan}, \bibinfo{author}{F.~Vi\'{e}gas},
  \bibinfo{author}{O.~Vinyals}, \bibinfo{author}{P.~Warden},
  \bibinfo{author}{M.~Wattenberg}, \bibinfo{author}{M.~Wicke},
  \bibinfo{author}{Y.~Yu}, \bibinfo{author}{X.~Zheng},
  \bibinfo{title}{{TensorFlow}: Large-Scale Machine Learning on Heterogeneous
  Systems}, \urlprefix\url{http://tensorflow.org/}, \bibinfo{note}{software
  available from tensorflow.org}, \bibinfo{year}{2015}.

\bibitem[{Battiti(1992)}]{battiti1992first}
\bibinfo{author}{R.~Battiti}, \bibinfo{title}{First-and second-order methods
  for learning: between steepest descent and Newton's method},
  \bibinfo{journal}{Neural computation}
  \bibinfo{volume}{4}~(\bibinfo{number}{2}) (\bibinfo{year}{1992})
  \bibinfo{pages}{141--166}.

\bibitem[{Li et~al.(2012{\natexlab{b}})Li, Han, and Wang}]{6177672}
\bibinfo{author}{D.~Li}, \bibinfo{author}{M.~Han}, \bibinfo{author}{J.~Wang},
  \bibinfo{title}{Chaotic Time Series Prediction Based on a Novel Robust Echo
  State Network}, \bibinfo{journal}{IEEE Transactions on Neural Networks and
  Learning Systems} \bibinfo{volume}{23}~(\bibinfo{number}{5})
  (\bibinfo{year}{2012}{\natexlab{b}}) \bibinfo{pages}{787--799}, ISSN
  \bibinfo{issn}{2162-237X}, \urlprefix\url{10.1109/TNNLS.2012.2188414}.

\bibitem[{Shi and Han(2007)}]{shi2007support}
\bibinfo{author}{Z.~Shi}, \bibinfo{author}{M.~Han}, \bibinfo{title}{Support
  vector echo-state machine for chaotic time-series prediction},
  \bibinfo{journal}{Neural Networks, IEEE Transactions on}
  \bibinfo{volume}{18}~(\bibinfo{number}{2}) (\bibinfo{year}{2007})
  \bibinfo{pages}{359--372}.

\bibitem[{Kumar et~al.(2016)Kumar, Irsoy, Ondruska, Iyyer, Bradbury, Gulrajani,
  Zhong, Paulus, and Socher}]{pmlr-v48-kumar16}
\bibinfo{author}{A.~Kumar}, \bibinfo{author}{O.~Irsoy},
  \bibinfo{author}{P.~Ondruska}, \bibinfo{author}{M.~Iyyer},
  \bibinfo{author}{J.~Bradbury}, \bibinfo{author}{I.~Gulrajani},
  \bibinfo{author}{V.~Zhong}, \bibinfo{author}{R.~Paulus},
  \bibinfo{author}{R.~Socher}, \bibinfo{title}{Ask Me Anything: Dynamic Memory
  Networks for Natural Language Processing}, in: \bibinfo{editor}{M.~F.
  Balcan}, \bibinfo{editor}{K.~Q. Weinberger} (Eds.),
  \bibinfo{booktitle}{Proceedings of The 33rd International Conference on
  Machine Learning}, vol.~\bibinfo{volume}{48} of
  \emph{\bibinfo{series}{Proceedings of Machine Learning Research}},
  \bibinfo{publisher}{PMLR}, \bibinfo{address}{New York, New York, USA},
  \bibinfo{pages}{1378--1387}, \bibinfo{year}{2016}.

\end{thebibliography}
\end{document}